\documentclass[letterpaper, 10pt, conference]{format/ieeeconf}

\IEEEoverridecommandlockouts    
\usepackage{glossaries}

\usepackage{cite}
\usepackage{amsmath,amssymb,amsfonts}
\usepackage{algorithmic}
\usepackage{graphicx}
\usepackage{textcomp}
\usepackage{xcolor}
\usepackage{mwe}
\usepackage{wrapfig}
\usepackage[normalem]{ulem}
\usepackage{graphicx}
\usepackage{tabularx}
\usepackage{booktabs}
\usepackage{multirow}
\usepackage[utf8]{inputenc}
\usepackage{comment}
\usepackage{subcaption}  

\makeatletter
\DeclareRobustCommand\onedot{\futurelet\@let@token\@onedot}
\def\@onedot{\ifx\@let@token.\else.\null\fi\xspace}

\def\eg{\emph{e.g}\onedot}

\makeatother

\def\BibTeX{{\rm B\kern-.05em{\sc i\kern-.025em b}\kern-.08em
    T\kern-.1667em\lower.7ex\hbox{E}\kern-.125emX}}
\begin{document}

\title{\Huge Real-Time Glass Detection and Reprojection using Sensor Fusion Onboard Aerial Robots 
}

\author{Malakhi Hopkins, Varun Murali, Vijay Kumar and Camillo J. Taylor  \\
\thanks{We gratefully acknowledge the support of ARL DCIST CRA W911NF-17-2-0181, NSF Grant CCR-2112665, the NSF Graduate Research Fellowship Grant No. DGE-1845298, and the IoT4Ag Engineering Research Center funded by the National Science Foundation (NSF) under NSF Cooperative Agreement Number EEC-1941529.
}
}

\newcommand{\VK}[1]{\textcolor{purple}{\textbf{VK:} #1}} 
\newcommand{\VM}[1]{\textcolor{red}{\textbf{VM:} #1}} 
\newcommand{\MH}[1]{\textcolor{blue}{\textbf{MH:} #1}} 

\maketitle

\begin{abstract}

Autonomous aerial robots are increasingly being deployed in real-world scenarios, where transparent obstacles present significant challenges to reliable navigation and mapping. These materials pose a unique problem for traditional perception systems because they lack discernible features and can cause conventional depth sensors to fail, leading to inaccurate maps and potential collisions. To ensure safe navigation, robots must be able to accurately detect and map these transparent obstacles. Existing methods often rely on large, expensive sensors or algorithms that impose high computational burdens, making them unsuitable for low Size, Weight, and Power (SWaP) robots. In this work, we propose 
a novel and computationally efficient framework for detecting and mapping transparent obstacles onboard a sub-300g quadrotor. Our method fuses data from a Time-of-Flight (ToF) camera and an ultrasonic sensor with a custom, lightweight 2D convolution model. This specialized approach accurately detects specular reflections and propagates their depth into corresponding empty regions of the depth map, effectively rendering transparent obstacles visible. The entire pipeline operates in real-time, utilizing only a small fraction of a CPU core on an embedded processor. We validate our system through a series of experiments in both controlled and real-world environments, demonstrating the utility of our method through experiments where the robot maps indoor environments containing glass. Our work is, to our knowledge, the first of its kind to demonstrate a real-time, onboard transparent obstacle mapping system on a low-SWaP quadrotor using only the CPU. 

\end{abstract}

\section{Introduction}
\label{sec:introduction}

Safe and effective robot navigation is contingent on a robot’s ability to accurately perceive its environment. 
While prior work has successfully addressed volumetric mapping, most methods rely on exteroceptive sensors like Time-of-Flight (ToF), RGB-D sensors or LiDAR, which fail in the presence of transparent or reflective surfaces common in indoor spaces. 
These materials present a unique challenge as they lack discernible surface features and reflect light variably depending on the viewing angle, making it difficult for conventional depth sensors to obtain precise data. 
This can lead to inaccurate maps and potential collisions that damage the robot or its surroundings. 

\begin{figure} [t!]
    \centering
    \includegraphics[width=0.45\textwidth]{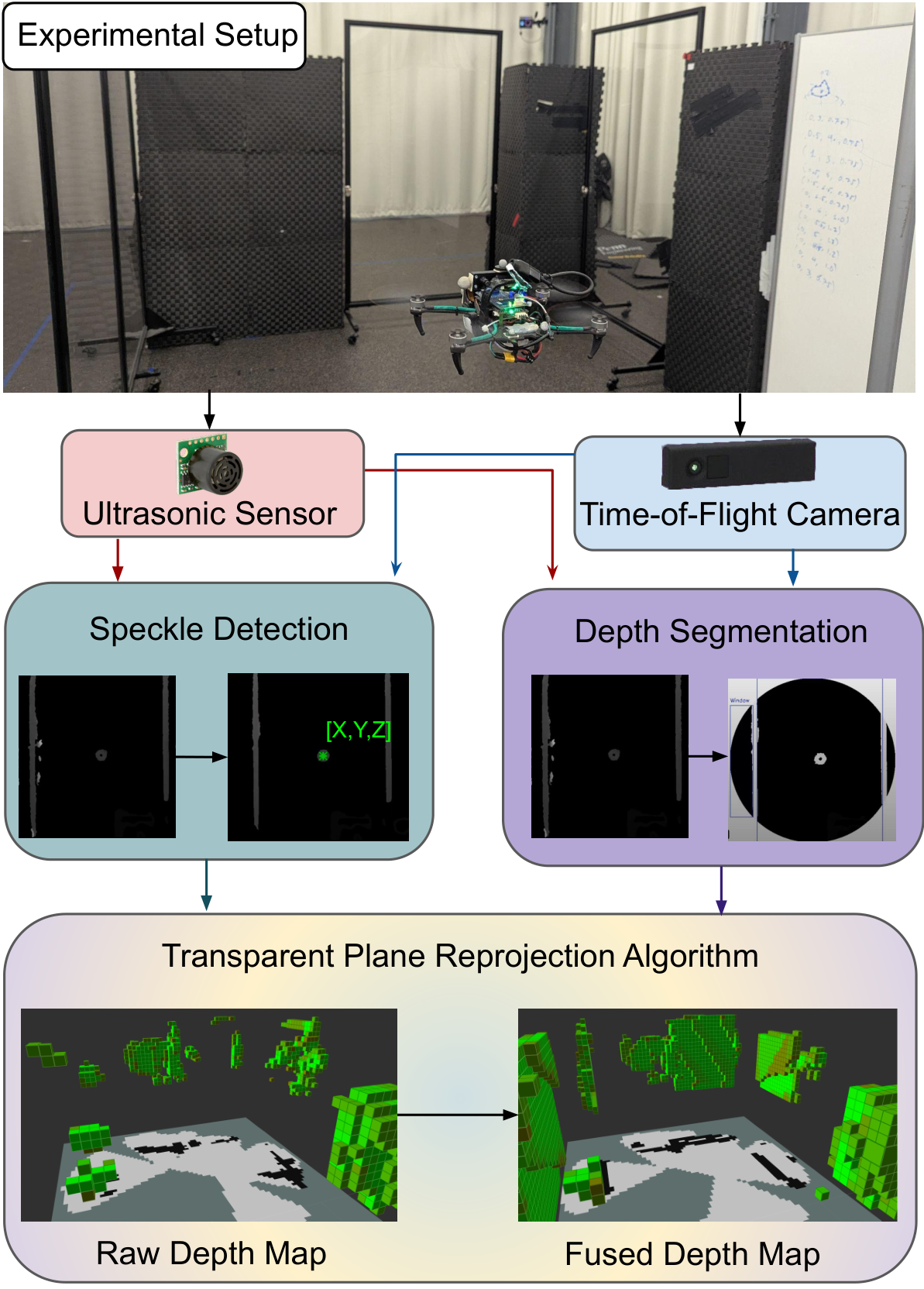}
    \caption{The experimental setup and algorithmic overview of our framework. Our methodology integrates data from an ultrasonic sensor and a ToF camera to detect specular reflections (speckles) on glass surfaces and segment the depth image based on empty space. This information is then used to convert the raw depth map into a fused depth map by reprojecting the speckle depth into its respective segmented region, effectively making glass visible for safe navigation and mapping.}
    \label{fig:motivation-figure}
    \vspace{-0.6cm}
    
\end{figure}

Researchers have explored various methods to address the challenge of detecting transparent glass obstacles in the environment. 
On one hand, some studies have focused on detecting glass using a variety of non-contact sensors, including RGB cameras and LiDARs.
These methods leverage advancements in machine learning to achieve high accuracy without physical contact. 
However, their performances are often influenced by factors such as lighting conditions and angular uncertainty, and they can be computationally expensive to run in real-time on small, low Size, Weight, and Power (SWaP) aerial robots. 
Additionally, high latency from these models can compromise the robot's operational speed for safety. 

On the other hand, an emerging class of solutions has explored active, contact-based methodologies. 
These systems may use visual sensors to identify potential glass surfaces and then actively perform a gentle tactile engagement to confirm or invalidate their presence. 
While this approach offers robustness by providing physical ground truth, it comes at the cost of increased mechanical complexity, and the need to physically interact with the environment, which can be physically intrusive and inefficient.
%
Motivated by the need for a non-intrusive and computationally-lightweight solution for Size, Weight, and Power (SWaP)-constrained platforms, we propose a novel, non-contact approach. To our knowledge, our work is the first of its kind to demonstrate a real-time, onboard transparent obstacle mapping system on a sub-300g quadrotor using only the CPU. In summary, our key contributions are as follows:
\begin{enumerate}
    \item A novel onboard glass obstacle mapping system for low-SWaP aerial robots that integrates compact Time-of-Flight (ToF) and Ultrasonic sensors to enable robust glass detection. 
    \item A lightweight, computationally efficient framework that operates in real-time on an embedded processor, utilizing a custom 2D convolution and depth image segmentation algorithm to detect specular reflections and accurately reproject glass planes. 
    \item An extensive experimental validation of our framework in real-time onboard a SWaP-constrained aerial robot, confirming its ability to accurately and efficiently handle glass planes.
\end{enumerate}
\section{Related Work}
\label{sec:related_work}

In robotics, accurate perception is key for reliable autonomous navigation. This section summarizes existing methods for detecting transparent obstacles, which fall into three categories: passive or non-contact, multi-sensor fusion, and active or contact-based engagement strategies.

\subsection{Passive, Non-Contact Glass Detection}
A significant body of research has focused on passive, non-contact methods to perceive transparent obstacles by exploiting their unique physical and visual properties.

\noindent \textbf{Methods Exploiting Physical Properties:}
Early approaches capitalized on the geometric and reflective properties of transparent materials. Methods like those by Yang et al. \cite{yang2010solving} utilize mirror symmetry to identify and reconstruct mirrors in structured environments. Similarly, research has leveraged the unique behavior of multi-return LiDAR \cite{koch2016detection, koch2017identification, koch2017effective, koch2017detection}, where differences in scan messages or point cloud disparities are used to detect reflective surfaces. While effective, these methods often operate offline, are limited to 2D mapping, or are restricted to specific environments. Other research has explored the use of light absorption in certain materials to detect them \cite{tibebu2021lidar}.

\noindent \textbf{Machine Learning for Perception:}
Recent advancements in machine learning have enabled sophisticated approaches to glass detection. Several works use machine learning to estimate depth from contextual clues in a monocular image alone \cite{bhat2023zoedepth, simon2023mononav, birkl2023midas}. Newer foundation models trained on massive datasets can even predict absolute depth \cite{yang2024depth}, but they are often too computationally expensive and have a high latency, making them impractical for real-time operation on low-SWaP aerial robots. The same issue applies to open-world segmentation models with large vision transformer backbones that are capable of detecting transparent objects \cite{ren2024grounded}. Research has also focused on dedicated networks for transparent object segmentation \cite{mei2020don, mei2022glass, xie2020segmenting, lin2025rgbd, neurips2022:gsds2022} and depth completion \cite{costanzino2023learning}, often relying on large datasets to train the models.

\noindent \textbf{Multi-Sensor Fusion Methodologies:}
Another class of solutions attempts to overcome the limitations of individual sensors by fusing data from multiple modalities. These methods combine sensor streams with complementary strengths to provide a more robust and complete representation of the environment. A comprehensive review of various sensor fusion methods in mobile robot navigation has been presented in recent literature, covering applications in mapping and path planning.

\noindent \textbf{RGB-D and Ultrasonic Fusion:}
A common fusion approach combines RGB-D cameras with ultrasonic sensors. The RGB-D camera can detect a window's frame or other opaque features, while the ultrasonic sensor provides a range reading within that region to confirm the presence of a transparent surface \cite{huang2018glass}. While this is effective for framed windows or objects with both transparent and opaque features, it is not a general solution for standalone glass panes. \cite{mao2015glass} incorporate a single ultrasonic sensor with a kinect2 depth camera to reconstruct objects that contatin glass. When an object with both transparent and opaque features is observed through the kinect sensor and the ultrasonic sensor detects range readings in the transparent sections of that object, a plane is filled in between the opaque features of the object that corresponds to the ultrasonic range detection.

\noindent \textbf{LiDAR and Other Sensor Fusion:}
LiDAR has been fused with various sensors to detect glass. For instance, some works combine LiDAR with a 3D sonar to better identify objects like mannequins and glass windows in an office environment \cite{tran2021environment}. 
Other approaches fuse LiDAR with infrared cameras, leveraging the LiDAR's ability to detect light return variance and the infrared camera's detection of specular reflections \cite{zhang2022glass}. A more complex approach fuses LiDAR with a polarization camera, where the LiDAR detects glass at low incidence angles and the polarization camera detects it at large incidence angles \cite{yamaguchi2020glass}. However, these methods often rely on expensive sensors, operate in 2D, or make assumptions about the environment's geometry.

\subsection{Active and Contact-Based Methods}
An emerging paradigm shifts away from purely passive perception by incorporating physical interaction as a source of sensing. These methods are often inspired by human intuition and are designed to provide absolute certainty about the presence of an obstacle.

\noindent \textbf{Contact-Resilient Aerial Vehicles:}
A number of studies have focused on designing aerial vehicles that can withstand collisions with obstacles \cite{shen2012indoor}. These designs include protective frames, collision bumpers, and compliant arms. While these mechanical designs can prevent crashes, they often come with increased weight and complexity, which compromises flight duration and maneuverability. Some of these methods also rely on passive collision response, which is inefficient and limits the utility of contact as a source of environmental sensing.

\noindent \textbf{Autonomous Navigation with Active Contact:}
Motivated by the limitations of passive methods, an active approach has been developed that combines visual detection with physical confirmation \cite{chen2025active}. This system uses a visual glass detection module to identify potential glass surfaces and then actively performs a gentle ``touch action" to confirm or invalidate their presence using a lightweight contact-sensing module. This provides a form of ``unassailable ground truth" that resolves ambiguities and false positives that purely visual systems might miss. While robust, this approach is inherently slower and more physically intrusive, requiring the vehicle to decelerate to perform the touch action and then return to a safe pose for replanning. The work presented in \cite{chen2025active} is to our knowledge the first to address transparent obstacles in aerial navigation through such an active approach.

In contrast to prior work, we specifically address the challenge of real-time, onboard, non-contact transparent obstacle detection and mapping for low-SWaP aerial robots. 
Our method leverages sensor fusion of low-cost sensors, and an efficient, custom-designed algorithm to exploit the properties of the sensor data, providing a robust, non-intrusive alternative to the contact-based methods while maintaining computational efficiency for constrained platforms.

\section{Methodology}
\label{sec:method}

This work proposes a novel system and method to overcome the limitations of conventional sensors and computationally heavy algorithms by providing a non-contact, computationally efficient, and light-invariant solution for transparent object mapping on low-SWaP aerial robots. We aim to enable a small flying robot to confidently detect, segment, and map stand-alone glass objects, generating a mask that closely approximates the bounds of the glass to ensure accurate mapping. This section details our methodology for transparent plane reprojection. We first describe the system's hardware (\S \ IV-A), followed by the morphology of a speckle (\S \ IV-B). Next we detail our speckle detection and glass segmentation algorithm and sensor fusion approach (\S \ IV-C). Finally, we explain the augmentation of the ToF depth image for glass plane mapping (\S \ IV-D). 

\subsection{System Design}
\label{sec:system_design}

Our base platform is the ModalAI Starling 2, which has a Qualcomm QRB5165 processor on a VOXL2 board. This setup includes 8 cores and 8GB LPDDR5 RAM. The robot's small size (280g) is well-suited for our experiments. We've integrated a PMD Flexx 2 VGA Time-of-Flight sensor, which provides a 640x480 resolution depth image and a 56° x 44° FoV. We've also added a MaxSonar MB1020 Ultrasonic sensor that provides a single-point depth measurement up to 5 meters. Both sensors are aligned to measure the same general region.

\begin{figure}[!th]
    \centering
    \includegraphics[width=0.9\linewidth]{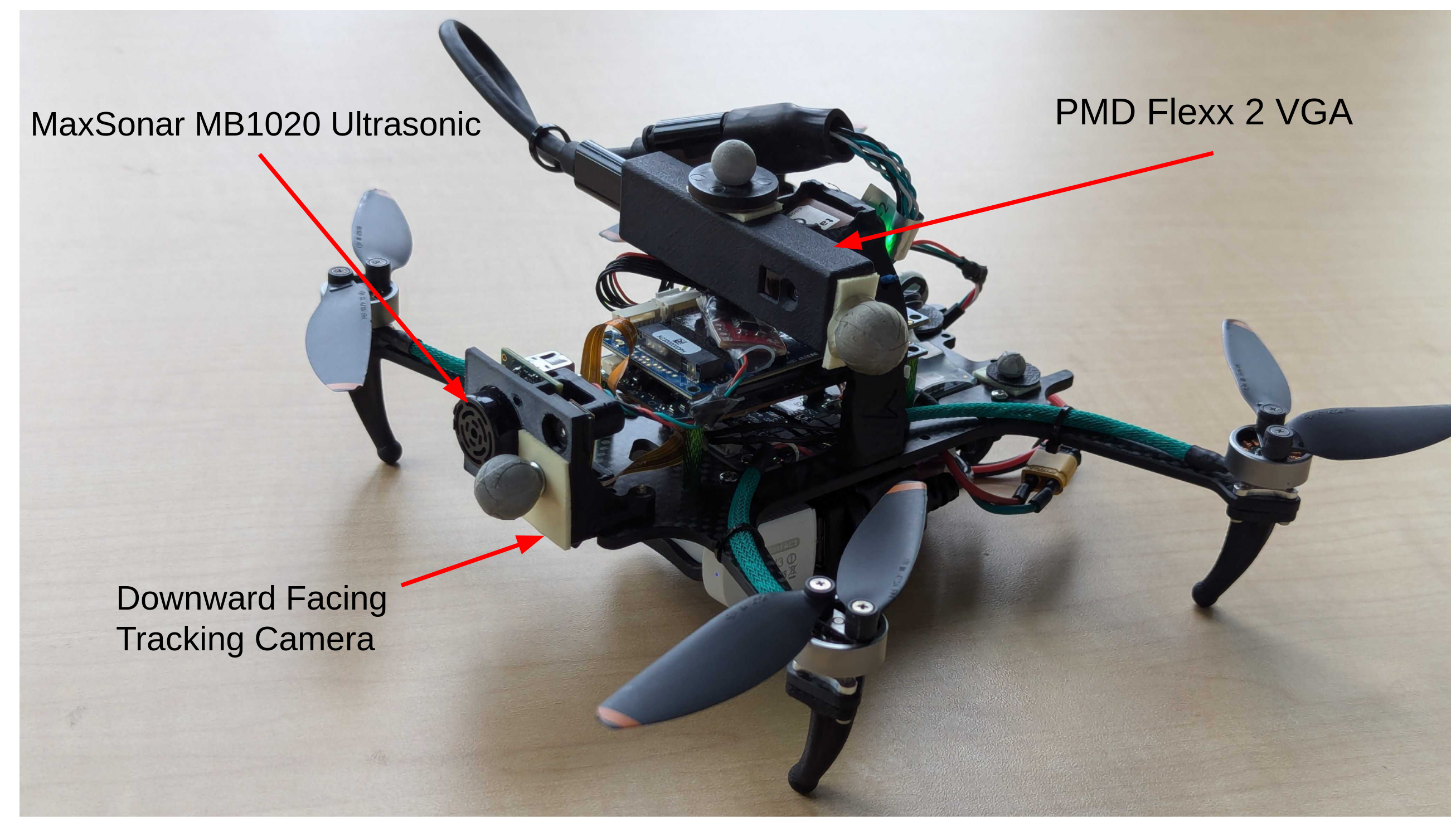}
    \caption{Our robot is a  ModalAI Starling 2 platform with a PMD Flexx 2 VGA Time-of-Flight sensor and a MaxSonar MB1020 Ultrasonic sensor.
    \vspace{-1em}
    }
    \label{fig:system}
\end{figure}

\subsection{Specular Reflection Morphology}
\label{sec:yolo}

\begin{figure}[!h]
\centering
\begin{tabular}{c}
{\includegraphics[width = 0.98\linewidth]{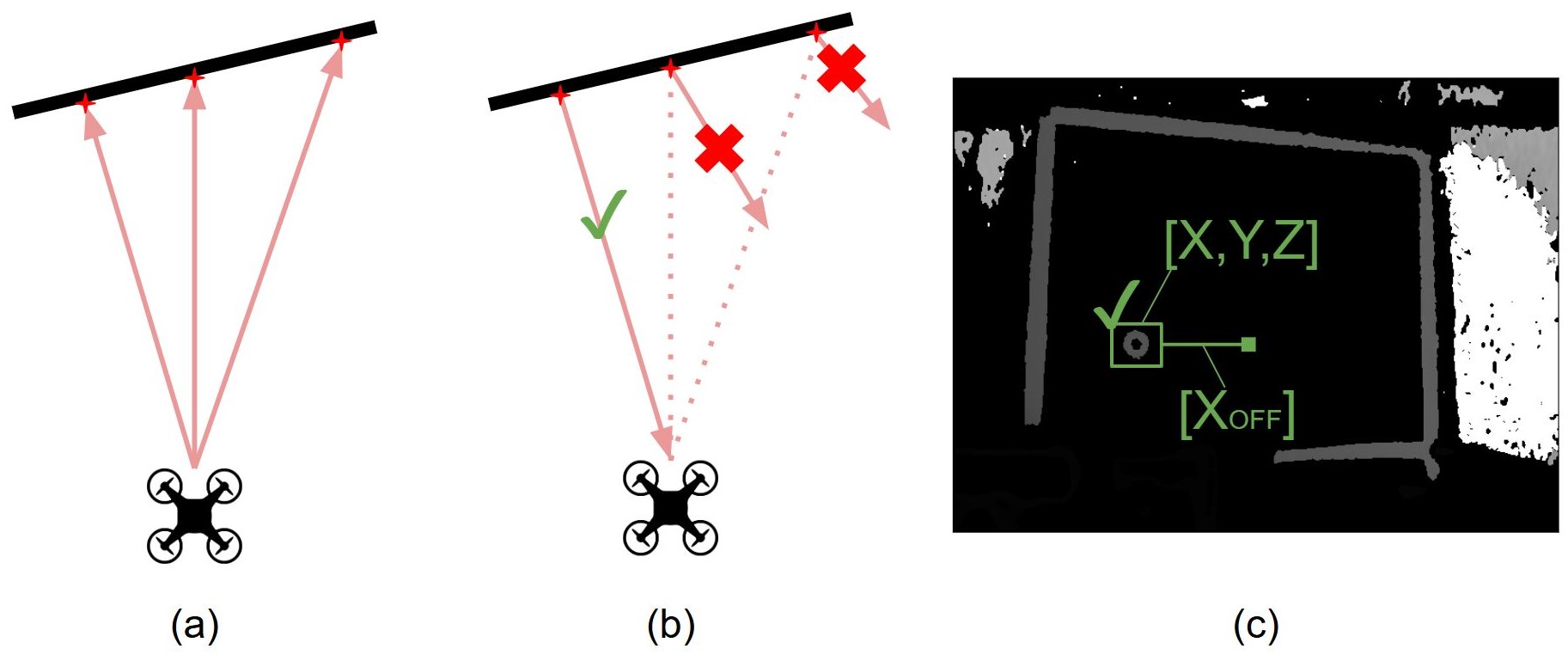}}
\end{tabular}
\caption{An Illustration of speckle formation. (a) Depicts three light rays from the ToF sensor approaching a tilted glass surface. (b) The light rays reflect, and the one perpendicular to the glass surface reflects back to the ToF sensor. (c) This returning ray creates the speckle, from which we can extract both the depth and the normal of the glass pane.
\vspace{-1em}
}
\label{fig:speckles_angle}
\end{figure}

In a Time-of-Flight (ToF) depth sensing context, a specular reflection (speckle) occurs when infrared light rays bounce off a smooth surface, like a glass pane, without scattering. This mirror-like reflection returns to the sensor at an angle equal to its incidence. When the ToF sensor is nearly perpendicular to the glass, a portion of these reflected rays return to the receiver as small, high-intensity speckles as shown in Figure \ref{fig:speckles_angle}. The surrounding glass appears as empty space because most light rays are reflected away from the sensor.

\begin{figure}[!h]
\centering
\begin{tabular}{c c c}
\subfloat[Zoomed in]{\includegraphics[width = 0.25\linewidth]{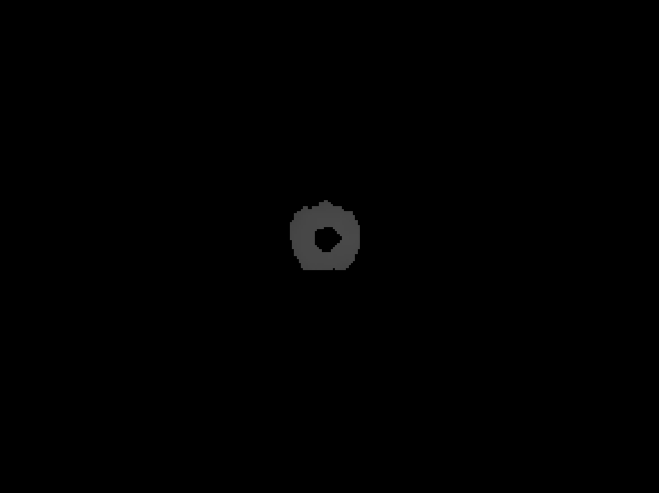}} &
\subfloat[Within 2 m]{\includegraphics[width = 0.25\linewidth]{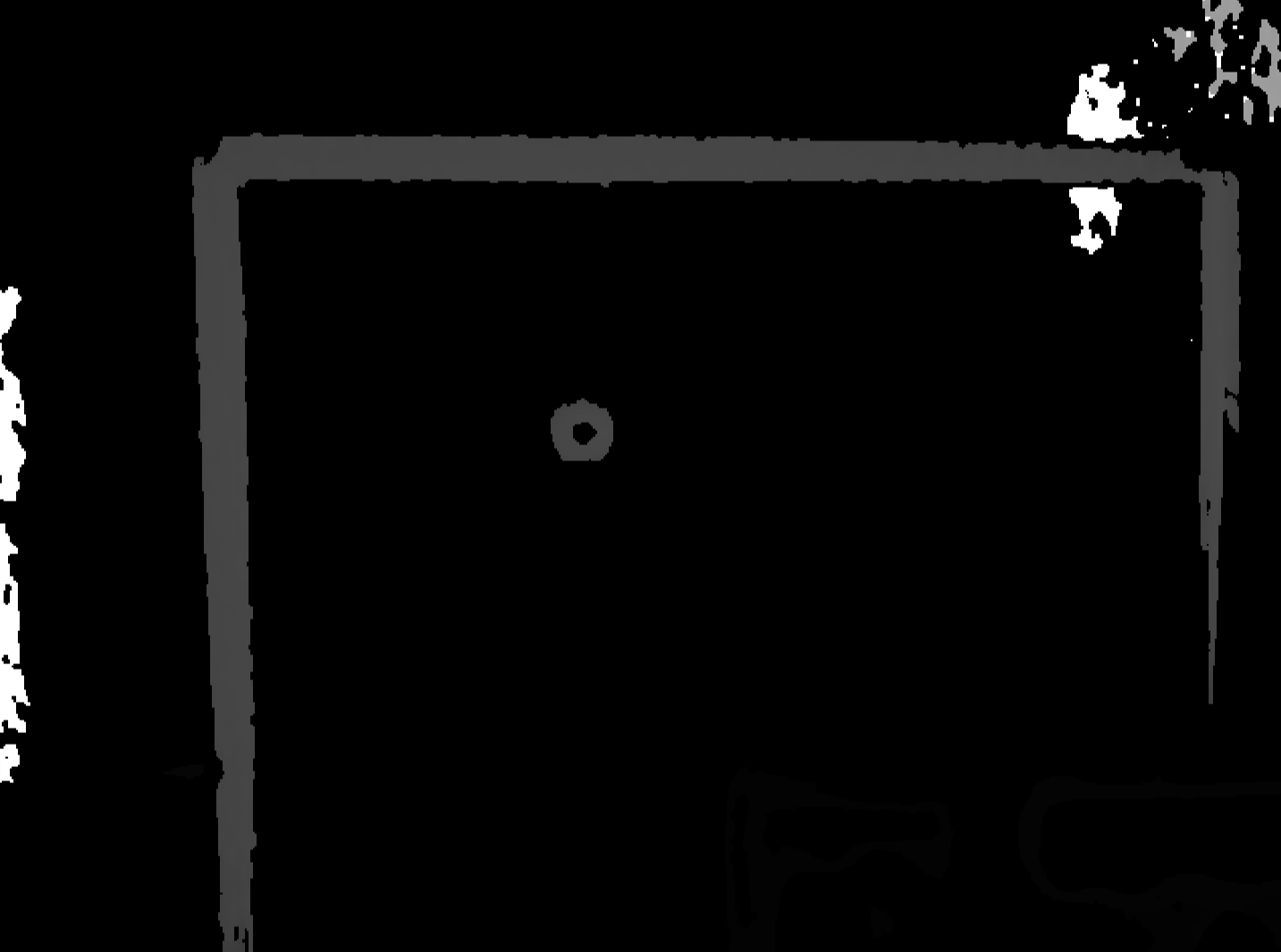}} &
\subfloat[Beyond 2 m]{\includegraphics[width = 0.25\linewidth]{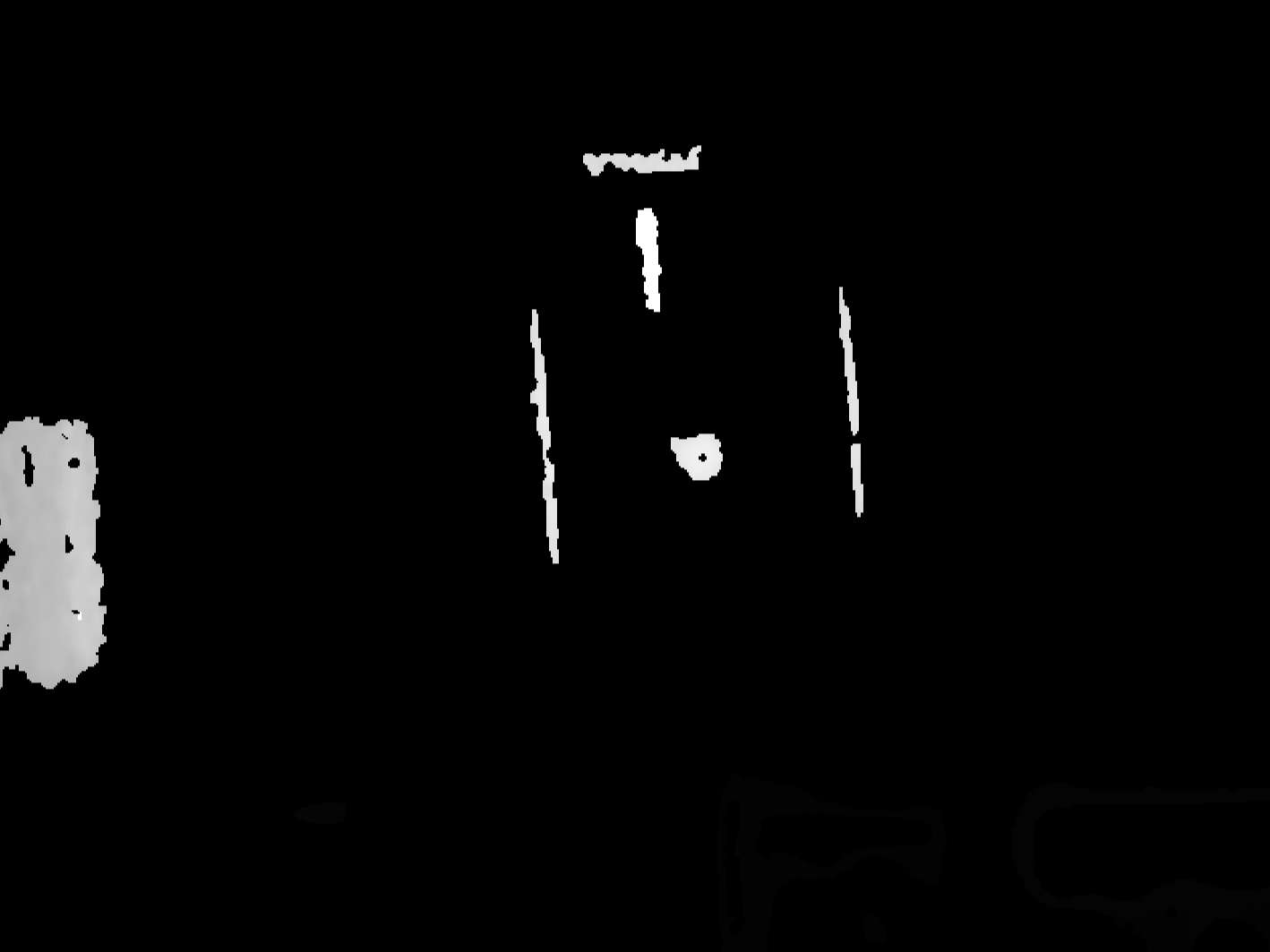}} \\
\end{tabular}
\caption{Examples of specular reflections (speckles) in a Time-of-Flight depth image. (a) A close-up view of a single, isolated speckle. (b) The same speckle within the context of a glass pane at a distance of 2 meters. (c) A different speckle and its environment at a distance greater than 2 meters.}
\label{fig:speckles}
\end{figure}

We utilize these speckles (Figure ~\ref{fig:speckles}) as they provide a valid depth measurement and the true depth of the glass pane. The speckle's location in the image also provides an estimate of the glass pane's normal vector. By analyzing the speckle's depth and position, our algorithm can determine the glass's distance and orientation, enabling us to map the invisible transparent surface as a tangible plane.

\subsection{Speckle Detection and Depth Segmentation Algorithms}


%
To enhance speckle detection, our approach uses the sonar sensor measurement as a filter to remove background clutter from the ToF depth image. Pixels in the ToF image with depth values greater than a small threshold above the sonar's measurement are filtered out, ensuring speckle detection focuses on foreground objects and empty space.

Speckle detection is performed using two custom-designed 2D convolution kernels. These kernels are applied to a Region of Interest (ROI) centered on the image.

\noindent \textbf{Bright Circle Kernel:} A uniform circular mask used to detect bright, reflective points from the ToF sensor's light return. The kernel $K_B$ is defined by:
$$K_B(x, y) = \begin{cases} 1 & \text{if } \sqrt{x^2+y^2} \le r_B \\ 0 & \text{otherwise} \end{cases}$$
where $r_B$ is the kernel's radius (\eg, 21 pixels). The kernel is normalized by dividing by the sum of its elements to ensure a local average.

\noindent \textbf{Dark Ring Kernel:} A kernel designed to detect the absence of data (the 'dark hole') immediately surrounded by valid depth data. The kernel $K_R$ is defined by:
$$K_R(x, y) = \begin{cases} 1 & \text{if } r_{in} < \sqrt{x^2+y^2} \le r_{out} \\ 0 & \text{otherwise} \end{cases}$$
where $r_{in}$ is the inner radius (\eg, 11 pixels) and $r_{out}$ is the outer radius (\eg, 21 pixels). This kernel is also normalized.

The normalized depth image is convolved with both the bright circle and dark ring kernels, producing two response maps. Potential speckle locations are identified as local maxima in both response maps that fall within a score range of 0.3 to 0.9. A successful detection requires a pair of high-scoring peaks from both kernels to be in close proximity, with a Euclidean distance less than the outer kernel's radius.


To validate detected speckles and reduce false positives, we apply a series of filters.

    \noindent \textbf{Circularity Check:} The raw depth data patch around each detected peak is segmented. The circularity of this contour is calculated as:
    $$C = \frac{4\pi \times \text{Area}}{\text{Perimeter}^2}$$
    This value is compared against a threshold (\eg, 0.5) to ensure the detected reflection has a circular, blob-like shape, which is a key characteristic of the speckles.

    \noindent \textbf{Empty Space Verification:} This 
    step distinguishes genuine glass speckles from reflections off background objects. It verifies that the space around the detected speckle is empty. To do this efficiently, an integral image of the binarized depth map is computed, which allows for rapid, constant-time queries of the pixel sum within any rectangular region. We check the pixel sum in eight rectangular regions surrounding the speckle's bounding box. If the ratio of filled pixels to total pixels within these regions is below a low threshold (\eg, 0.07), the speckle is considered isolated within a glass plane.

    \noindent \textbf{Temporal Consistency:} A final filter operates on a \textbf{tracking-by-detection} principle to ensure identified features are persistent and not transient sensor noise. A speckle is confirmed and passed to the mapping algorithm only after its \texttt{required\_count} (\eg, 1-3 detections) is exceeded across multiple consecutive frames. To prevent the accumulation of false positives and old detections, a \texttt{max\_age} parameter is used to expire and remove tracks that have not been seen for a specified duration.

\subsection{Transparent Plane Reprojection}

The final stage of our methodology involves segmenting empty regions in the depth map and reprojecting the confirmed transparent planes. The algorithm first identifies the empty regions in the depth image and applies a non-maximum suppression (NMS) algorithm to merge redundant empty regions, ensuring a single, accurate representation of each transparent plane. The algorithm then locates the empty region corresponding to a validated speckle location. This entire segmented region is then populated with the speckle's depth value obtained from the sonar-filtered ToF image.
A horizontal depth gradient is applied across the populated region to model the expected tilt of the glass plane. This gradient is proportional to the speckle's measured depth and the speckle's horizontal distance from the image center. This synthesized depth data effectively renders the transparent obstacle visible as a solid surface in the depth map. This fused depth map can then be used by the robot's navigation stack for safe path planning and obstacle avoidance. 

\section{Experimental Results}
\label{sec:experiments}

In this section, we describe the set of experiments performed to validate the method and their results. 
First, we report the results of our speckle detection algorithm on the validation set to show that the model is able to find the speckle in real-world scenarios.

\subsection{Speckle Detection and Glass Segmentation}

\begin{figure*}[!h]
\centering
\begin{tabular}{c c c c c c c}
{\includegraphics[width = 0.12\textwidth]{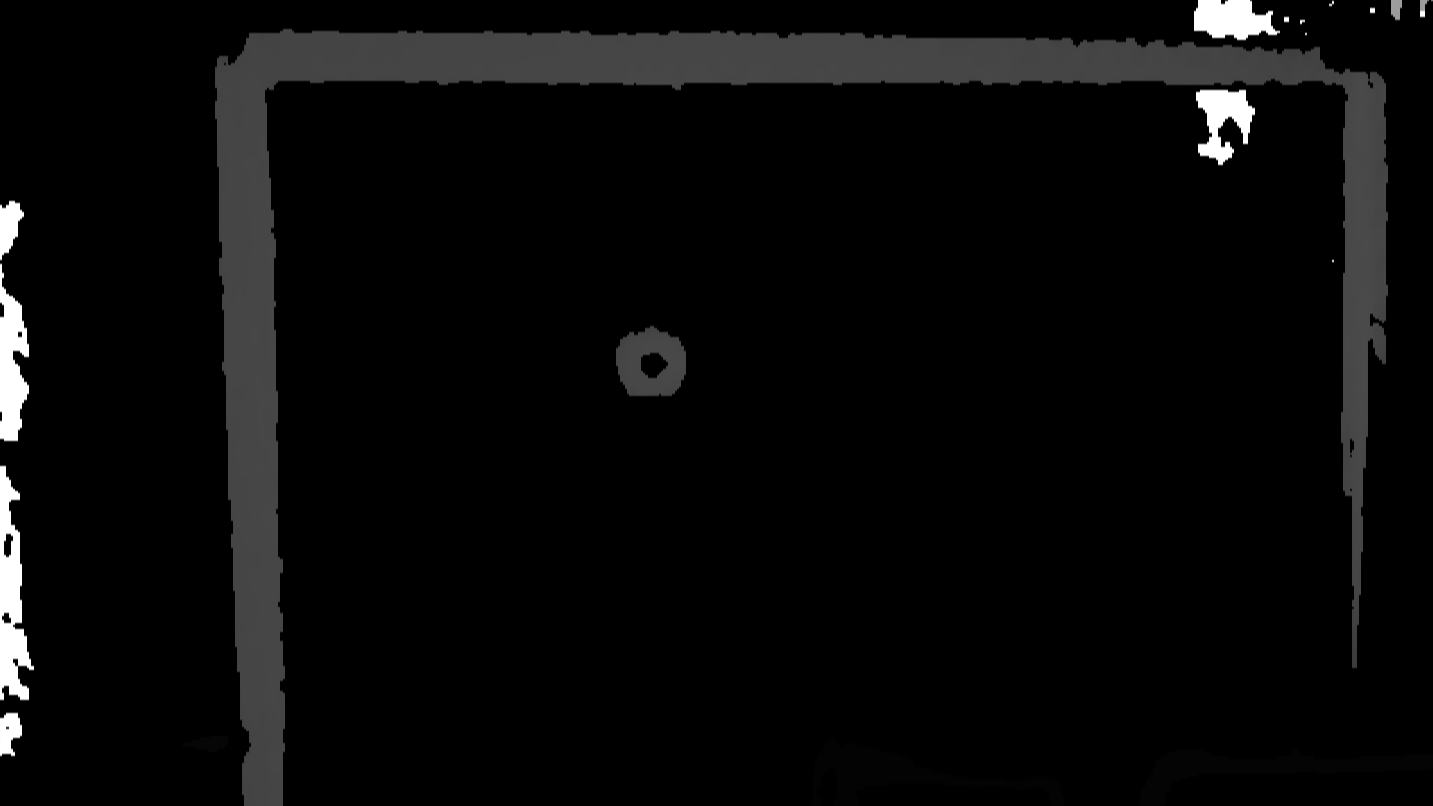}} &
{\includegraphics[width = 0.12\textwidth]{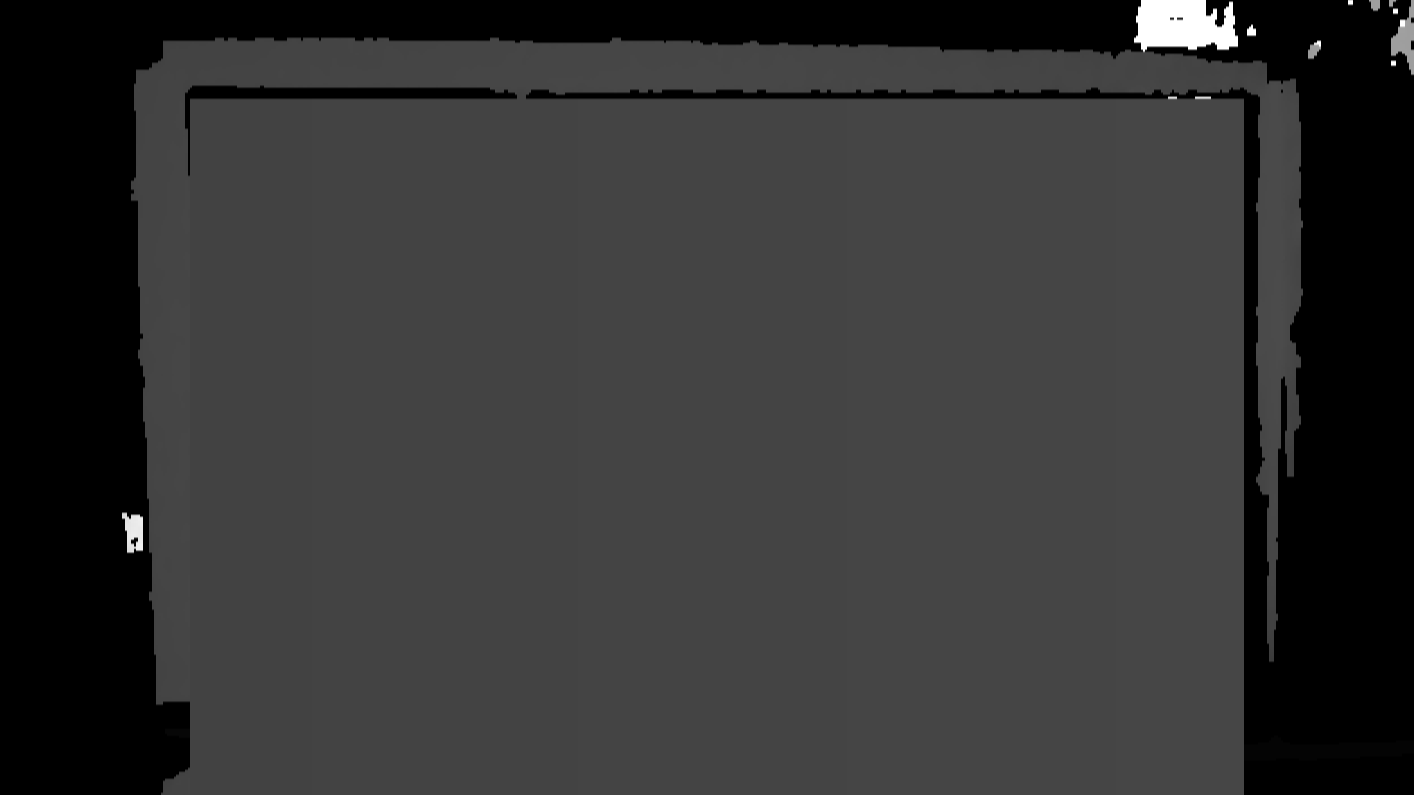}} &
{\includegraphics[width = 0.12\textwidth]{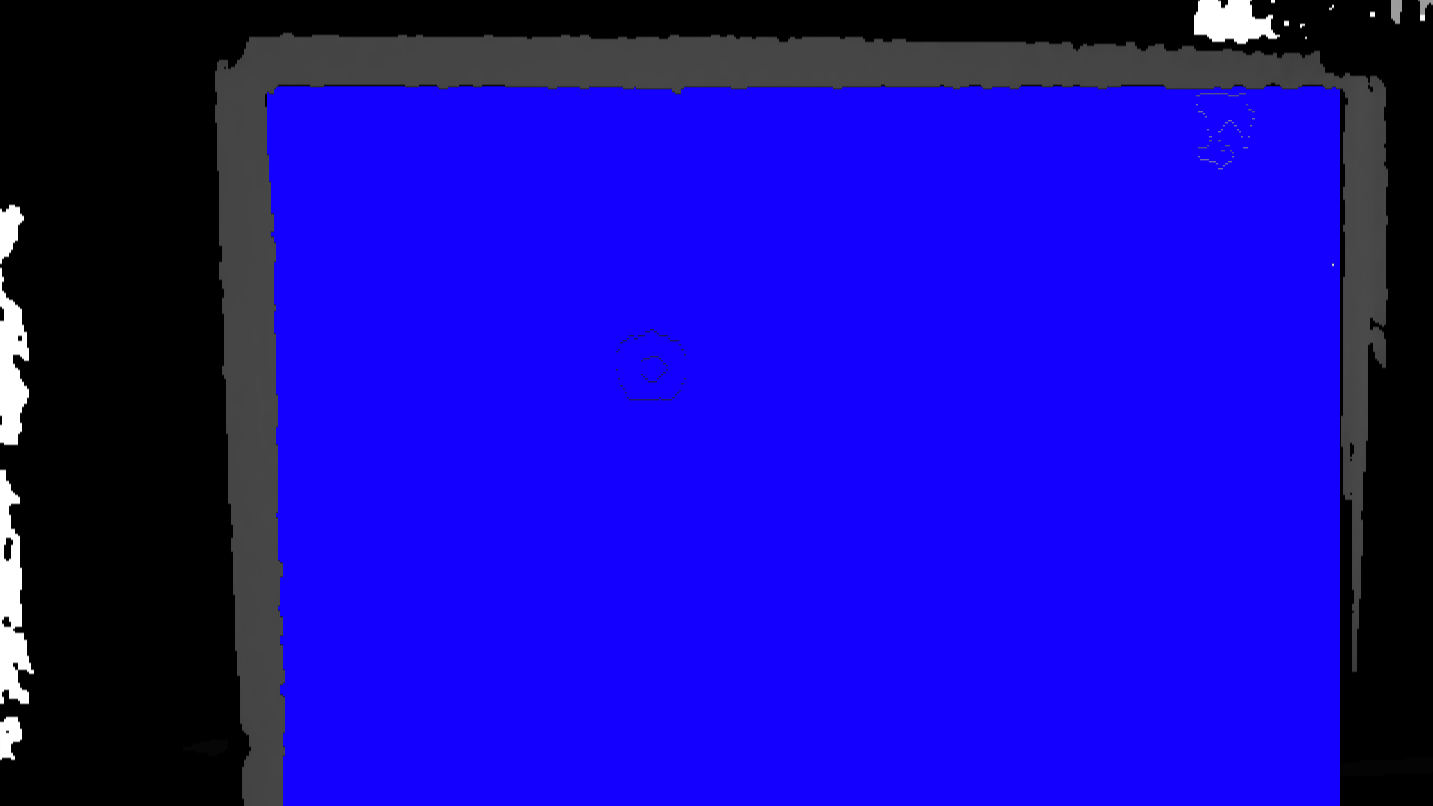}} &
{\includegraphics[width = 0.12\textwidth]{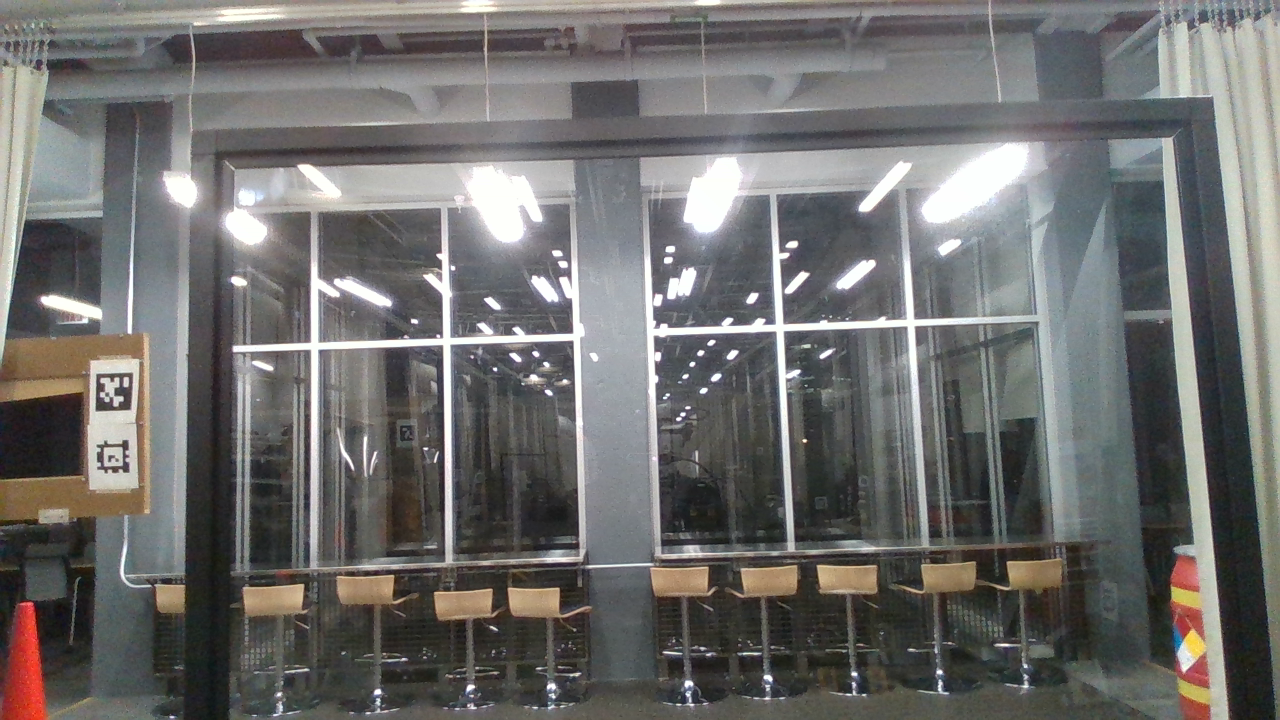}} &
{\includegraphics[width = 0.12\textwidth]{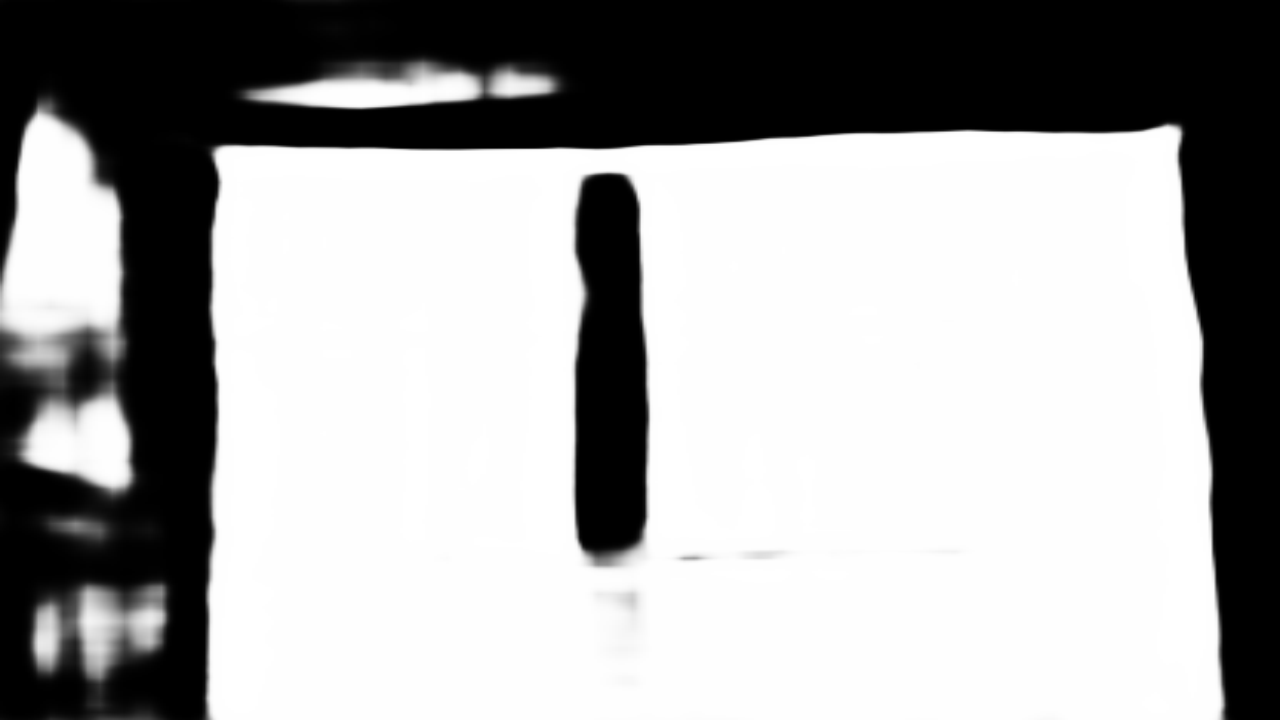}} &
{\includegraphics[width = 0.12\textwidth]{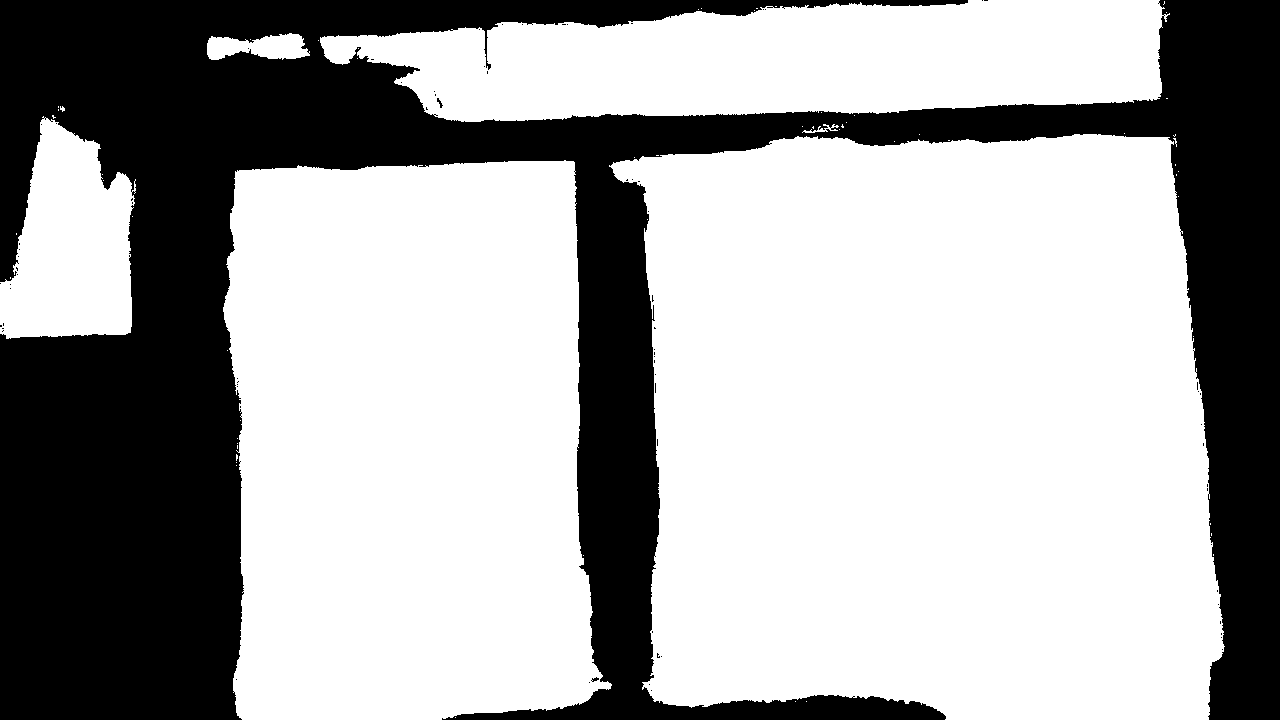}} &
{\includegraphics[width = 0.12\textwidth]{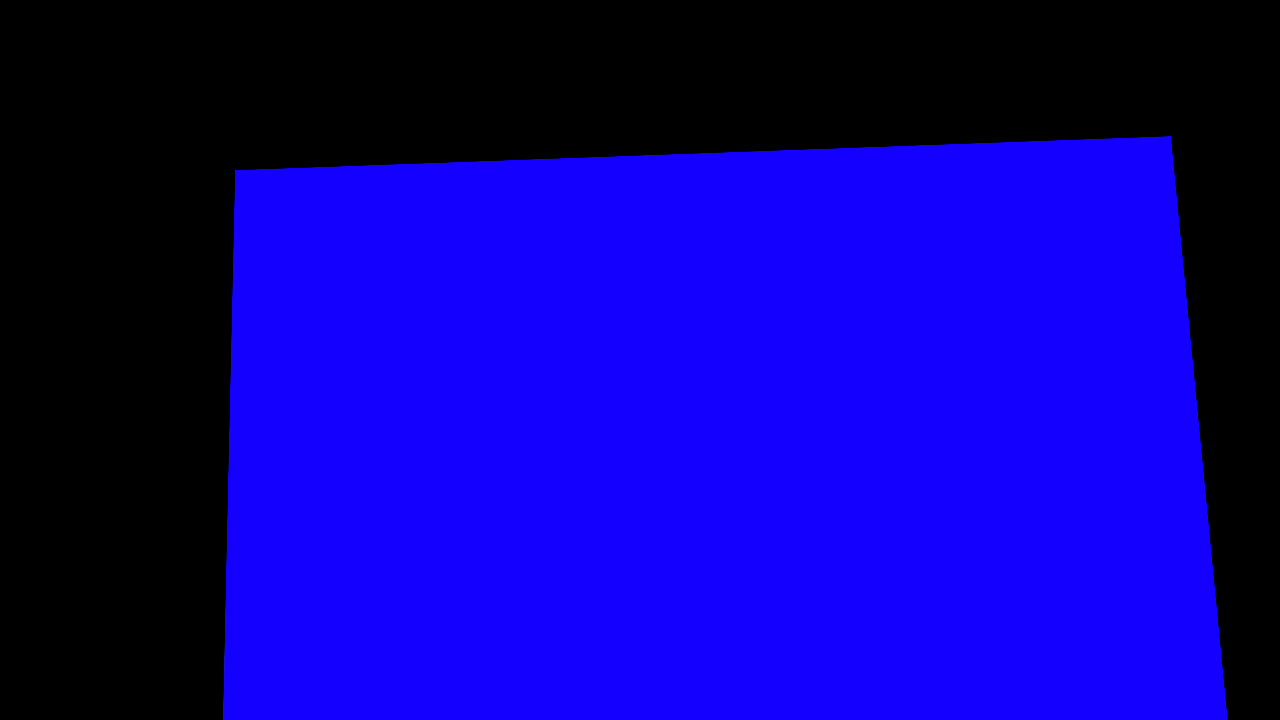}}\\
{\includegraphics[width = 0.12\textwidth]{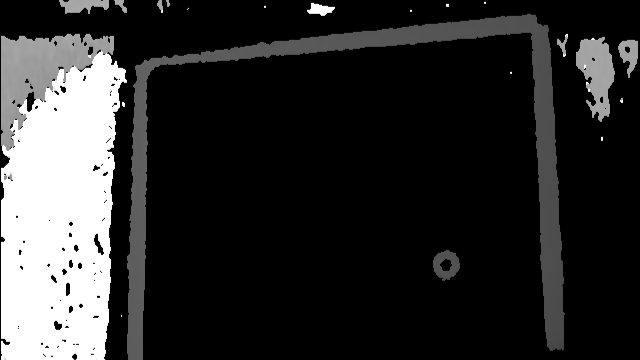}} &
{\includegraphics[width = 0.12\textwidth]{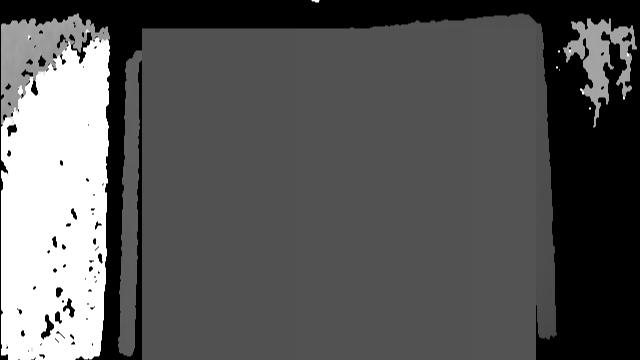}} &
{\includegraphics[width = 0.12\textwidth]{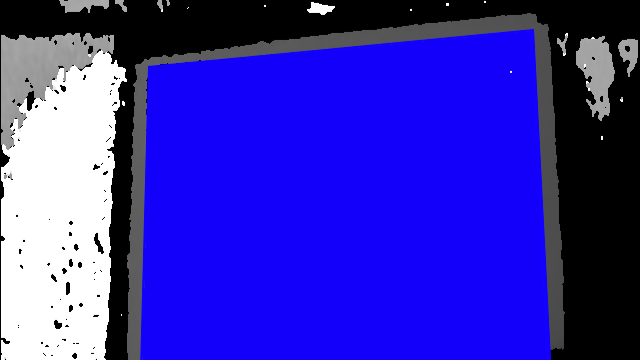}} &
{\includegraphics[width = 0.12\textwidth]{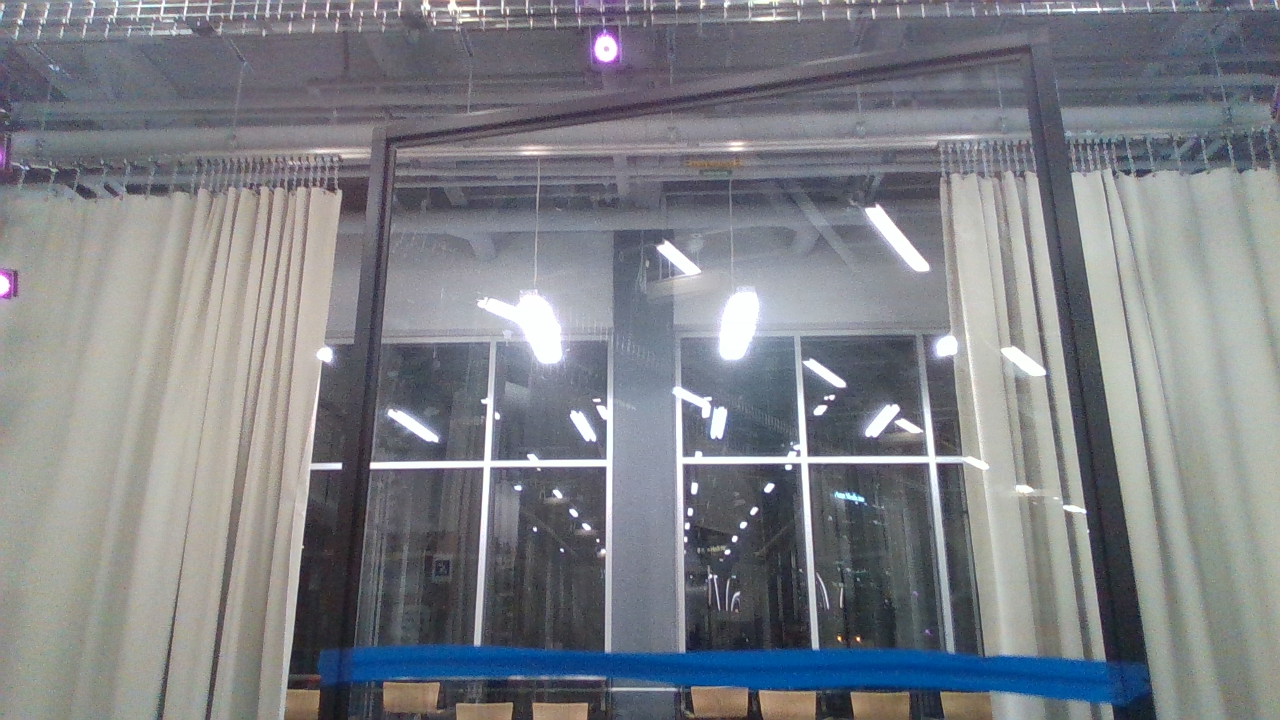}} &
{\includegraphics[width = 0.12\textwidth]{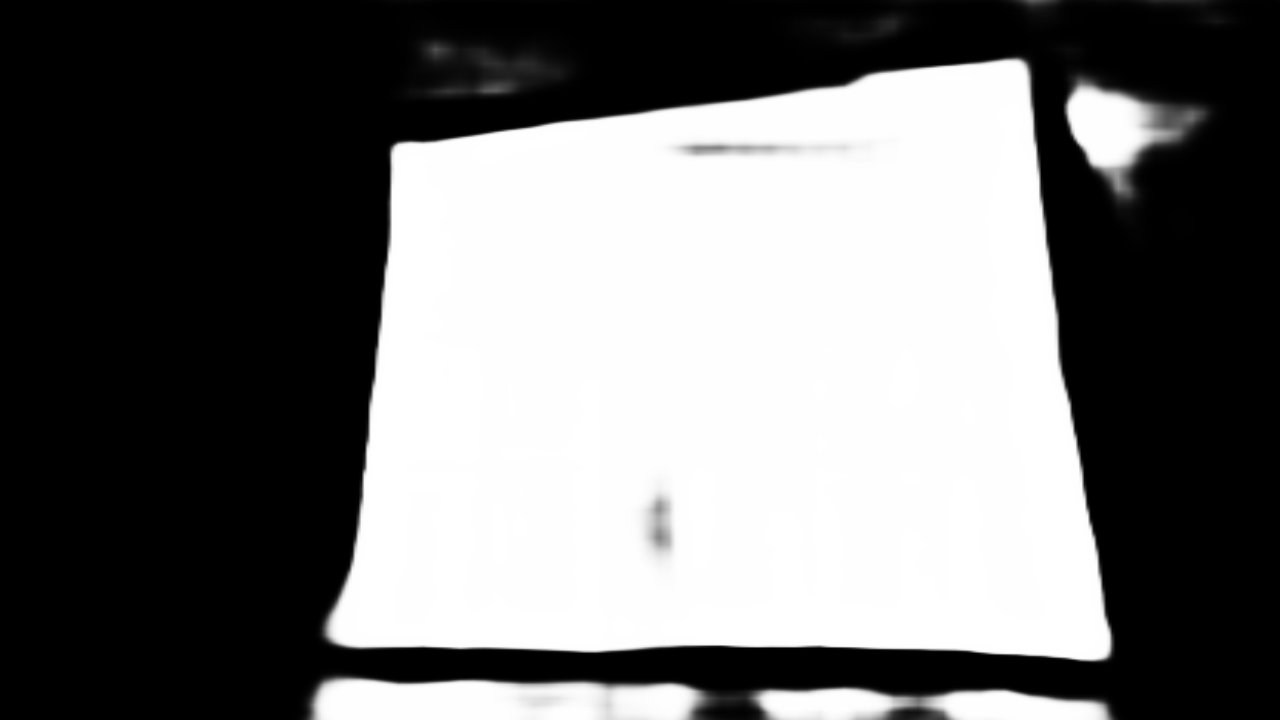}} &
{\includegraphics[width = 0.12\textwidth]{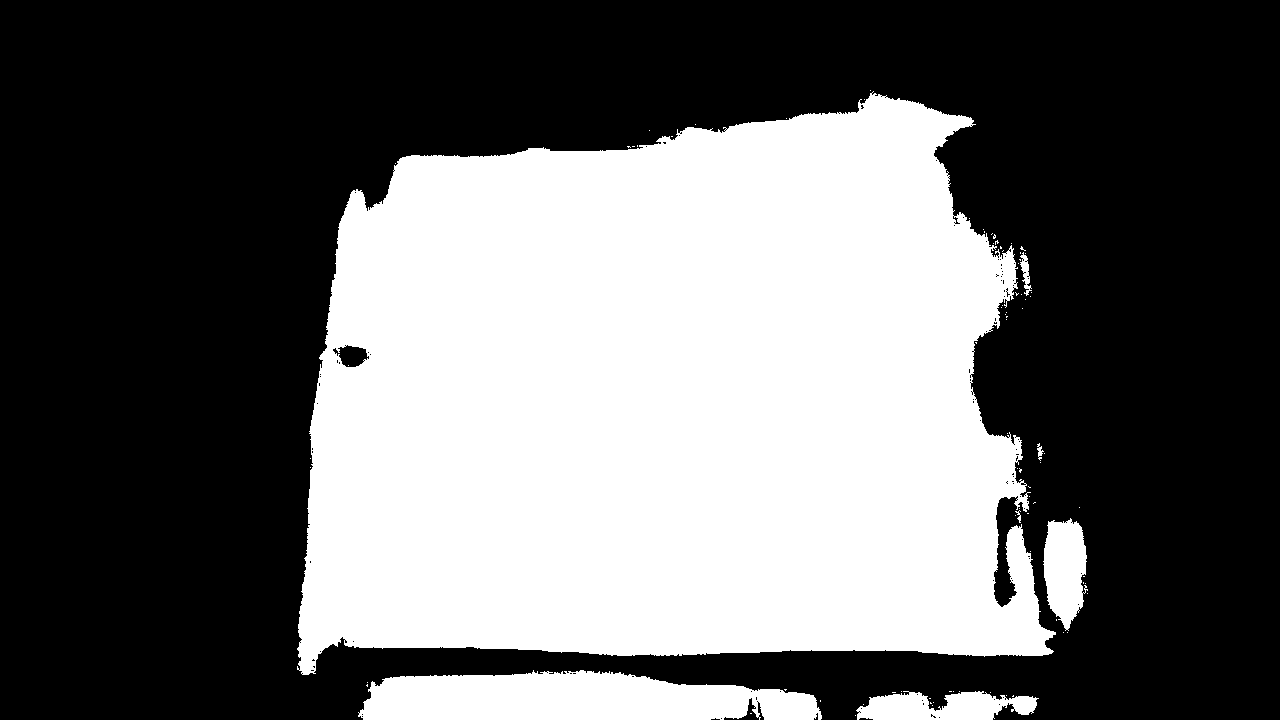}} &
{\includegraphics[width = 0.12\textwidth]{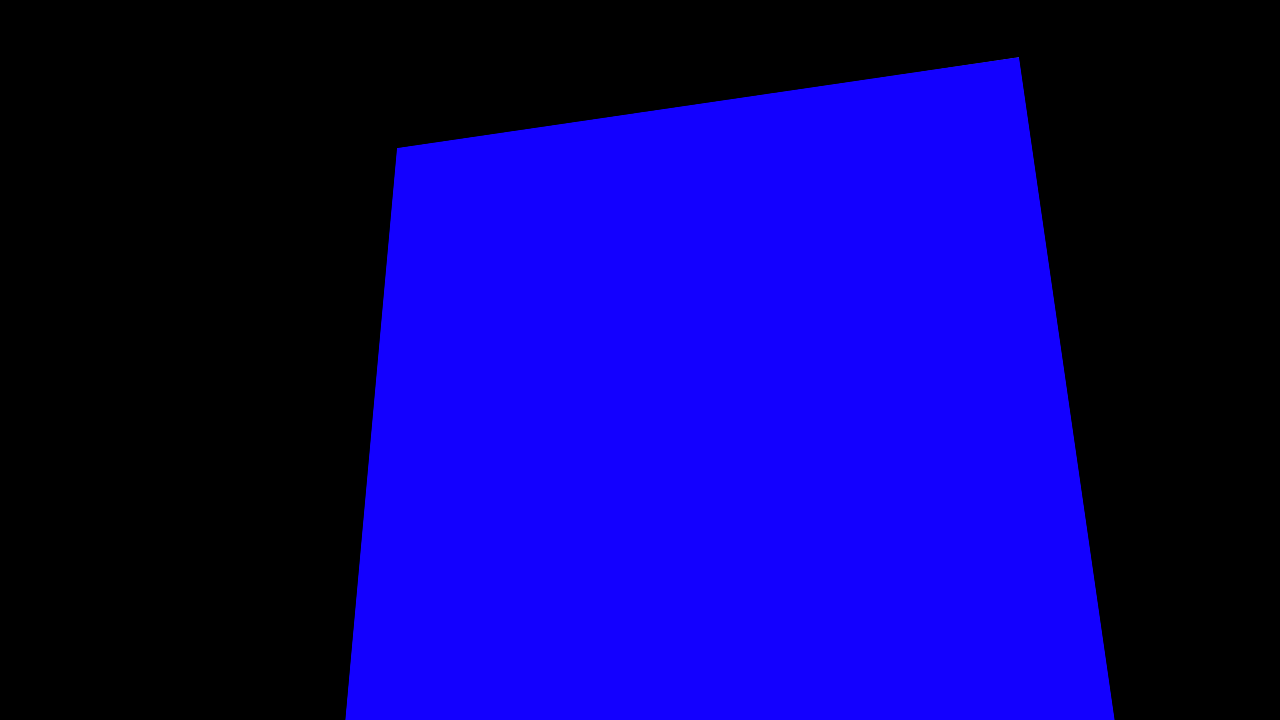}}\\
{\includegraphics[width = 0.12\textwidth]{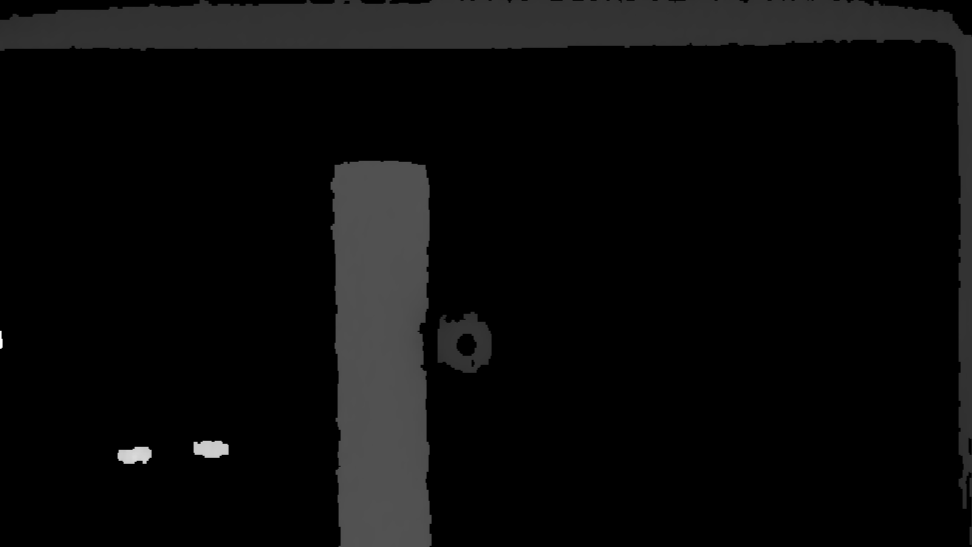}} &
{\includegraphics[width = 0.12\textwidth]{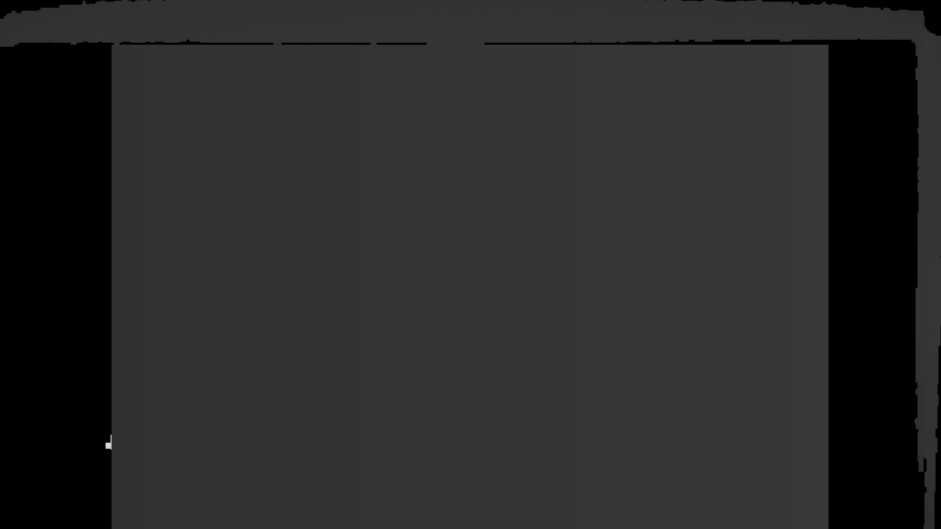}} &
{\includegraphics[width = 0.12\textwidth]{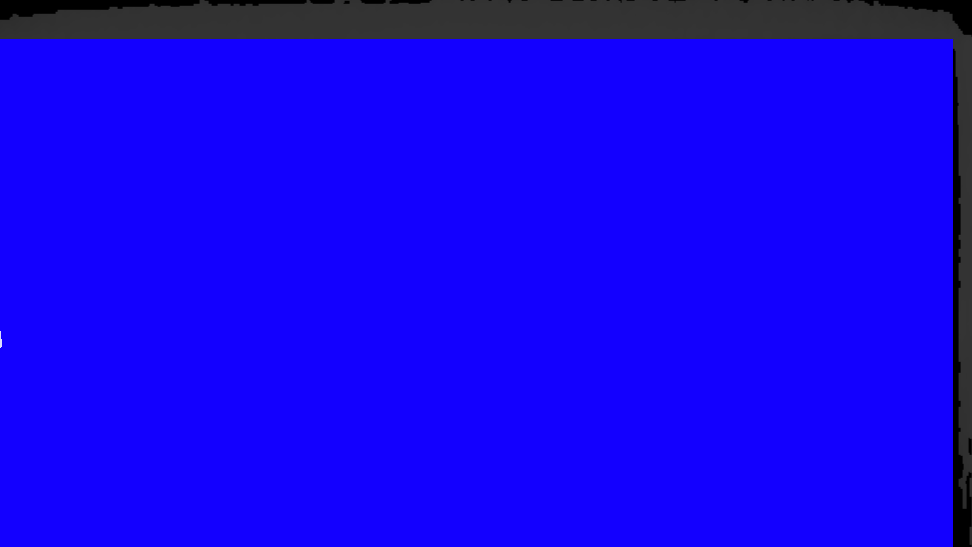}} &
{\includegraphics[width = 0.12\textwidth]{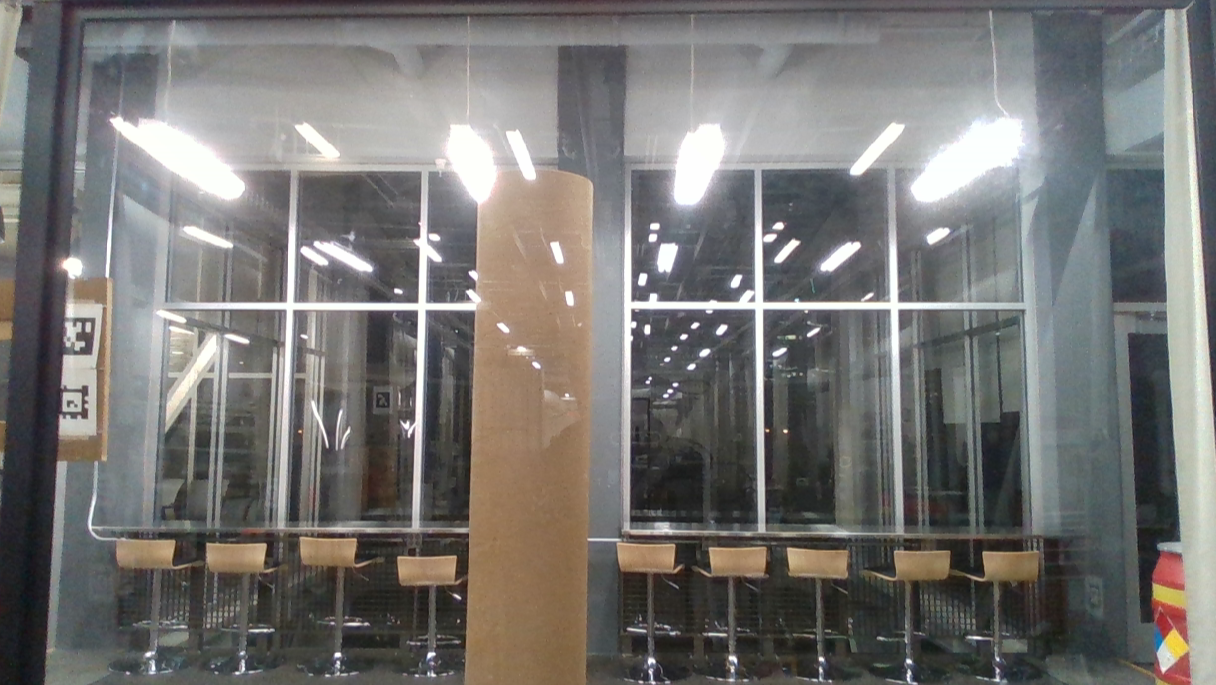}} &
{\includegraphics[width = 0.12\textwidth]{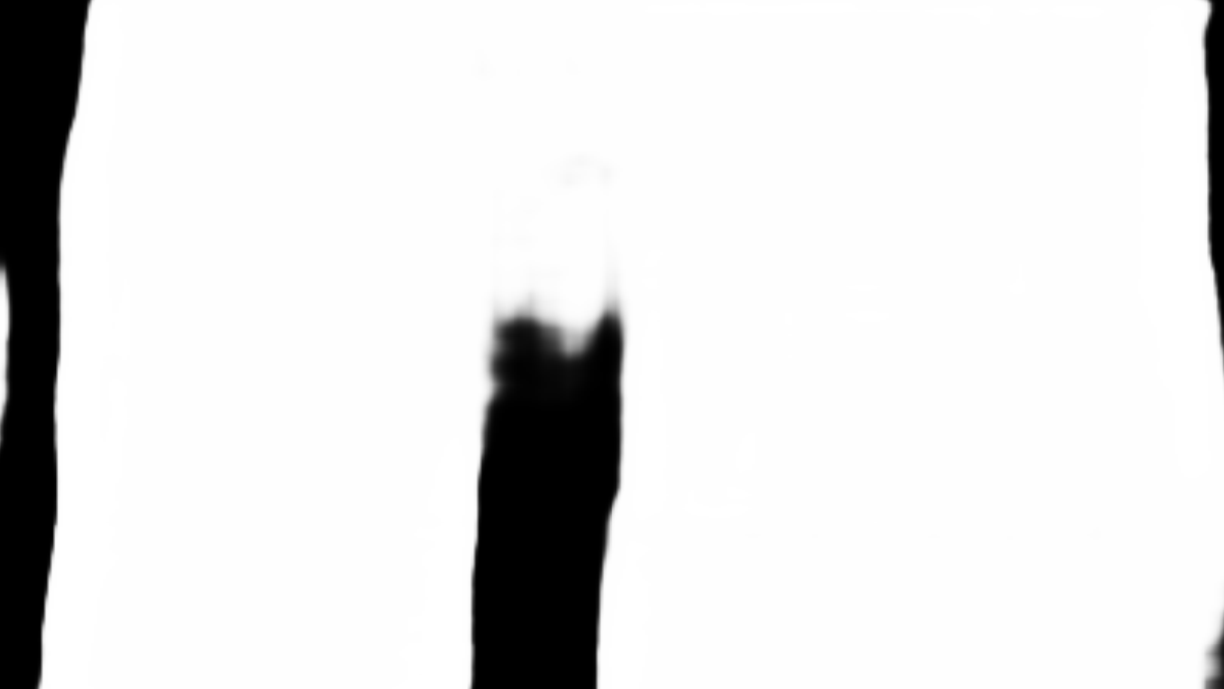}} &
{\includegraphics[width = 0.12\textwidth]{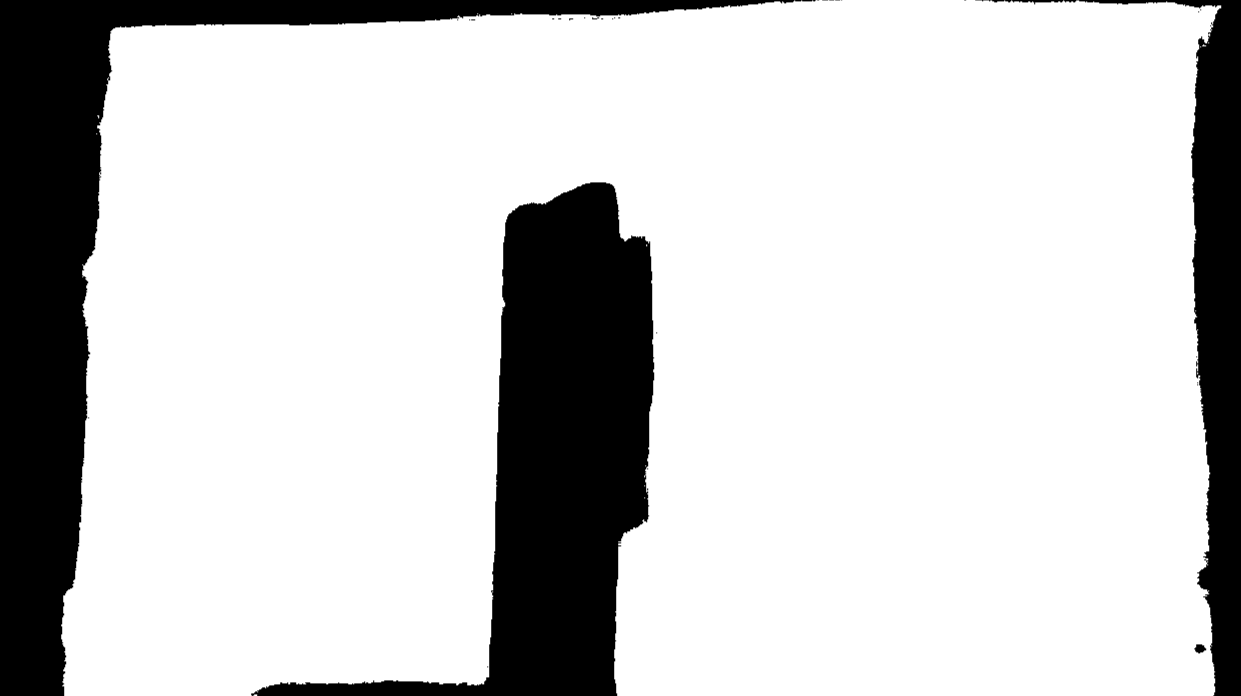}} &
{\includegraphics[width = 0.12\textwidth]{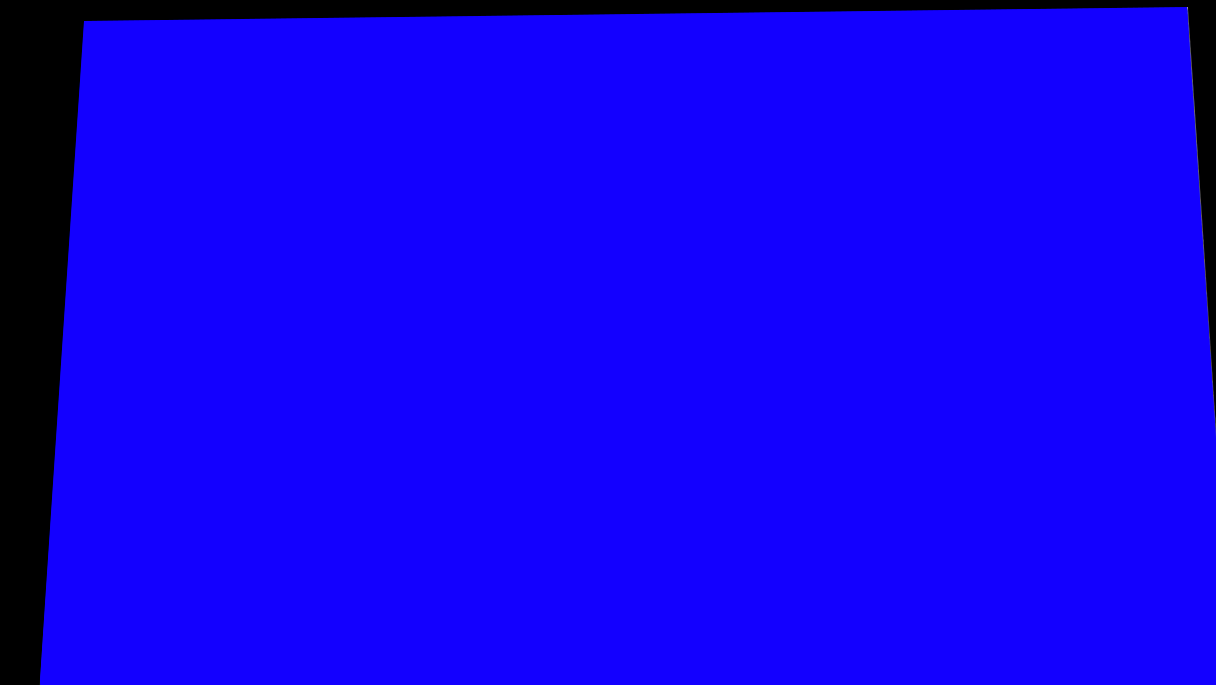}}\\
\subfloat[Baseline]
{\includegraphics[width = 0.12\textwidth]{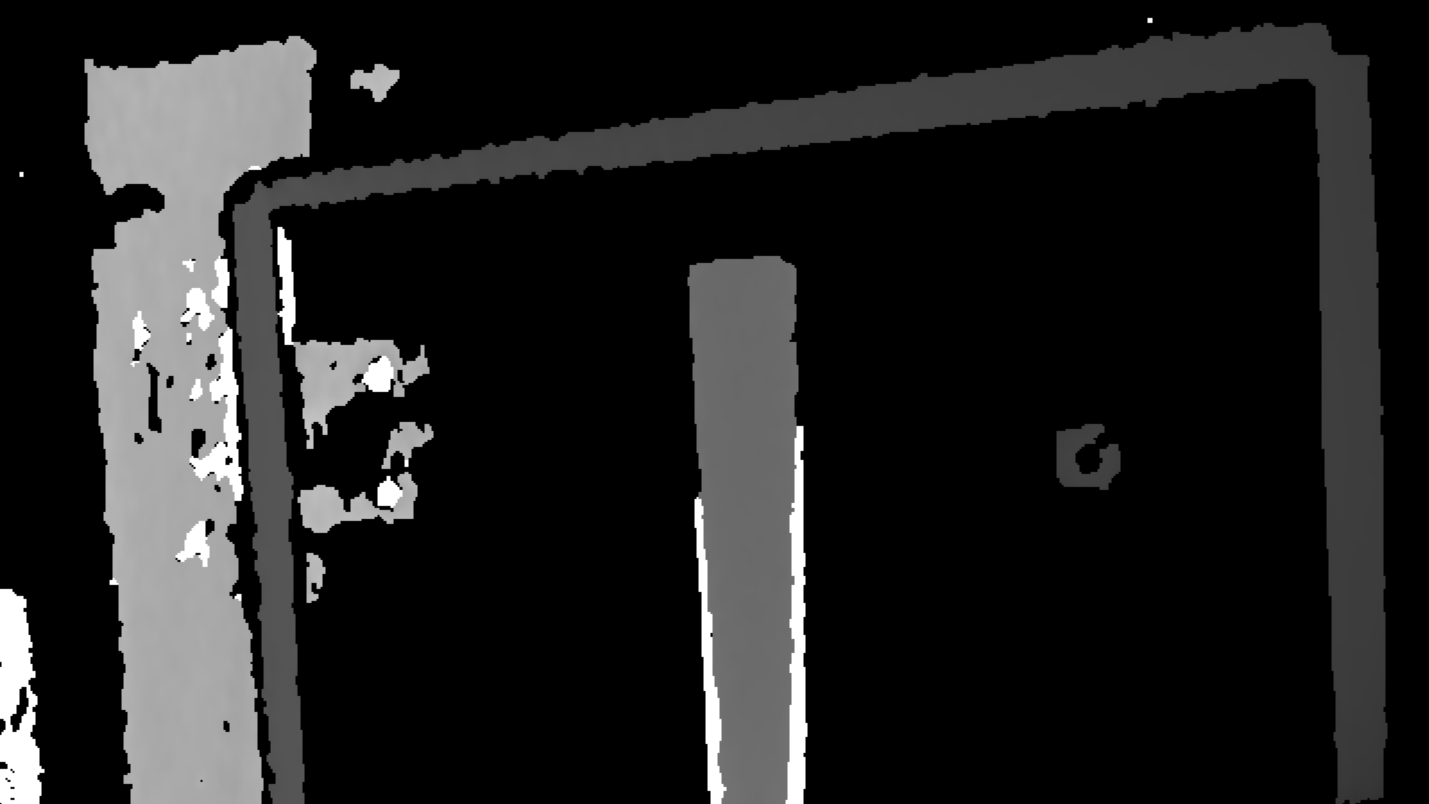}} &
\subfloat[Ours]
{\includegraphics[width = 0.12\textwidth]{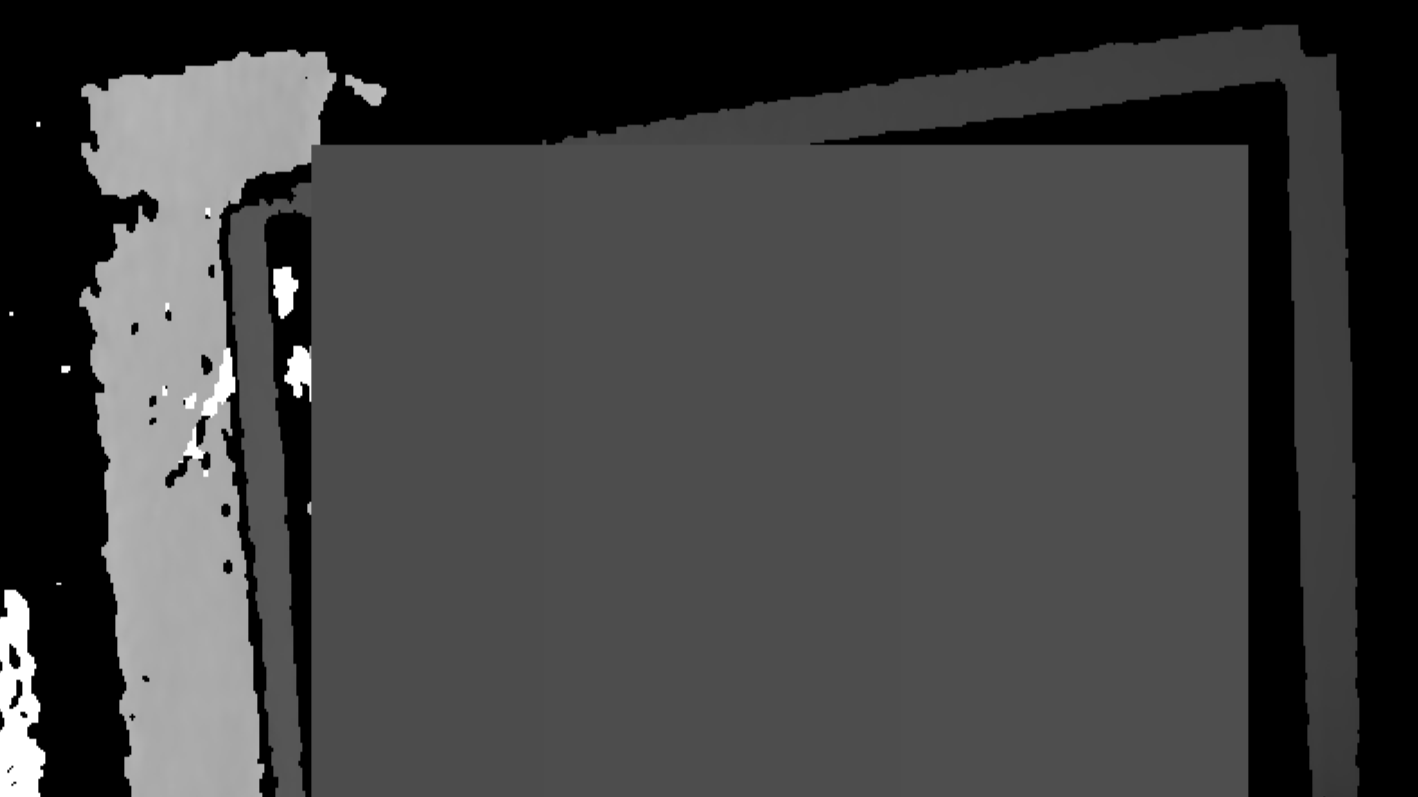}} &
\subfloat[GT]
{\includegraphics[width = 0.12\textwidth]{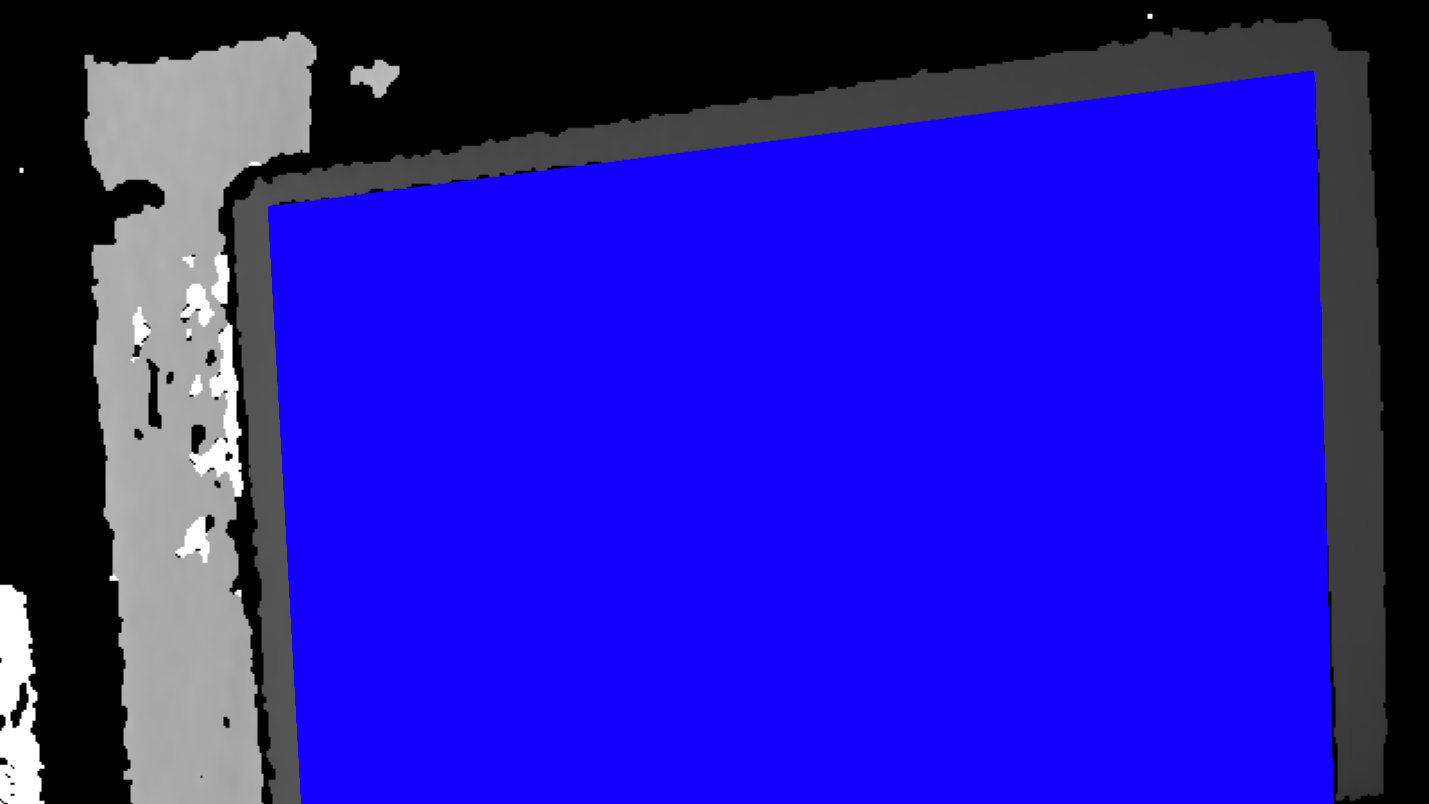}} &
\subfloat[Baseline]
{\includegraphics[width = 0.12\textwidth]{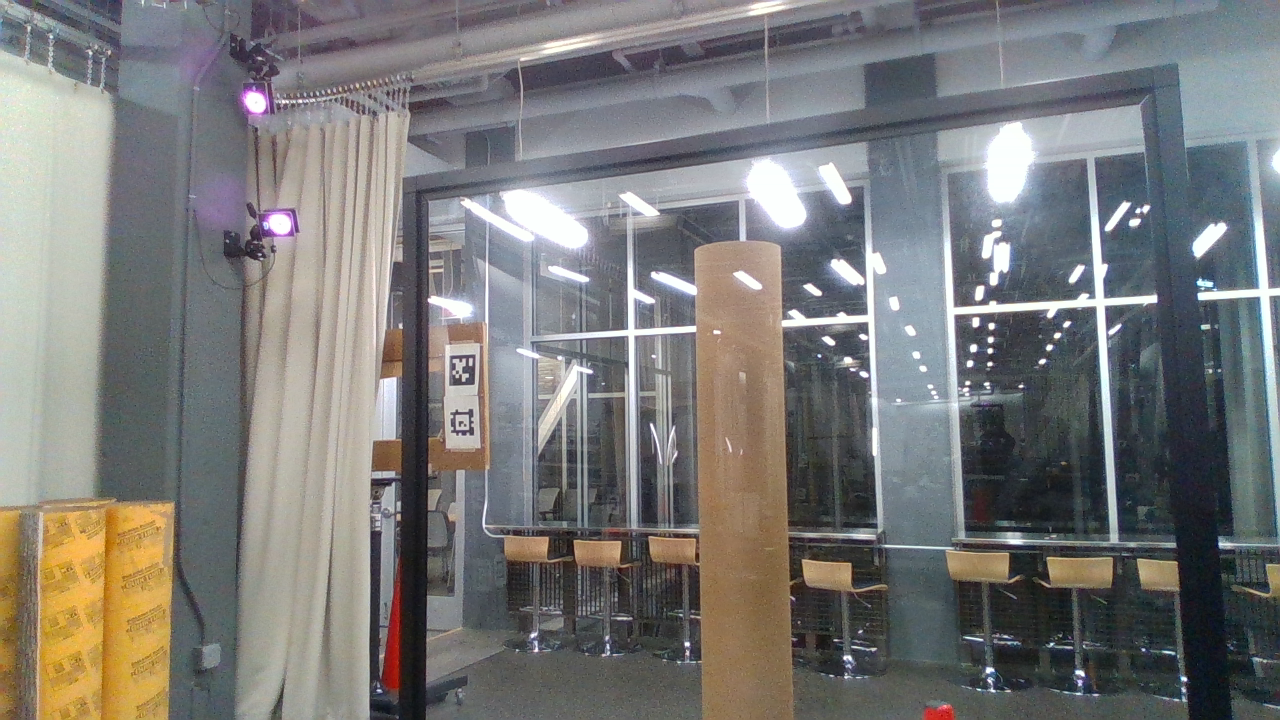}} &
\subfloat[GDNet]
{\includegraphics[width = 0.12\textwidth]{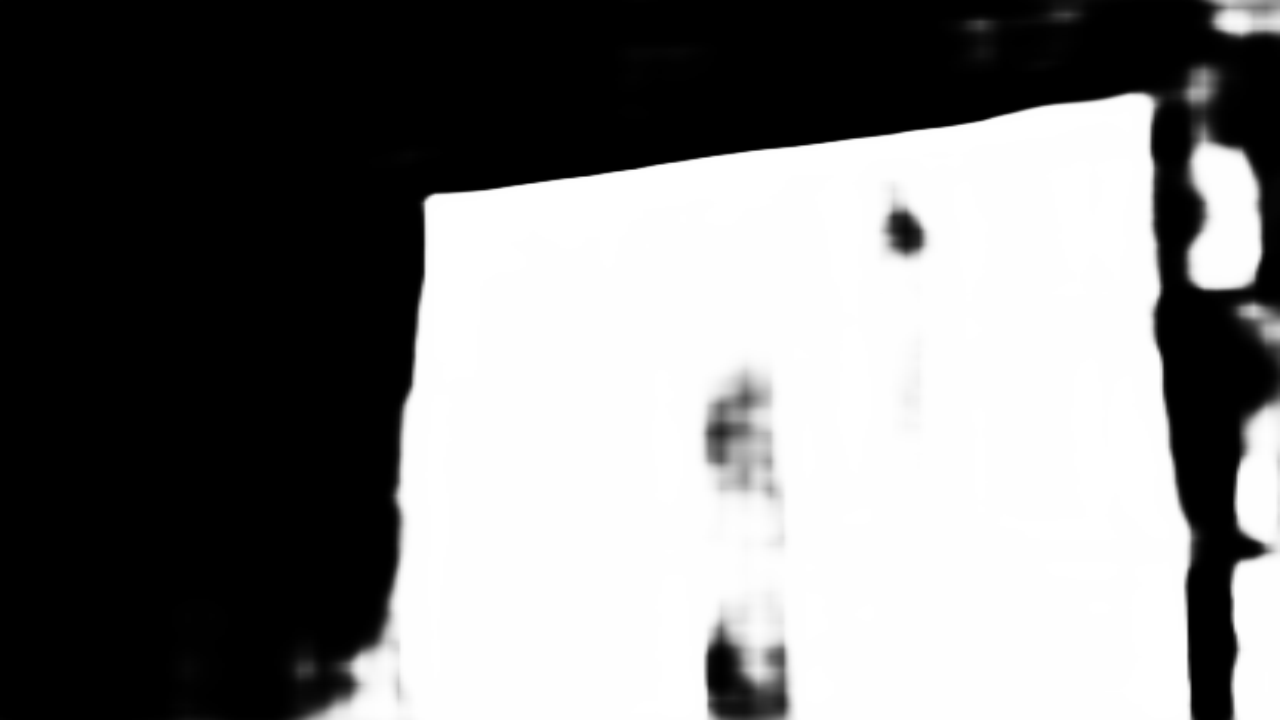}} &
\subfloat[GlassSemNet]
{\includegraphics[width = 0.12\textwidth]{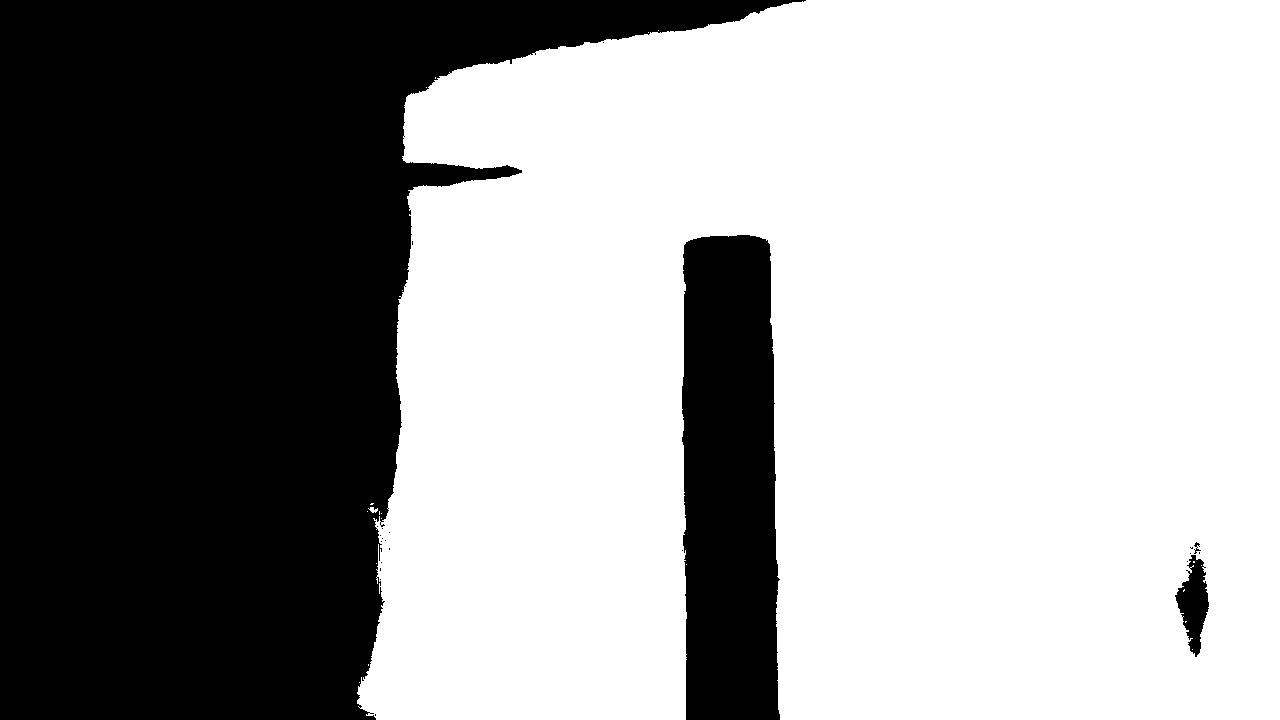}} &
\subfloat[GT]
{\includegraphics[width = 0.12\textwidth]{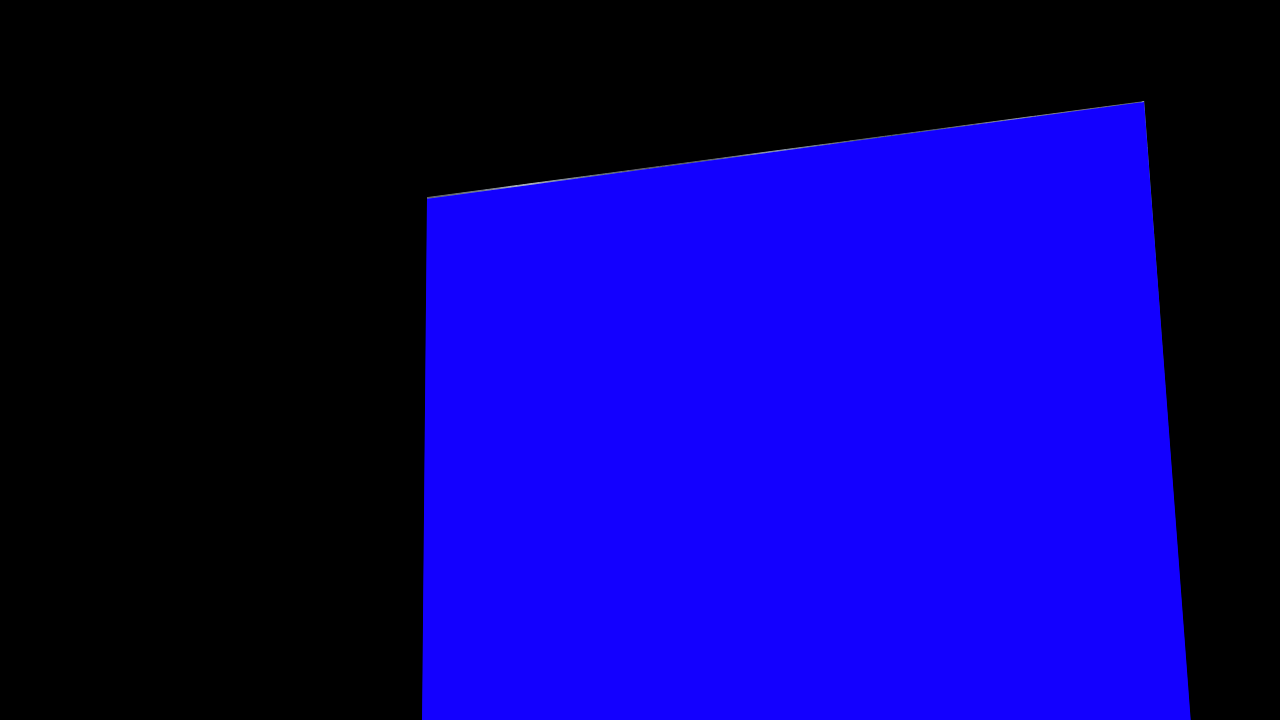}}\\
\end{tabular}
\caption{Qualitative results for Speckle Detection and Glass Segmentation. The figure illustrates the performance of our algorithm (b) in segmenting transparent planes from both clear and cluttered environments. Rows 1 and 2 show results from head-on and angled approaches to a clear glass pane, while rows 3 and 4 show results from similar approaches with background objects. 
\vspace{-1em}
}
\label{fig:mapping_percent}
\end{figure*}

We evaluated the performance of our speckle detection and transparent plane segmentation algorithms through a series of controlled experiments designed to measure their accuracy in detecting a single glass pane while differentiating it from background objects. Table~\ref{tab:performance_summary} displays the quantitative results and Figure~\ref{fig:mapping_percent} displays the qualitative results.

\begin{table}[!h]
    \centering
    \caption{Summary of Speckle Detection and Empty Space Segmentation Performance. 
    }
    \label{tab:performance_summary}
    \begin{tabularx}{0.5\textwidth}{l l l l}
        \toprule
        \textbf{Experiment Type} & \textbf{Precision (\%)} & \textbf{Recall (\%)} &  \textbf{mIOU(\%)}\\
        \midrule
        \textbf{Head-on} (Clear) (Ours) & \textbf{92.3}\% & \textbf{96}\% & \textbf{82.5}\% \\
        Head-on (Clear) (GSN) & 71.8\% & 54.1\% & 42.3\% \\
        Head-on (Clear) (GD) & 62.7\% & 89.7\% & 58.2\% \\
        \hline 
        \textbf{Angled} (Clear) (Ours)& \textbf{85.7}\% & 83.3\% & \textbf{67.6}\% \\
        Angled (Clear) (GSN) & 76.4\% & 64.4\% & 52.9\% \\
        Angled (Clear) (GD) & 68.2\% & \textbf{97.2}\% & 66.7\% \\
        \midrule
        Head-on (Background) & 92.1\% & 79.6\% & 74.7\% \\
        Angled (Background) & 100\% & 41.3\% & 41.8\%  \\
        Head-on (Bg w/o Sonar) & 80.2\% & 65.7\% & 40.4\% \\
        Angled (Bg w/o Sonar) & 100\% & 41.3\% & 36.6\%  \\
        \bottomrule
    \end{tabularx}
    \vspace{-2em}
\end{table}

We conducted two controlled tests with a hand-carried quadrotor to approach a glass pane under different conditions:
(i) \textbf{Head-on approach}: The quadrotor moved directly toward the glass; and (ii) \textbf{Angled approach}: The quadrotor approached the glass at a 5-degree angle.

For each test, we measured the precision, recall, and mIOU metrics. In these experiments, with a clear glass pane, the algorithm performed exceptionally well with a precision of \textbf{92.3\%} and a recall of \textbf{96\%} for head-on approach, and a precison of \textbf{85.7\%} and a recall of \textbf{83.3\%} for angled approach, as shown in Table~\ref{tab:performance_comparison}.
These results demonstrate its fundamental capability to detect transparent surfaces.

We conducted experiments with various objects placed behind the glass pane to evaluate the effectiveness of the sonar sensor in filtering background interference. The algorithm successfully used the sonar sensor data to identify the glass and filter out the objects behind it with a precison of \textbf{92.1\%} and recall of \textbf{79.6\%} for head-on approach, and a precison of \textbf{100\%} and a recall of \textbf{41.3\%} for angled approach, though performance was slightly degraded compared to the clear glass experiments. 
This performance drop at steeper angles is expected, as the algorithm's depth threshold becomes more ambiguous.

To confirm the sonar sensor's impact, we repeated these experiments with the sonar depth filter deactivated. The head-on results were noticeably worse head-on, with a precision of \textbf{80.2\%} and a recall of \textbf{65.7\%} for head-on approach, a precision of \textbf{100\%} and a recall of \textbf{41.3\%} for angled approach. This confirms that the sonar is a reliable way to filter background objects. However, since the sensor is fixed and provides only a 1D measurement, its reliability degrades as the angle of incidence increases. This issue could be mitigated by either decreasing the predefined sonar depth filtering threshold or by making the threshold proportional to the horizontal translation of the speckle, which would make the filter more robust to angled glass panes. 
We also note that in the angled tests, the speckle's location was not always directly in front of a background object.

In addition to speckle detection, we evaluated our transparent plane segmentation algorithm by comparing its output to ground truth glass location and depth data. The quantitative results demonstrate the algorithm's performance across various scenarios.

    \noindent \textbf{Clear Experiments:} The algorithm performed well in clear environments, achieving an mIOU of \textbf{82.5\%} for head-on approaches and \textbf{67.6\%} for angled approaches.
    
    \noindent \textbf{Background Experiments:} When background objects were present, the mIOU decreased to \textbf{74.7\%} for head-on and \textbf{41.8\%} for angled approaches indicating a performance drop when faced with complex backgrounds and viewing angles.
    
    \noindent \textbf{Without Sonar Filter:} Experiments conducted without the ultrasonic sensor's depth filter showed a significant performance degradation, with a head-on mIOU of \textbf{40.4\%} and an angled mIOU of \textbf{36.6\%}.

These results indicate that while our system can segment transparent surfaces, its performance is most reliable in clear, head-on scenarios and decreases with complex backgrounds and viewing angles. We also note that the robot's sonar mount physically obstructs a portion of the lower-right corner of the image. This obstruction impedes the precision of the segmentation algorithm, as it may cause the algorithm to segment a larger area than intended when it cannot close figures with corners in that region.

We compare our method to existing methods, GDNet and GlassSemNet. Our approach achieves superior precision (\textbf{92.3\%}), recall (\textbf{96\%}) and mIOU (\textbf{82.5\%}) in the head-on approach and achieve superior precision (\textbf{85.7\%}) and mIOU (\textbf{67.6\%}) in the angled approach. While GDNet and GlassSemNet struggled identifying the singular glass pane among the opaque and trasparent objects behind it, GDNet accomplished a superior recall score of \textbf{97.2\%} in the angled approach. Our algorithm was the only one to successfully exclude background objects from the glass segmentation in every experiment. 

\subsection{Mapping Experiments}
Our experimental results validate the efficacy of our algorithm. We conducted a series of experiments, progressing from controlled, isolated scenarios to complex, real-world environments, with computation performed onboard the robot. Table~\ref{tab:experimental_results} displays our quantitative results for our algorithm's speckle detection and glass segmentation. The qualitative results in Figure 6 visually demonstrate the mapping completeness of our algorithm.

    
\begin{table}[!ht]
    \centering
    \caption{Summary of Experimental Results for Glass Detection and Mapping}
    \label{tab:experimental_results}
    \begin{tabularx}{0.48\textwidth}{l l c c c}
        \toprule
        \textbf{Experiment Type} & \textbf{Environment} & \textbf{Precision} &
        \textbf{Recall} &
        \textbf{mIOU} \\
        \midrule
        \multirow{2}{*}{\shortstack[l]{\\ \\ \\ \\ Controlled\\Experiments}} 
        & Vicon Room & 87.6\% & 85.7\% & 66.8\% \\& (Handheld) \\
        \cmidrule{2-5}
        & Vicon Room & 92.9\% & 72.3\% & 64.4\%\\& (Autonomous) \\
        \midrule
        \multirow{3}{*}{\shortstack[l]{\\ \\ \\ \\ \\ \\ \\ \\ Real-World\\Experiments}}
        & Half-Glass & 96.8\% & 78.9\% & 69.6\% \\& Room \\
        \cmidrule{2-5}
        & Glass & 82.6\% & 62\% & 55.7\%\\& Hallway \\
        \cmidrule{2-5}
        & Glass & 91.2\% & 50.4\% & 48.4\%\\& Perimeter \\
        \bottomrule
    \end{tabularx}
\end{table}

\subsection{Controlled Laboratory Experiments}

We first established a performance baseline using a series of controlled experiments to isolate and evaluate key components of our algorithm.

\begin{figure}[!h]
\centering
\begin{tabular}{c c c}
{\includegraphics[width = 0.28\linewidth]{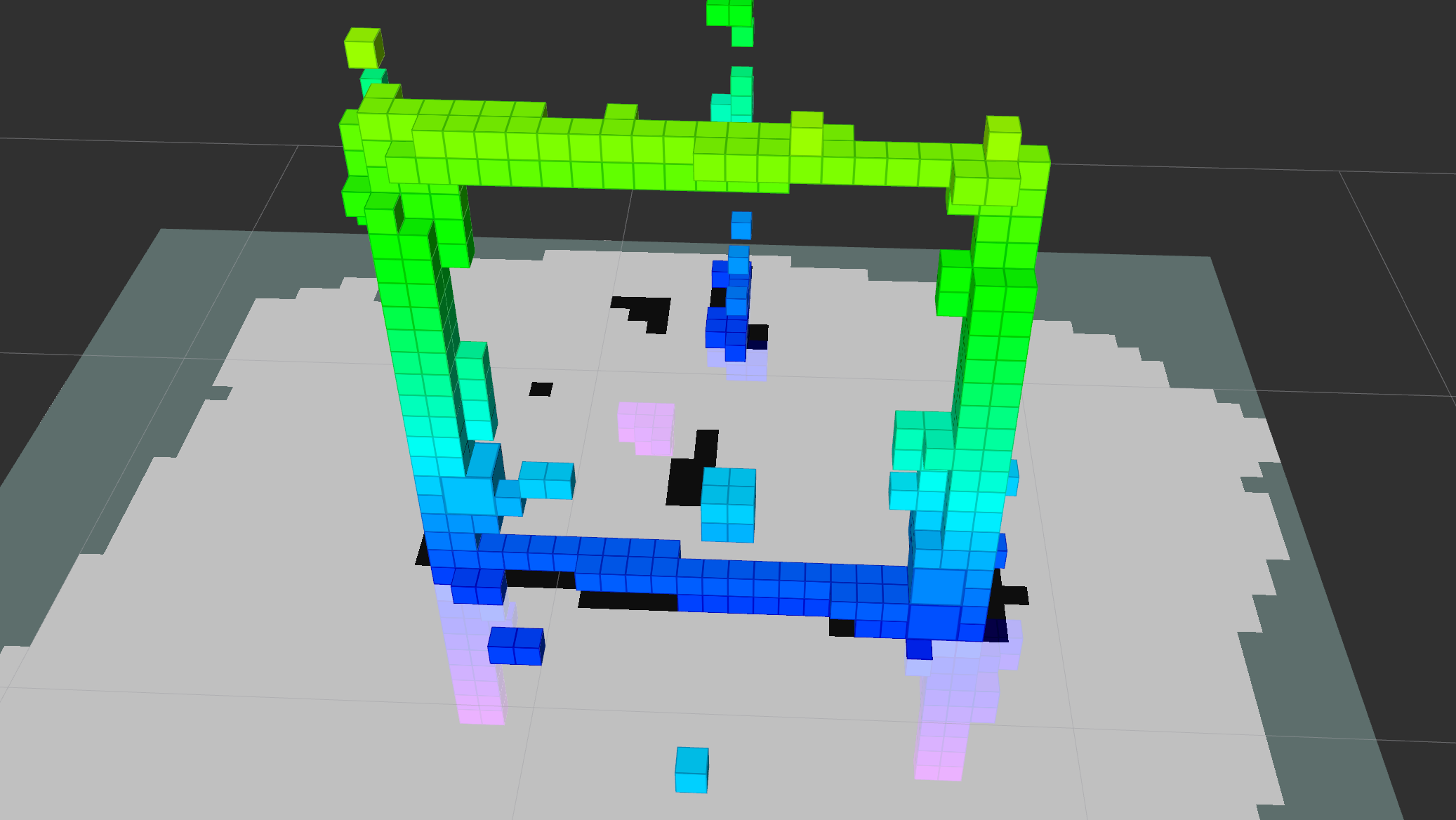}} &
{\includegraphics[width = 0.28\linewidth]{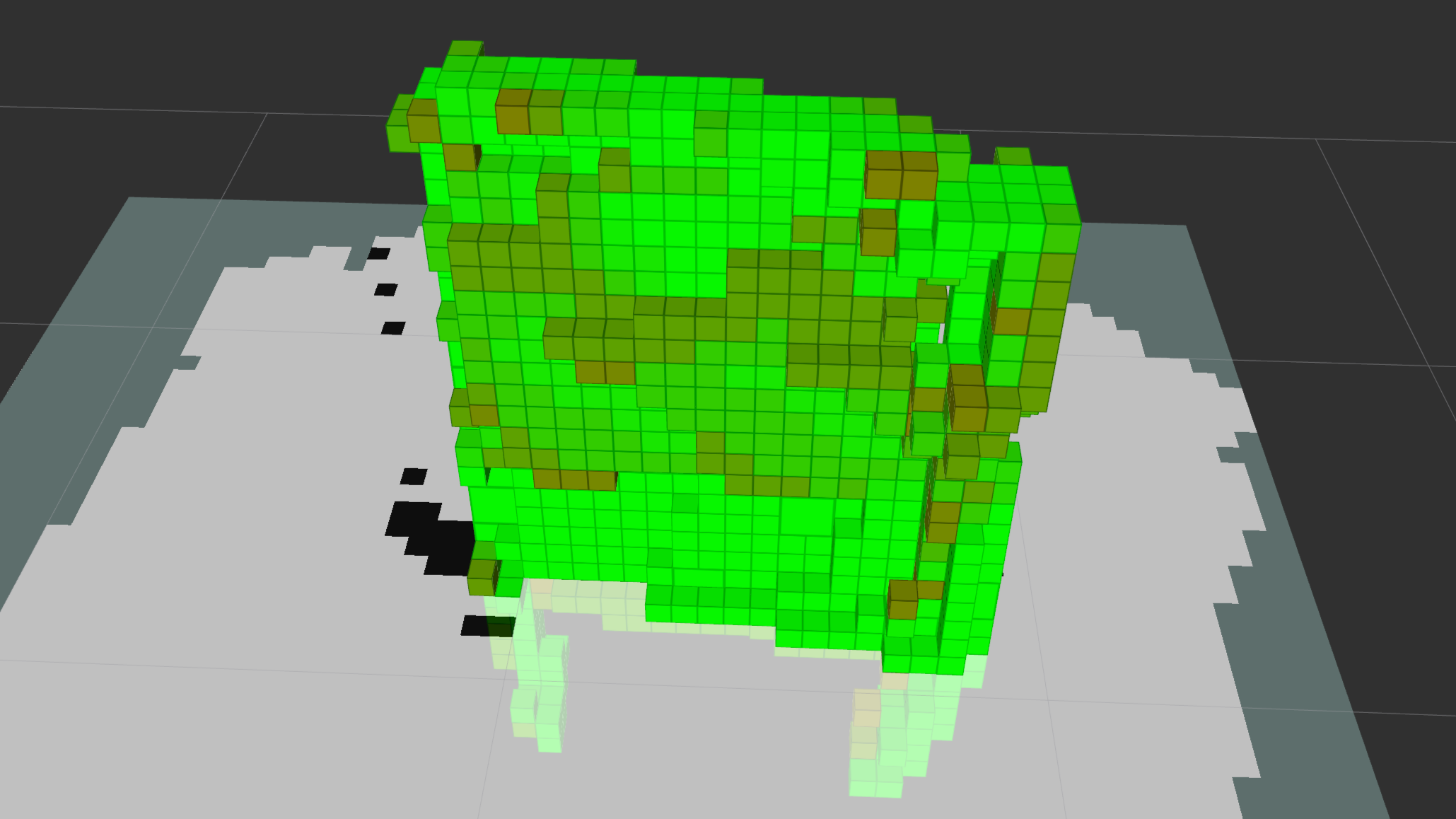}} &
{\includegraphics[width = 0.28\linewidth]{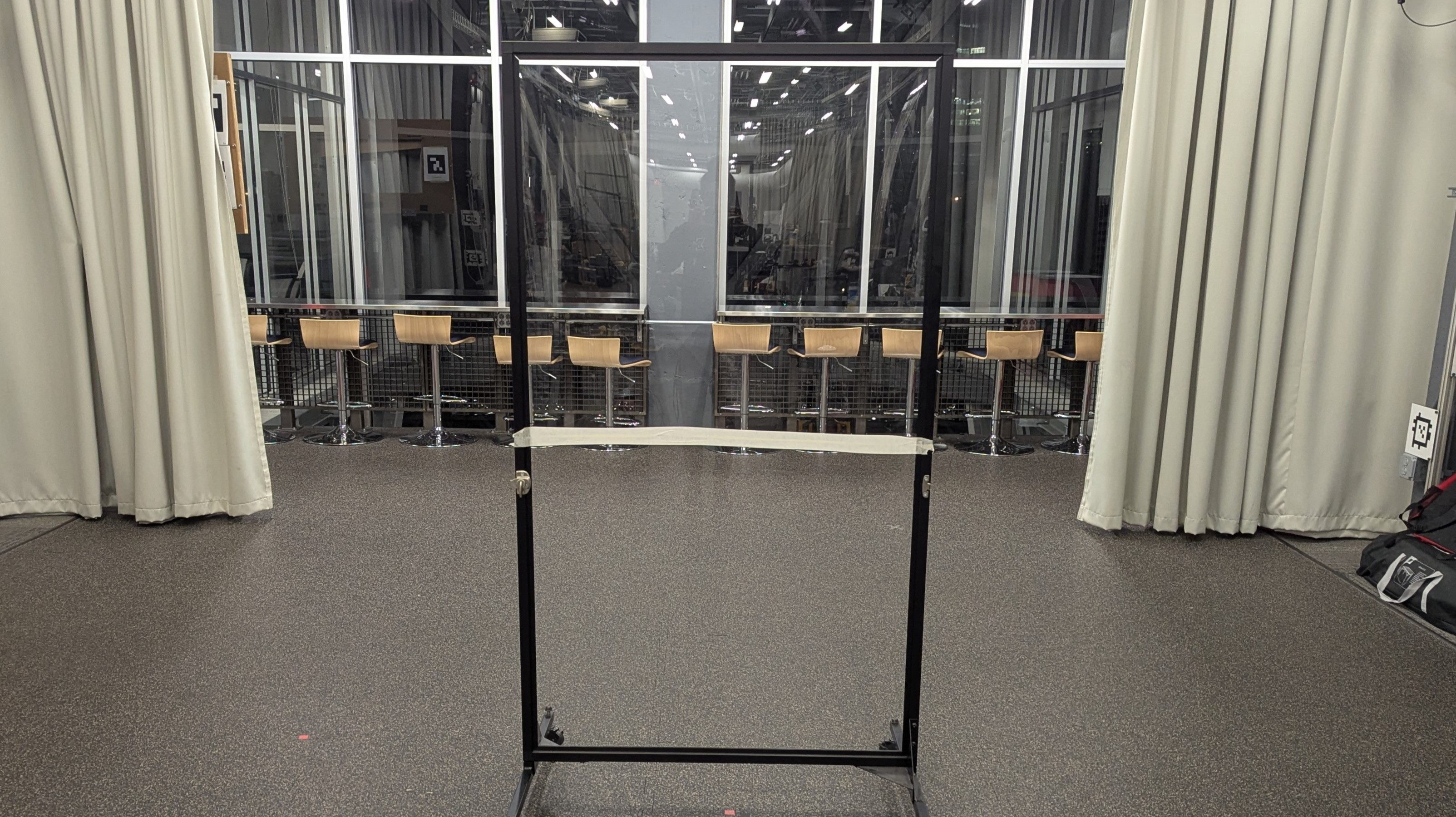}} \\
{\includegraphics[width = 0.28\linewidth]{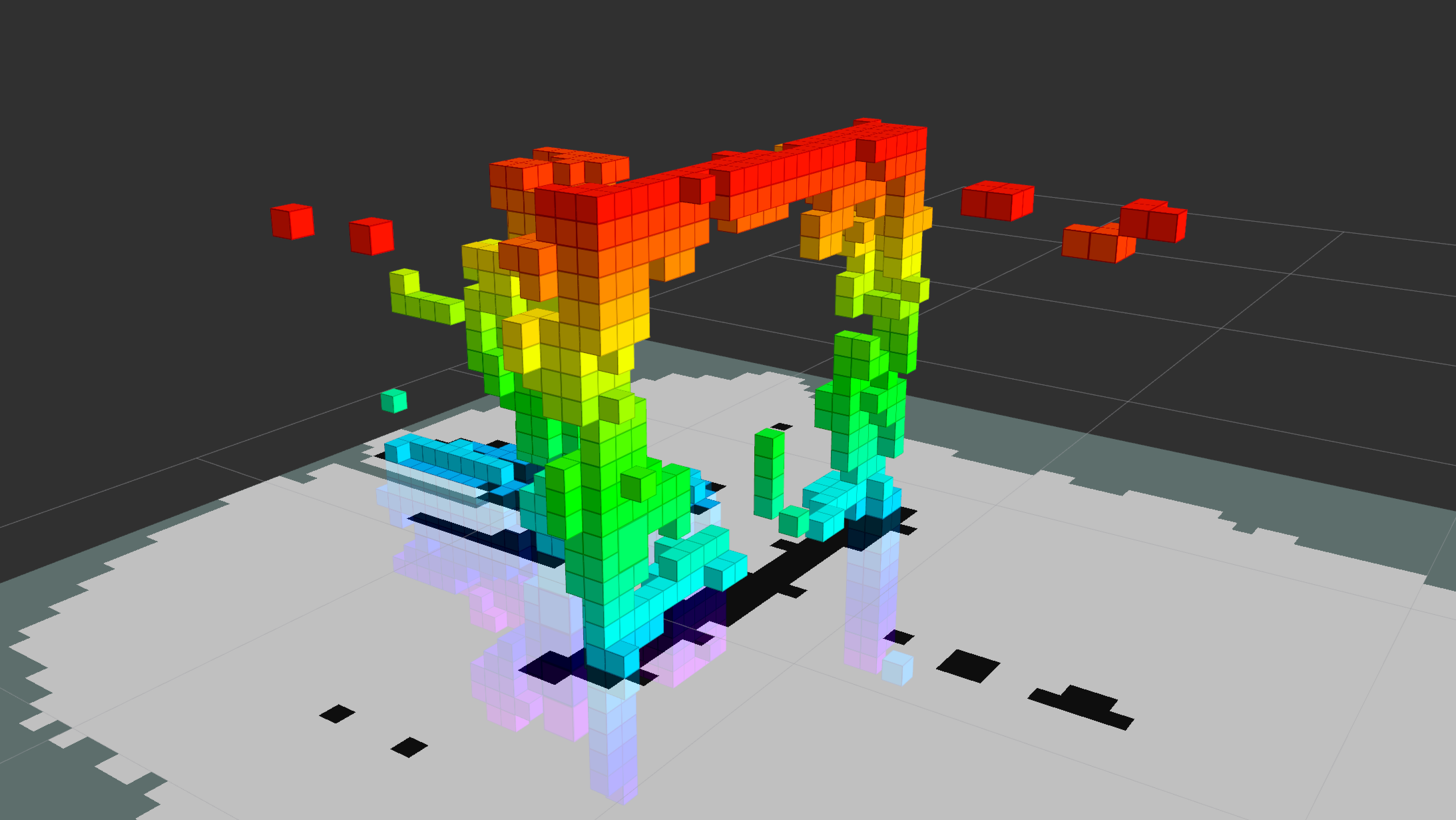}} &
{\includegraphics[width = 0.28\linewidth]{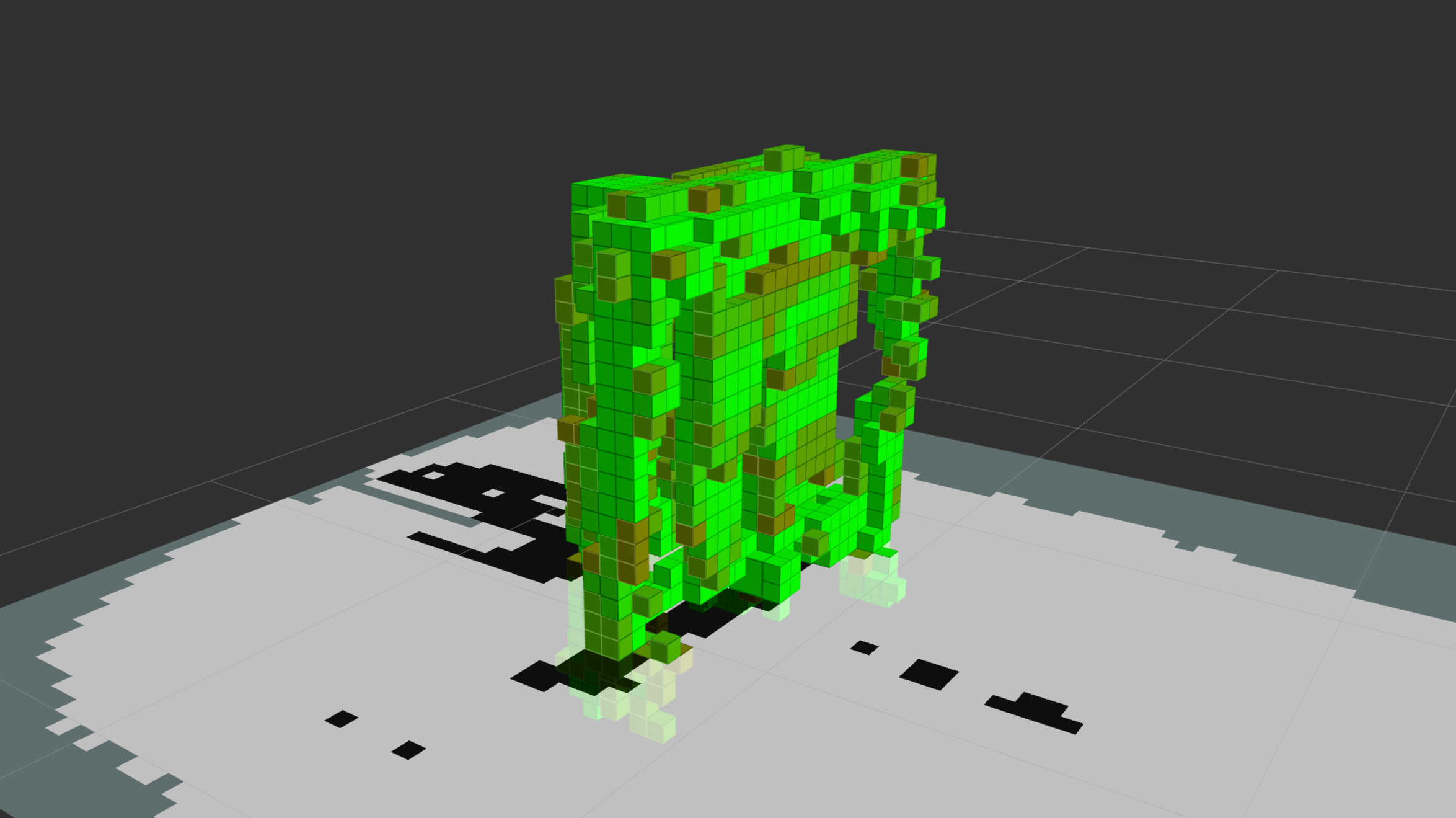}} &
{\includegraphics[width = 0.28\linewidth]{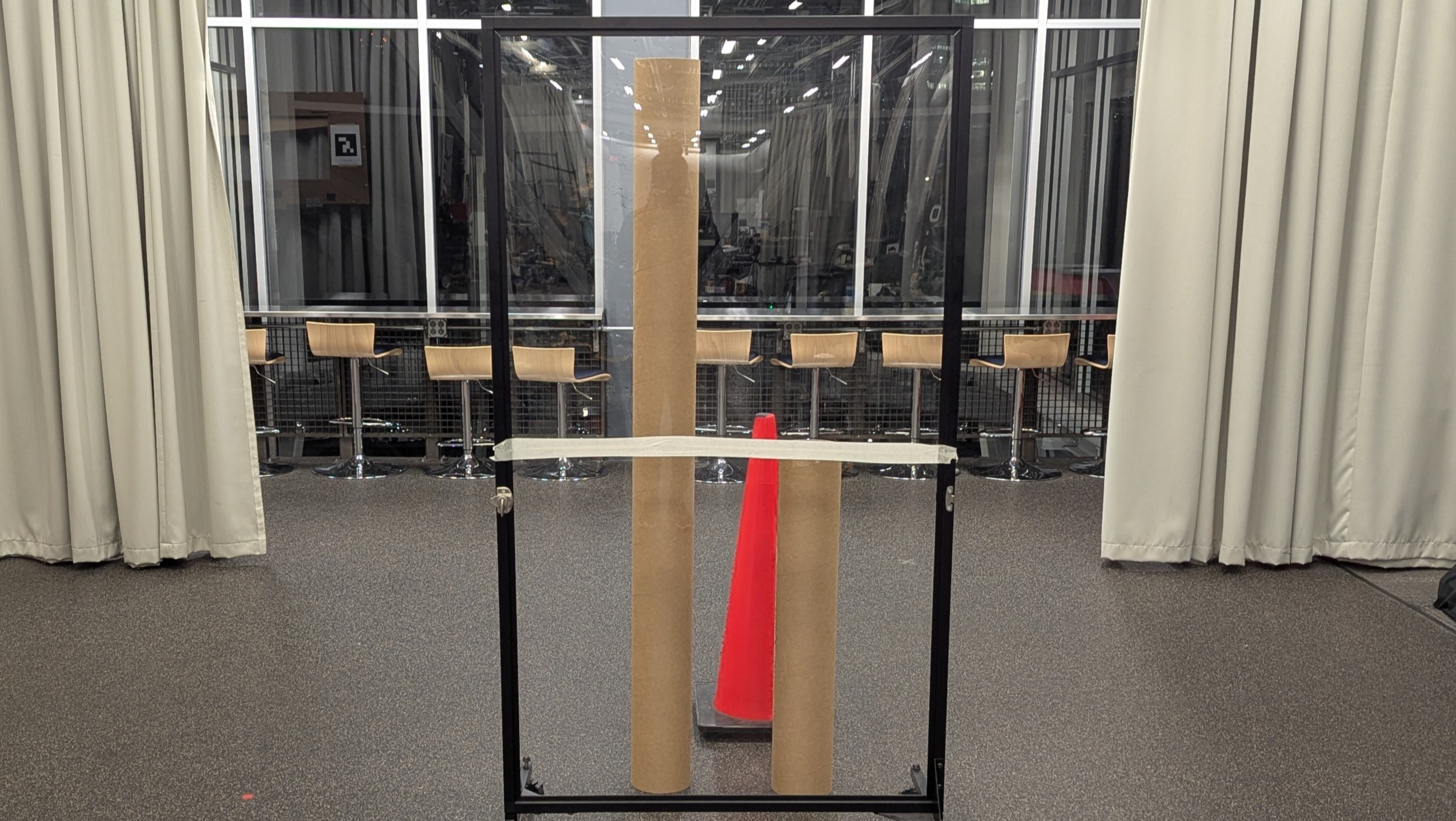}} \\
{\includegraphics[width = 0.28\linewidth]{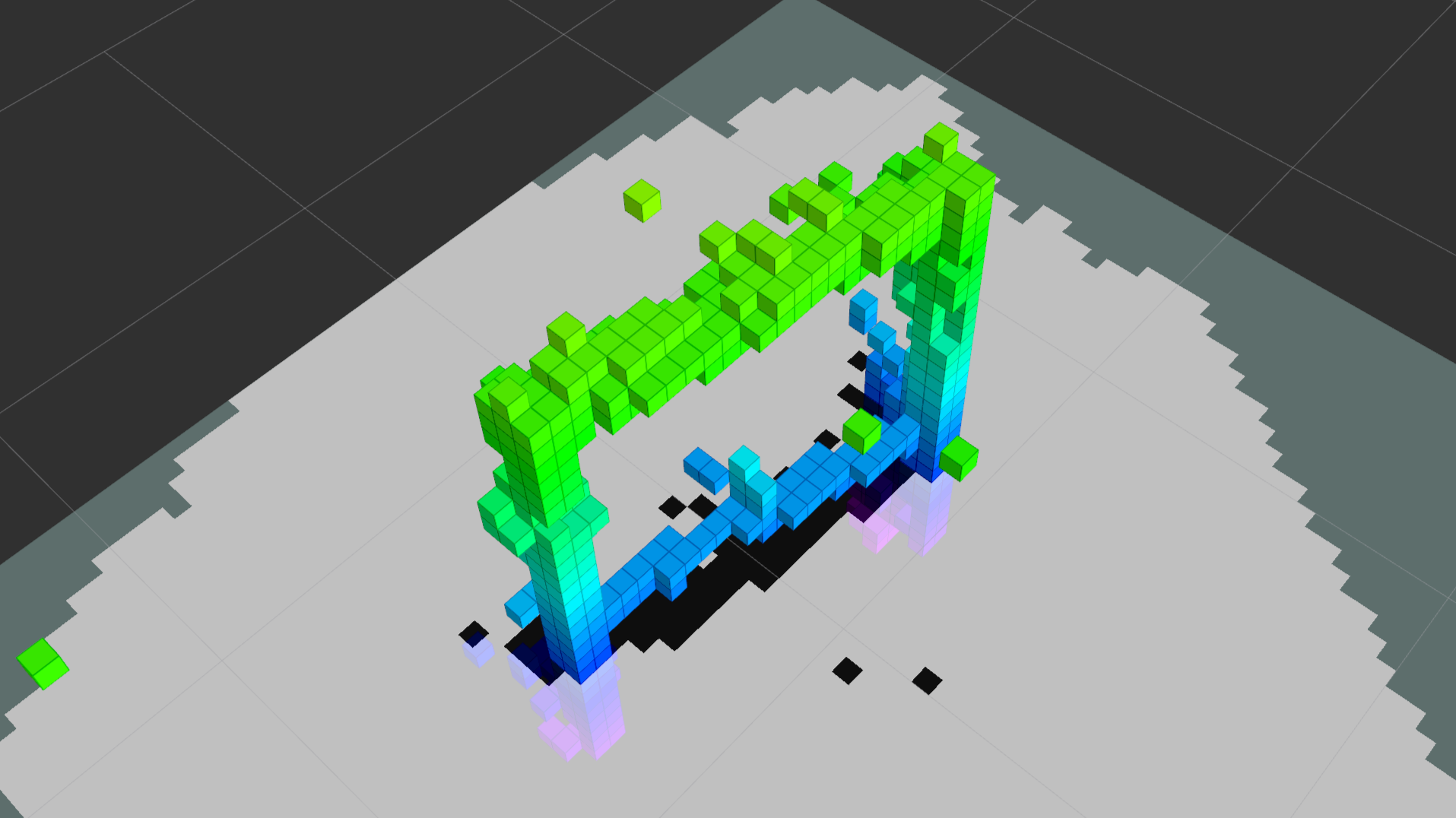}} &
{\includegraphics[width = 0.28\linewidth]{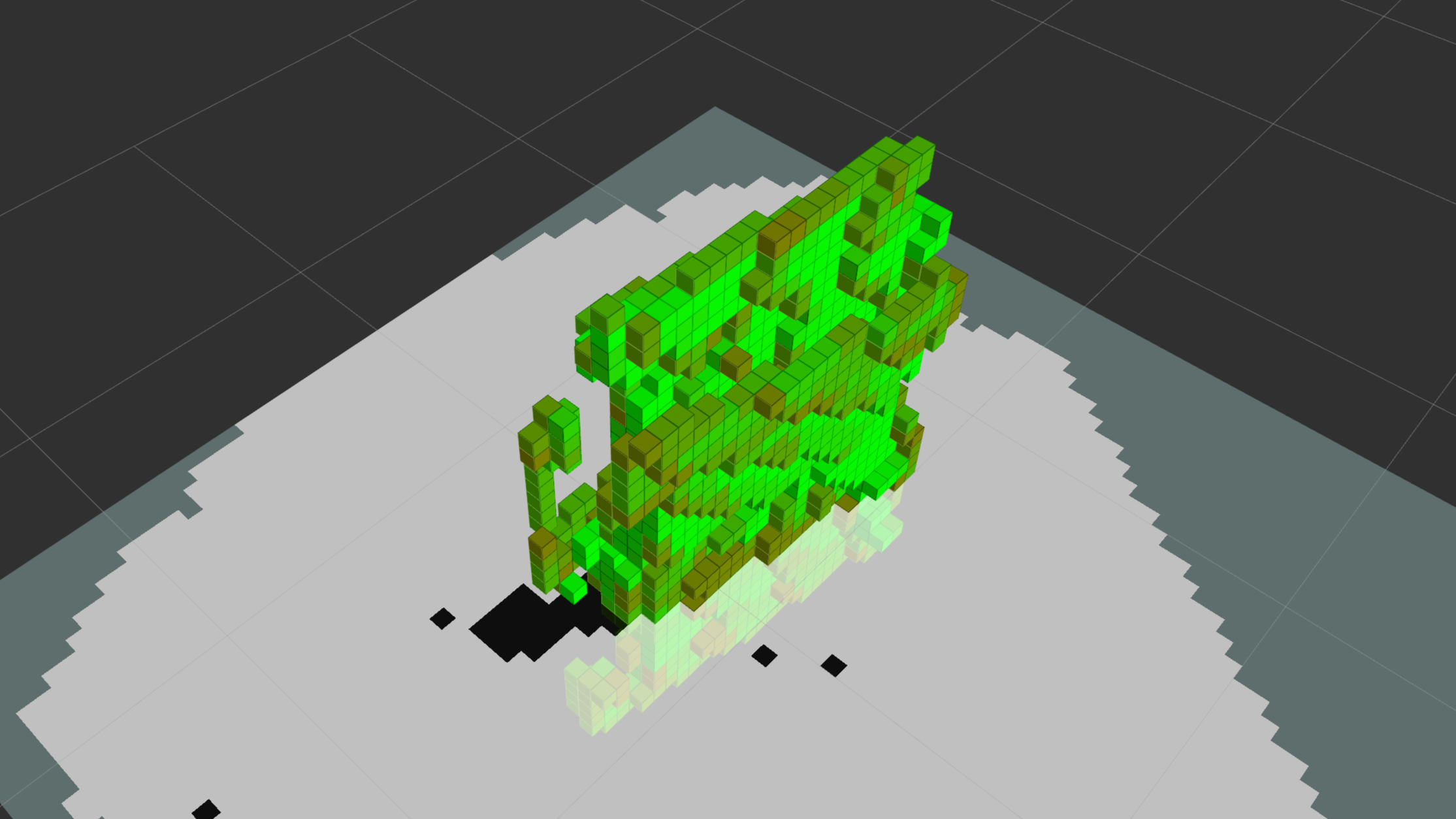}} &
{\includegraphics[width = 0.28\linewidth]{images/newimages/pane-clear_1.jpg}} \\
{\includegraphics[width = 0.28\linewidth]{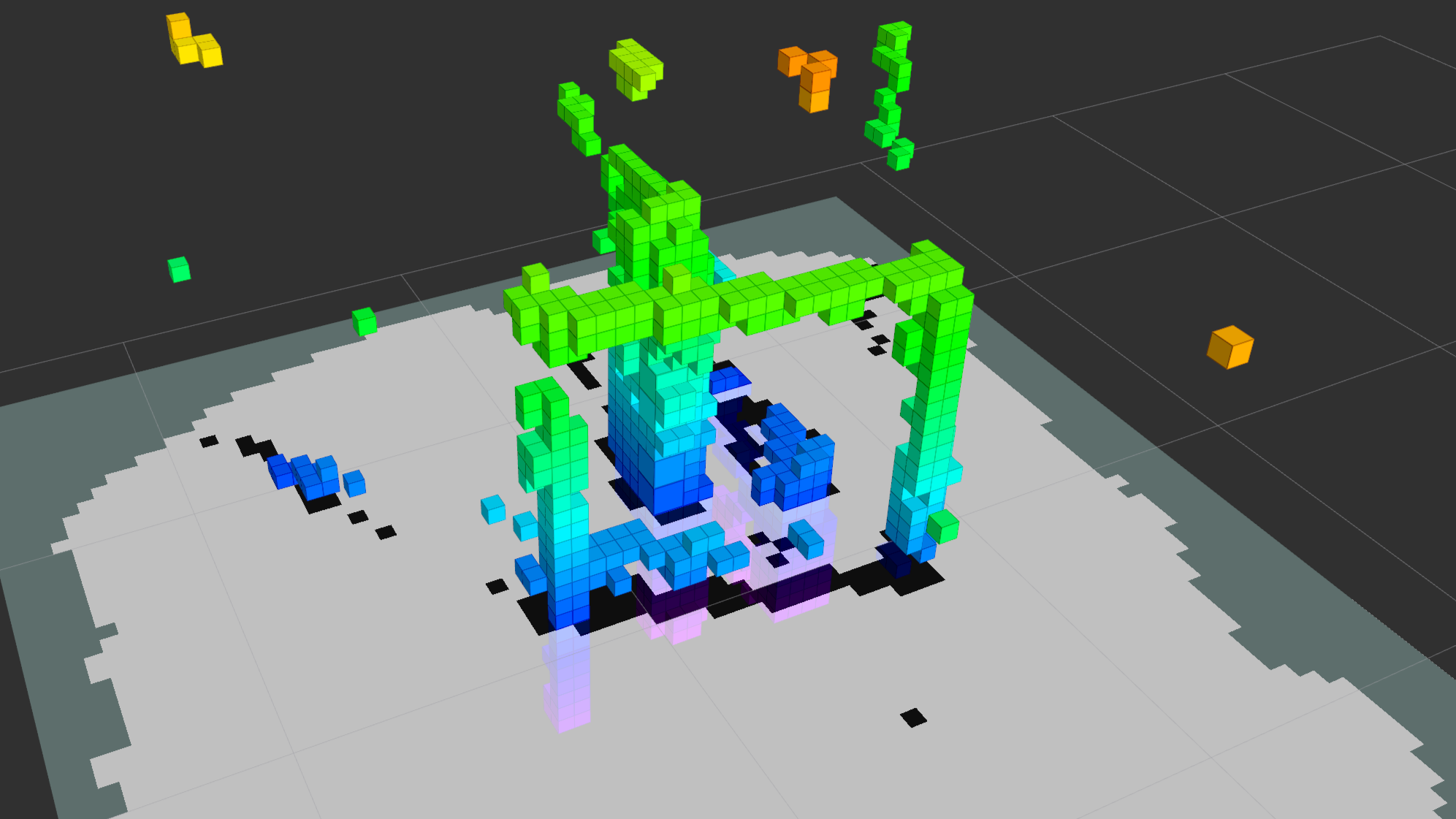}} &
{\includegraphics[width = 0.28\linewidth]{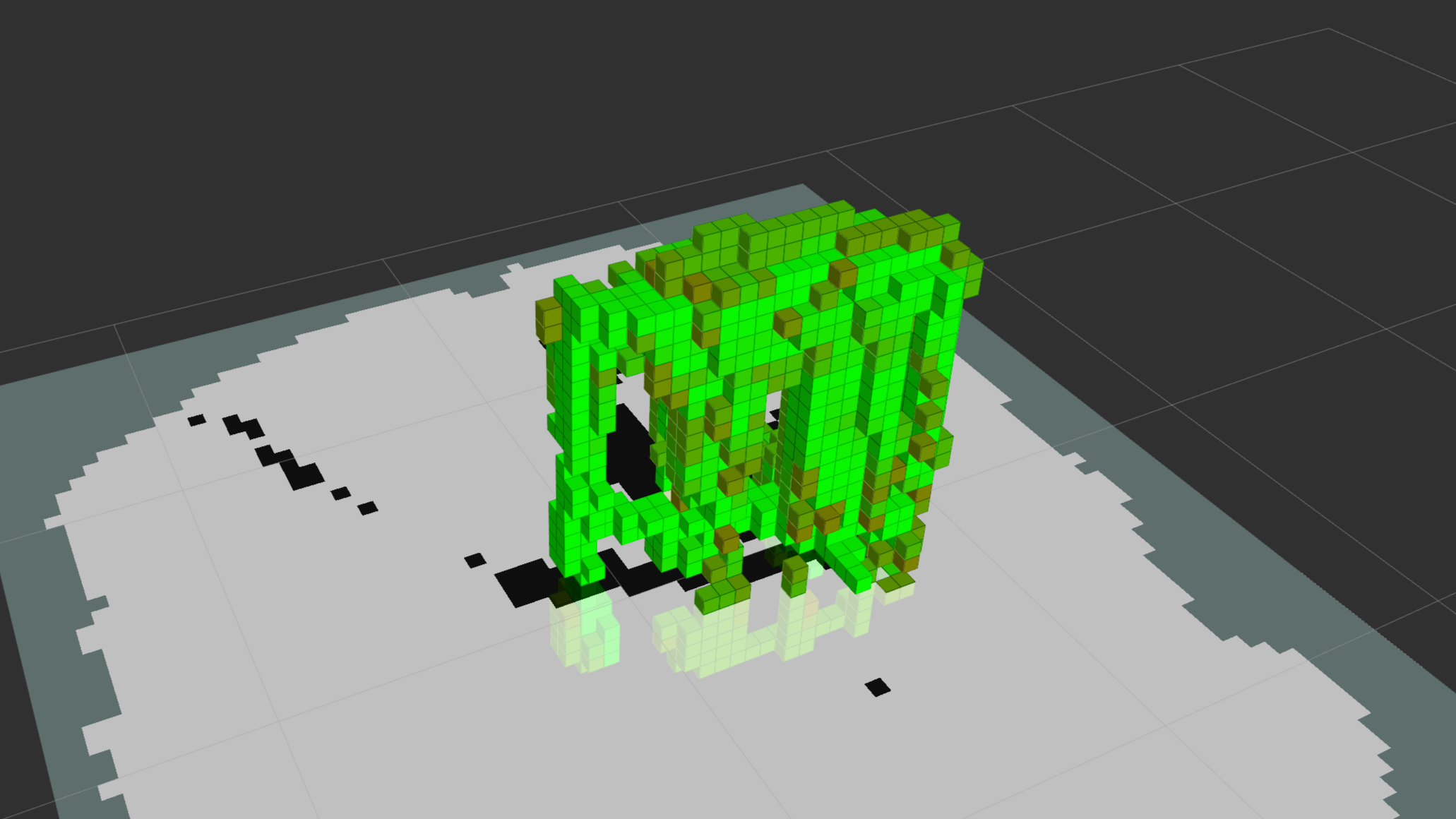}} &
{\includegraphics[width = 0.28\linewidth]{images/newimages/pane-occluded_1.jpg}} \\
{\includegraphics[width = 0.28\linewidth]{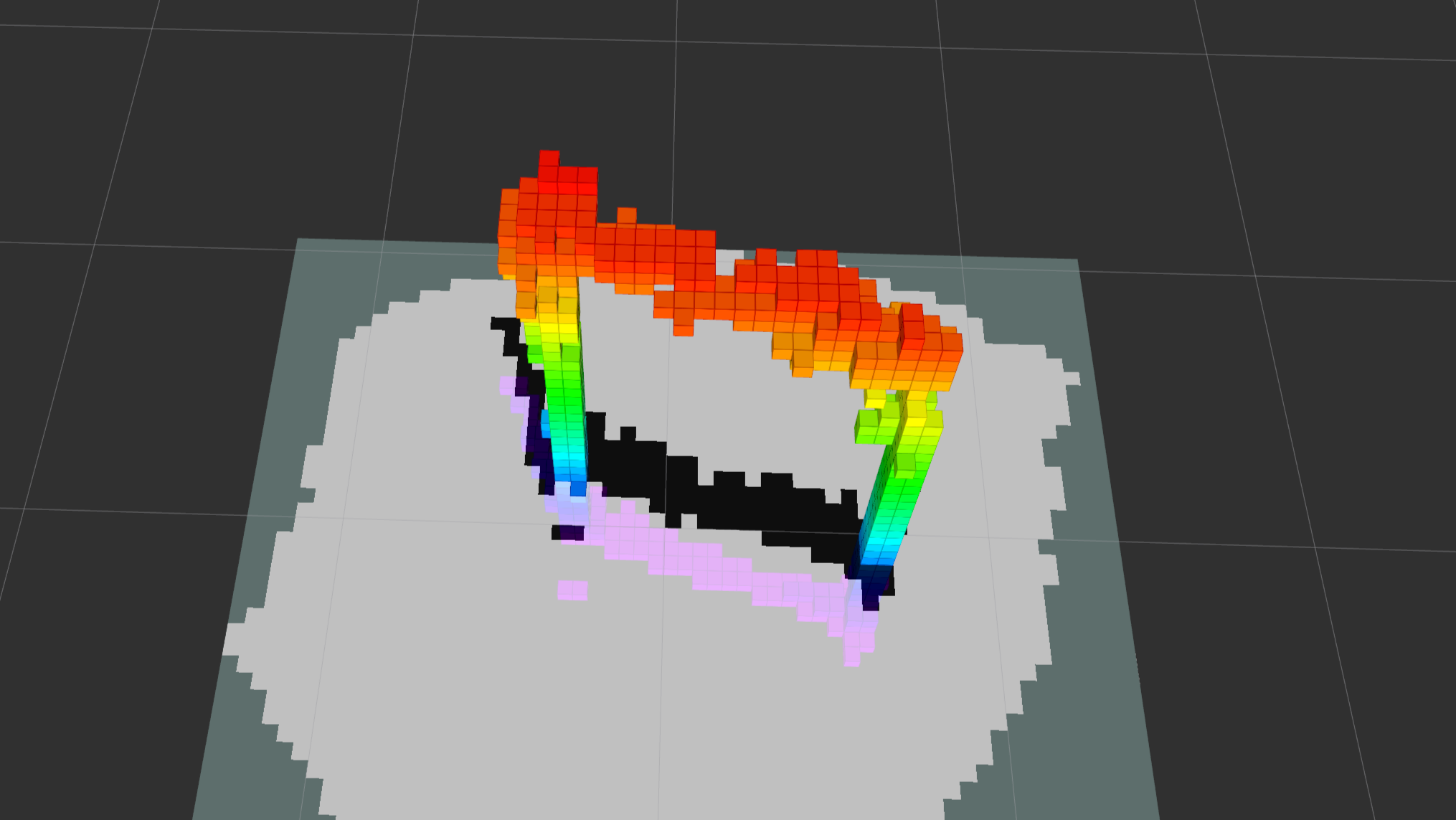}} &
{\includegraphics[width = 0.28\linewidth]{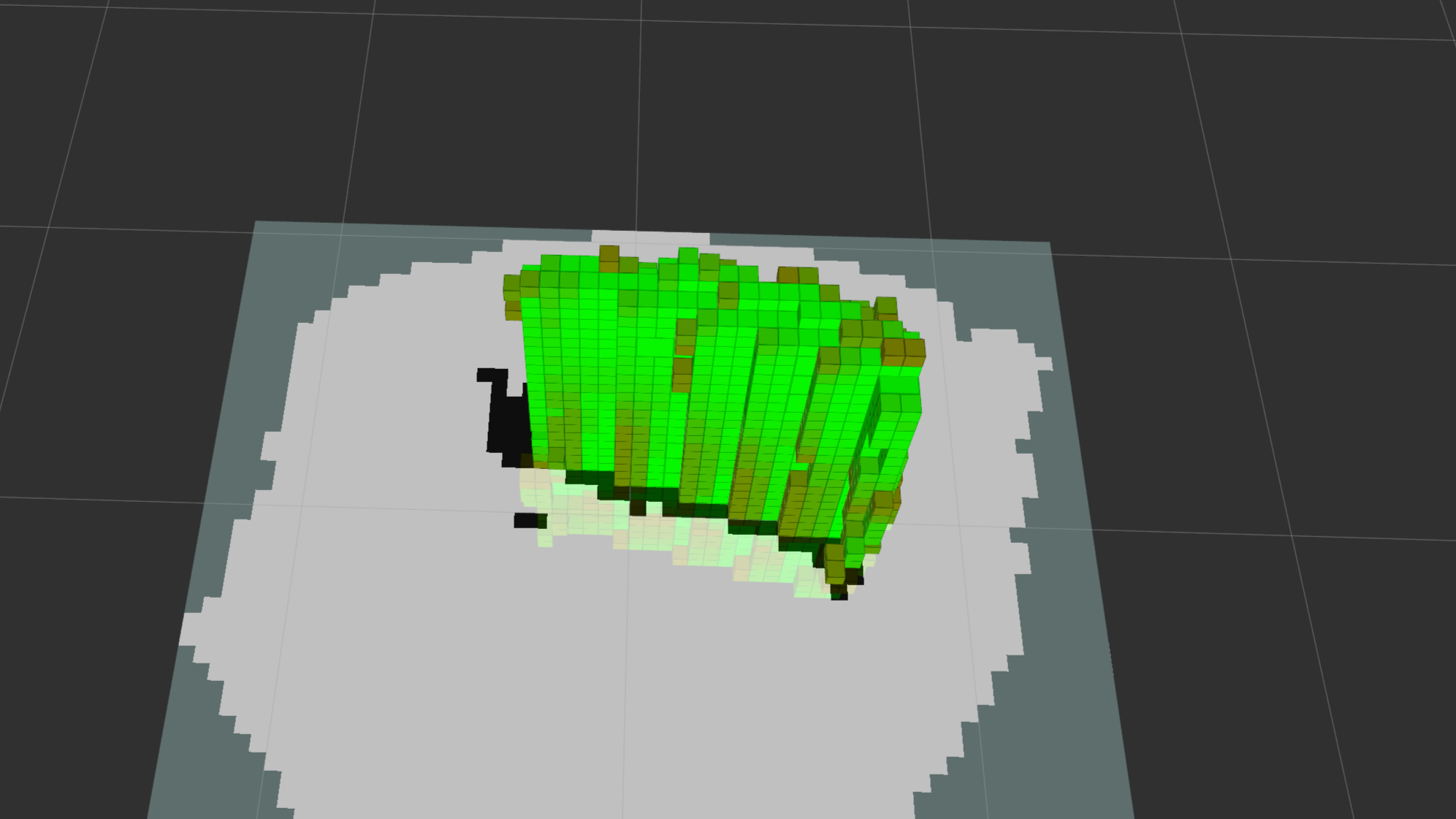}} &
{\includegraphics[width = 0.28\linewidth]{images/newimages/pane-clear_1.jpg}} \\
\subfloat[Baseline]
{\includegraphics[width = 0.28\linewidth]{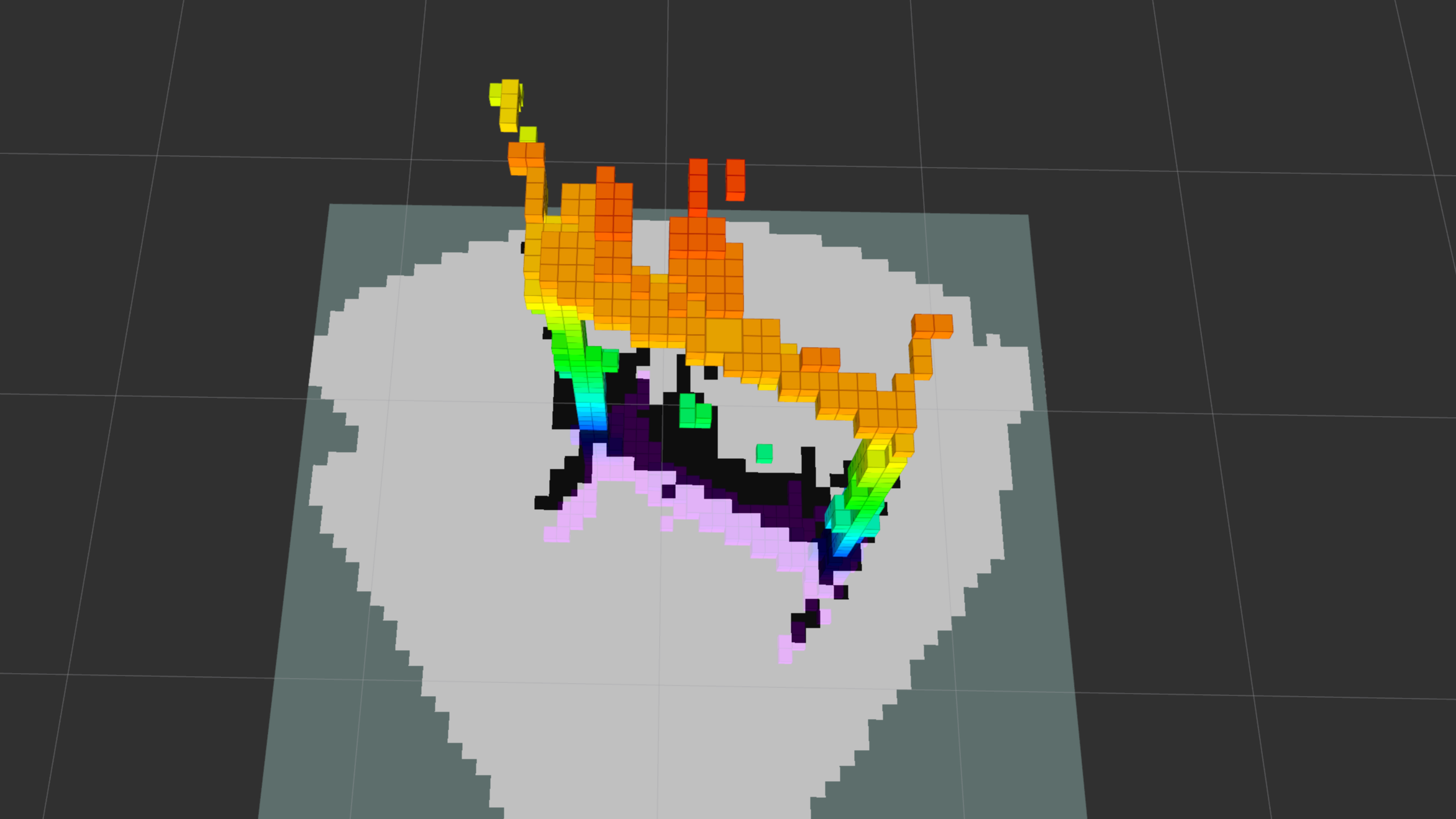}} &
\subfloat[Ours]
{\includegraphics[width = 0.28\linewidth]{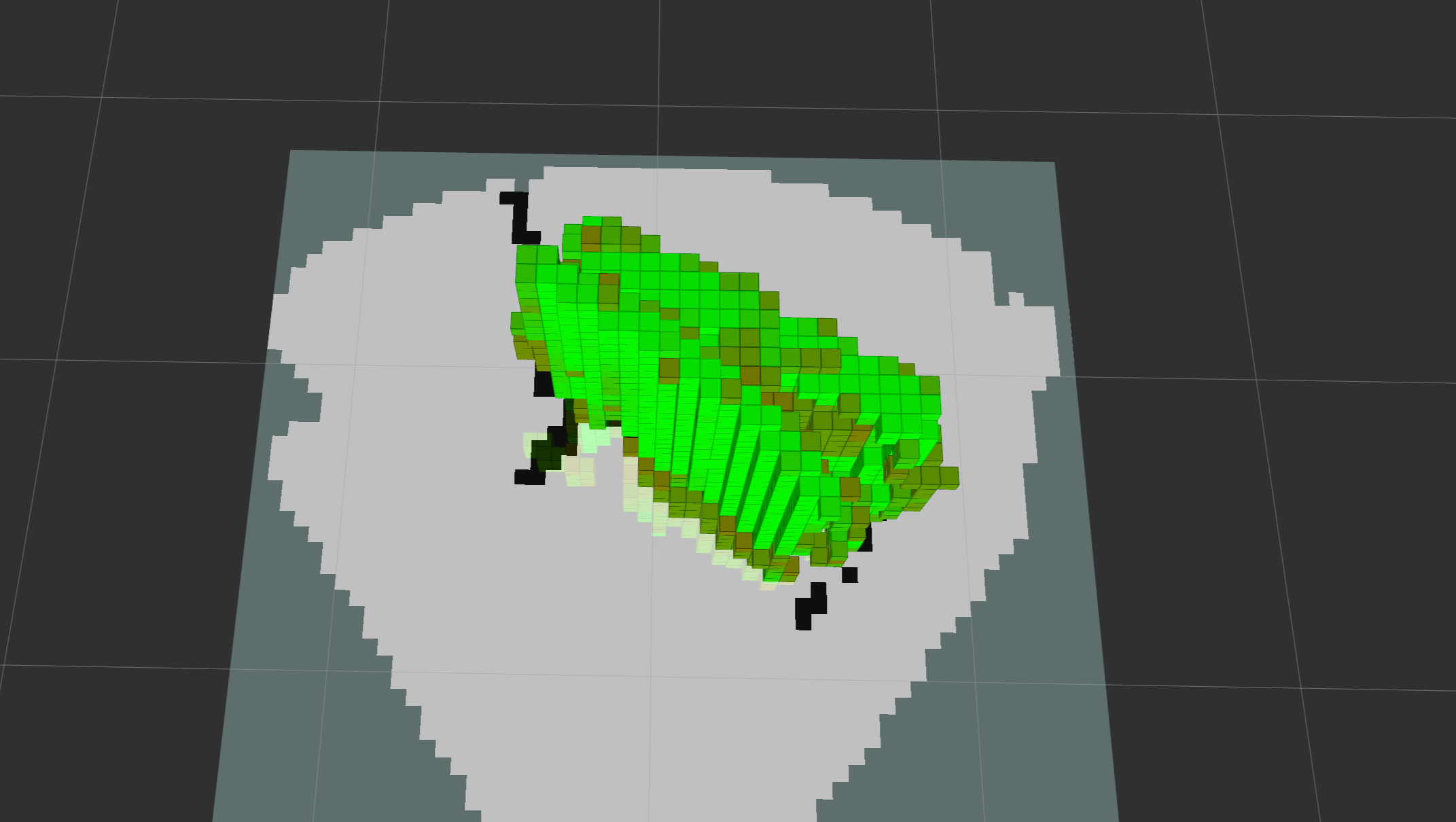}} &
\subfloat[Scene]
{\includegraphics[width = 0.28\linewidth]{images/newimages/pane-clear_1.jpg}} \\
\end{tabular}
\caption{Qualitative results for Mapping Experiments. Each row displays a different experimental condition from top to bottom: a head-on approach to a clear glass pane; a head-on approach with background objects; an approach at a 5-degree angle; an approach at a 5-degree angle with background objects; an approach at a 10-degree angle; and an approach at a 15-degree angle. The columns show the (a) baseline depth map, (b) our algorithm's reprojected map, and the (c) scene view. 
\vspace{-2em}
}
\label{fig:occupancy_maps}
\end{figure}

\noindent \textbf{Single Glass Pane Experiments: }
We tested the algorithm's fundamental ability to detect and map during all four of the aforementioned single glass pane experiments. To further display the algorithms ability to estimate the normal of the glass plane, we conducted two more experiments in which the glass pane was angled at 10 degrees and 15 degrees. The robot approached the glass and our algorithm correctly identified the speckle and corresponding empty space and the mapped depth was consistent to the glass pane border and ground truth tape measurements.

\noindent \textbf{Vicon Experiments: }
We conducted mapping experiments within a Vicon motion capture space to provide ground truth for the robot's pose. We built a mock glass room within the Vicon space in a rectangular shape, containing opaque walls as well as 5 glass pane walls. We tested two scenarios: a hand-carried, smooth trajectory and an autonomous flight test using Vicon odometry. Our algorithm was processed onboard the robot in both scenarios. 
The results are shown in Table~\ref{tab:experimental_results}.
The minimal performance degradation 
in the autonomous flight 
indicates the algorithm's robustness.

\subsection{Real-World Environment Experiments}

\begin{figure*}[!h]
\centering
\begin{tabular}{c c c c c}
{\includegraphics[width = 0.17\linewidth]{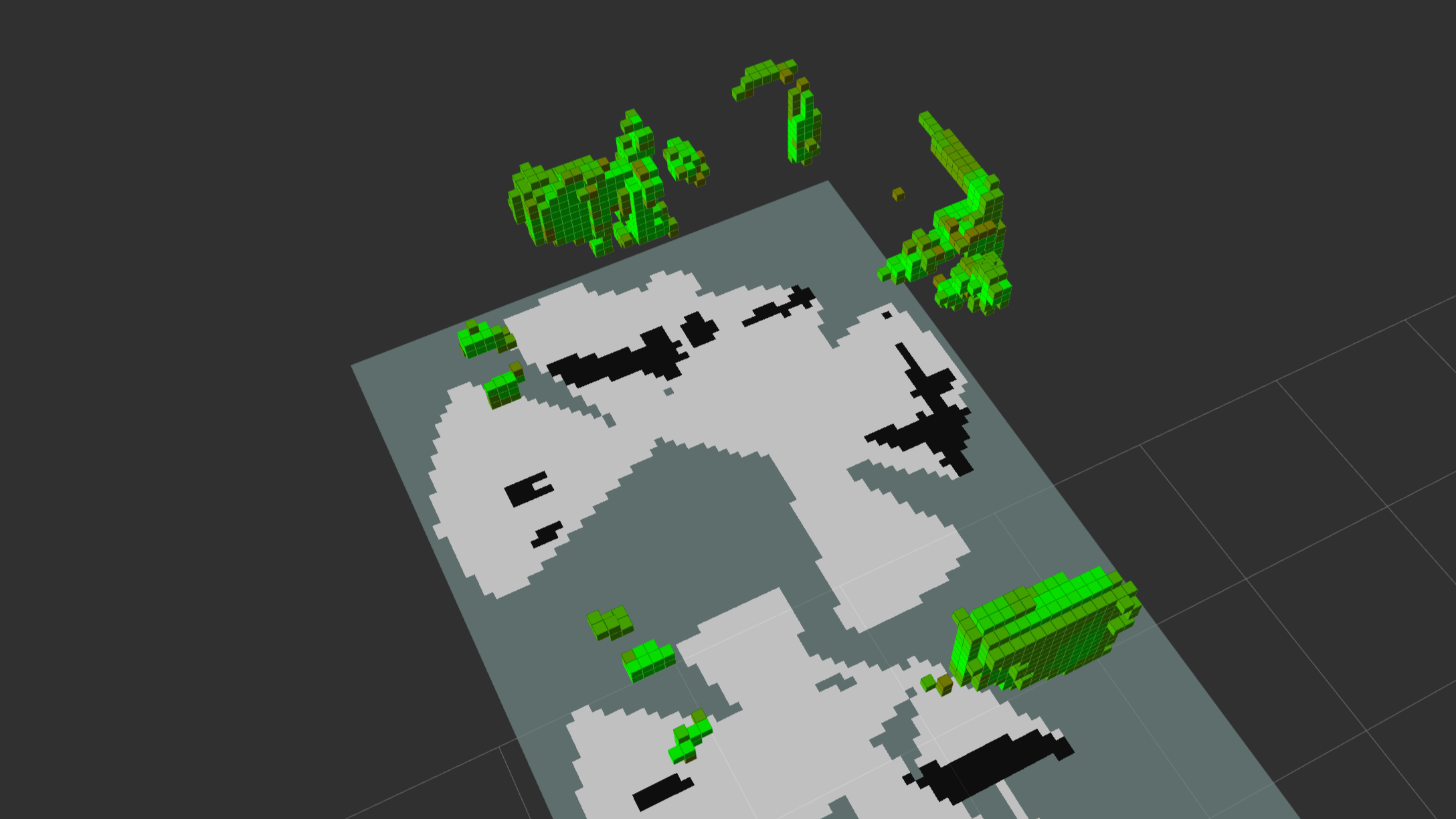}} &
{\includegraphics[width = 0.17\linewidth]{images/newimages/vicon-4.png}} &
{\includegraphics[width = 0.17\linewidth]{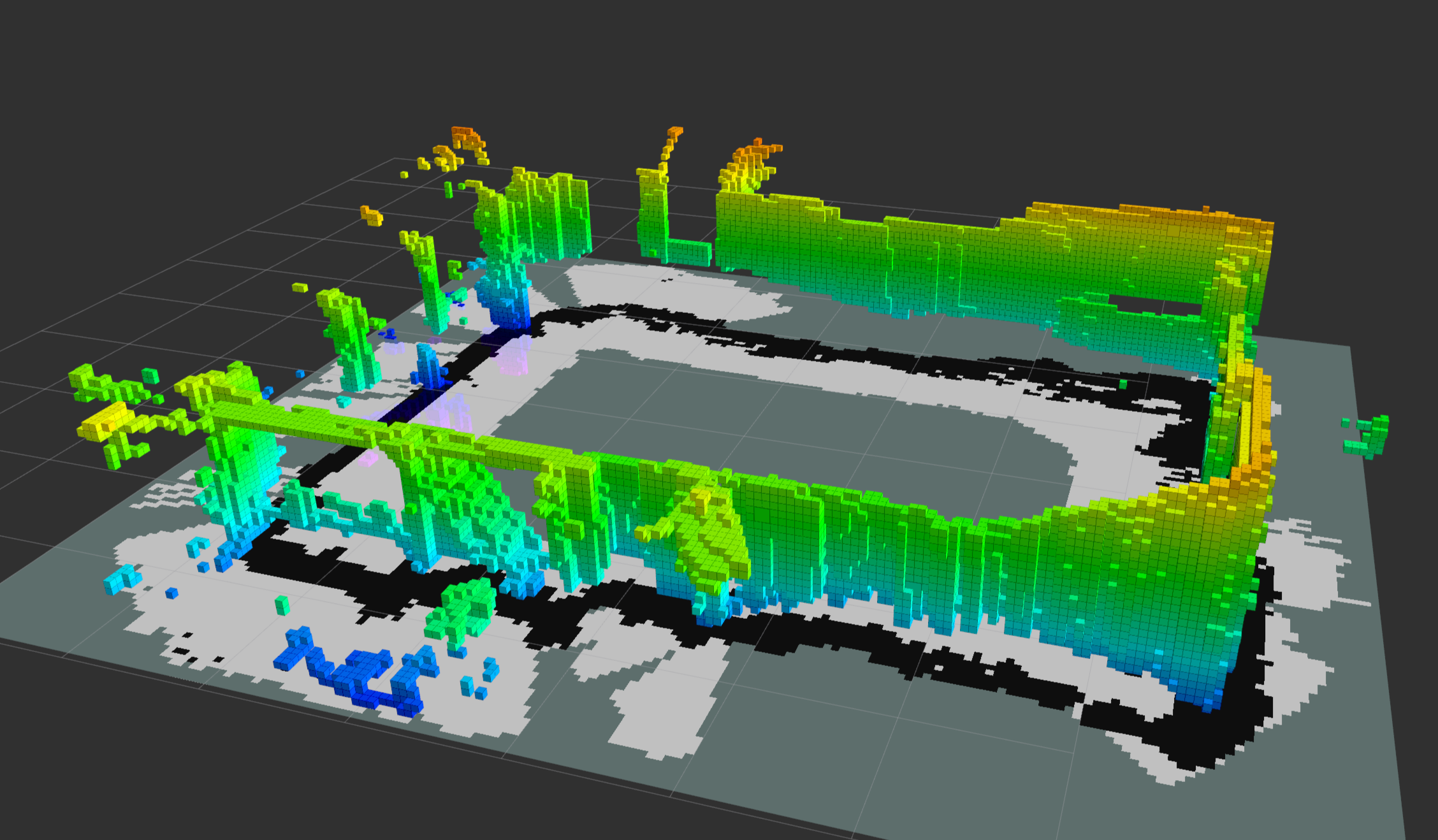}} &
{\includegraphics[width = 0.17\linewidth]{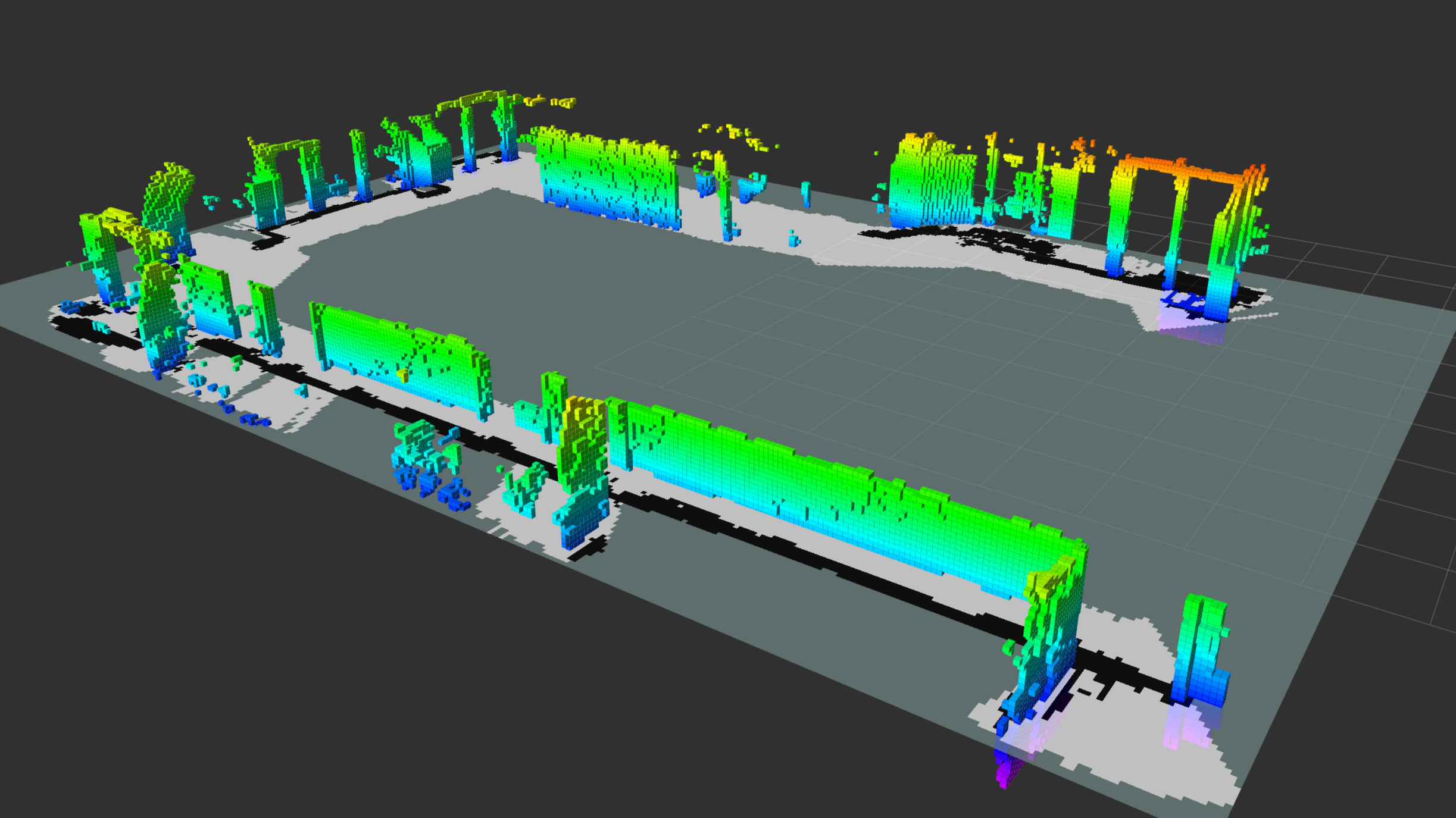}} &
{\includegraphics[width = 0.17\linewidth]{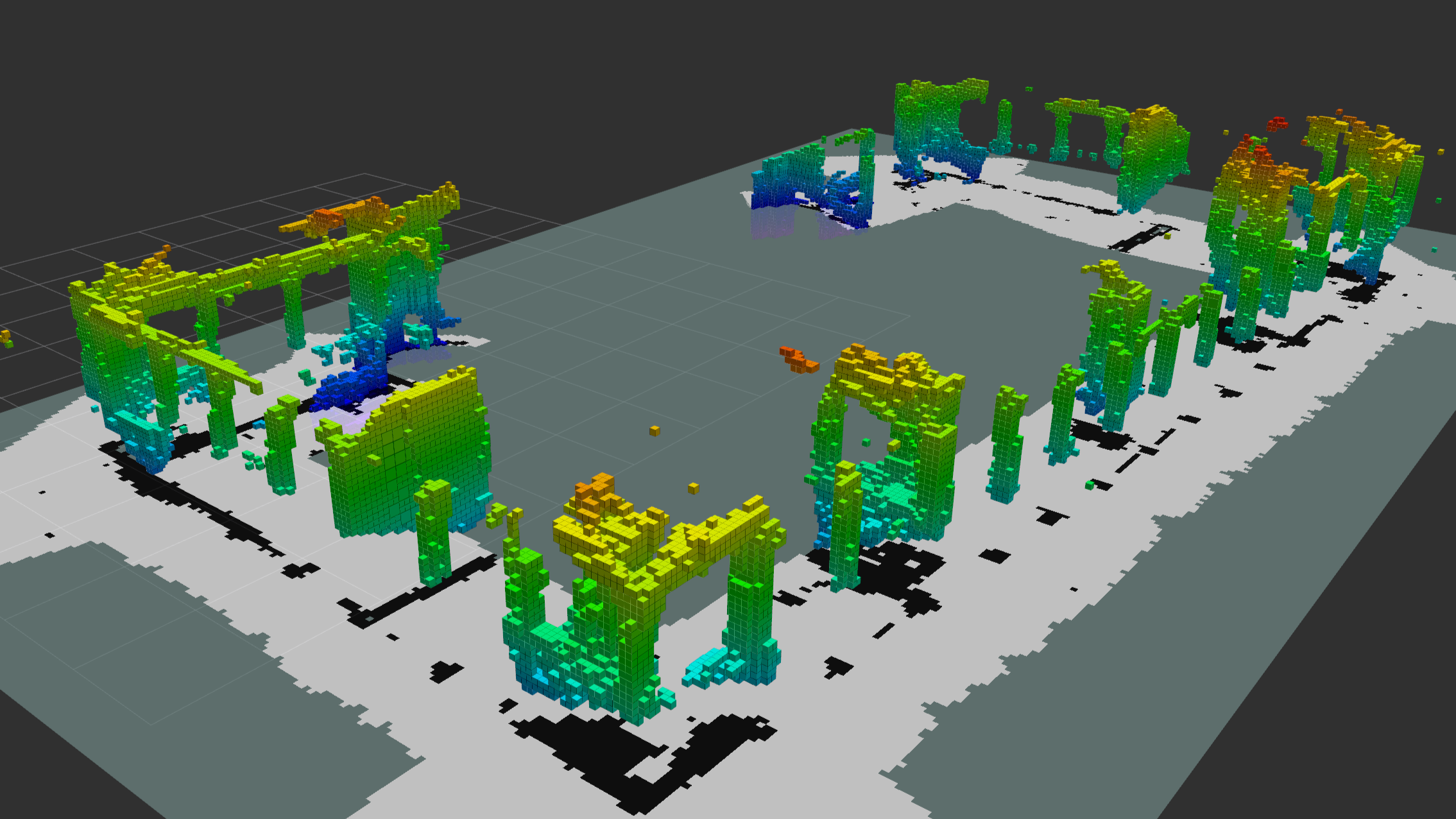}} \\
{\includegraphics[width = 0.17\linewidth]{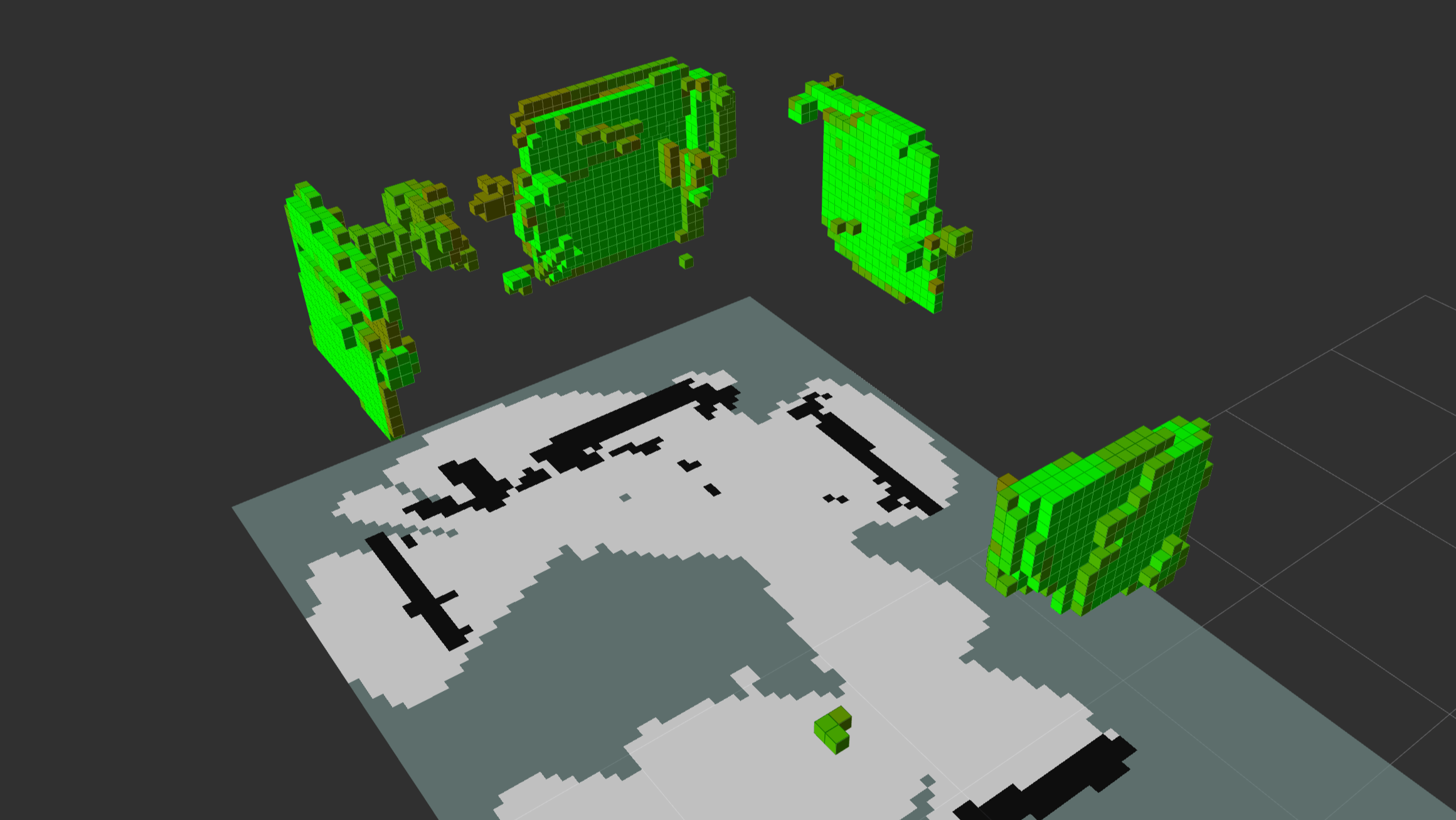}} &
{\includegraphics[width = 0.17\linewidth]{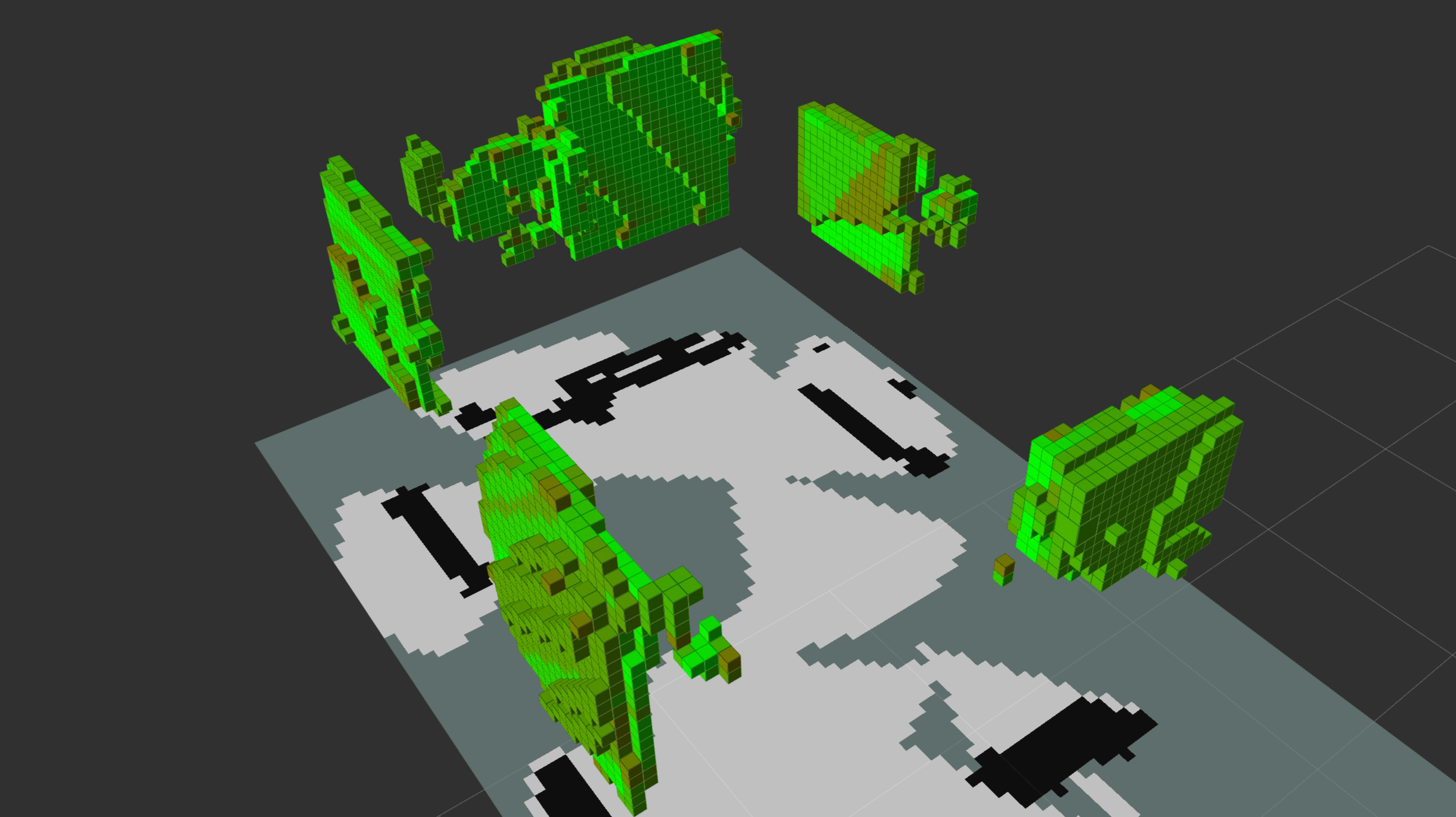}} &
{\includegraphics[width = 0.17\linewidth]{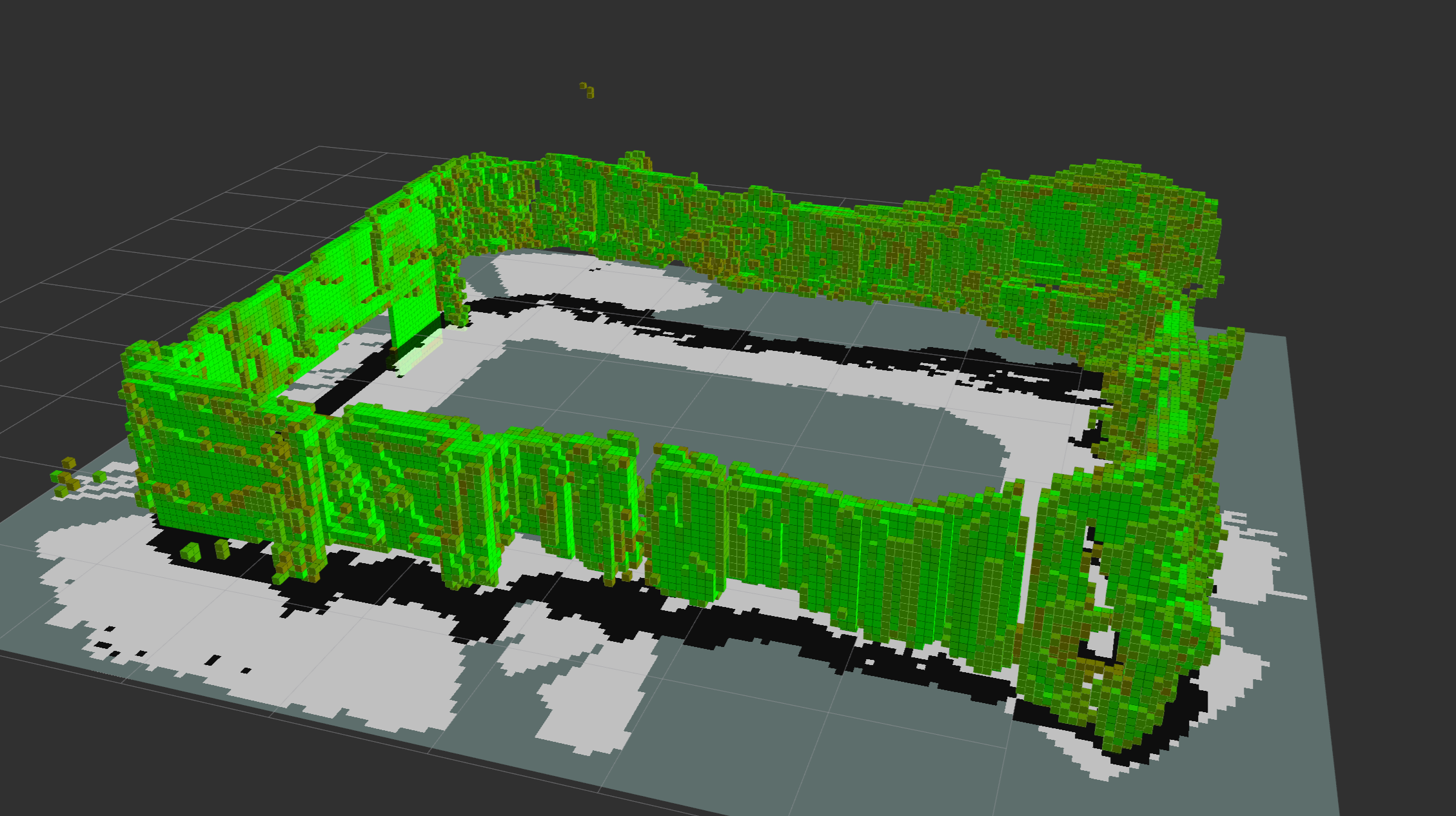}} &
{\includegraphics[width = 0.17\linewidth]{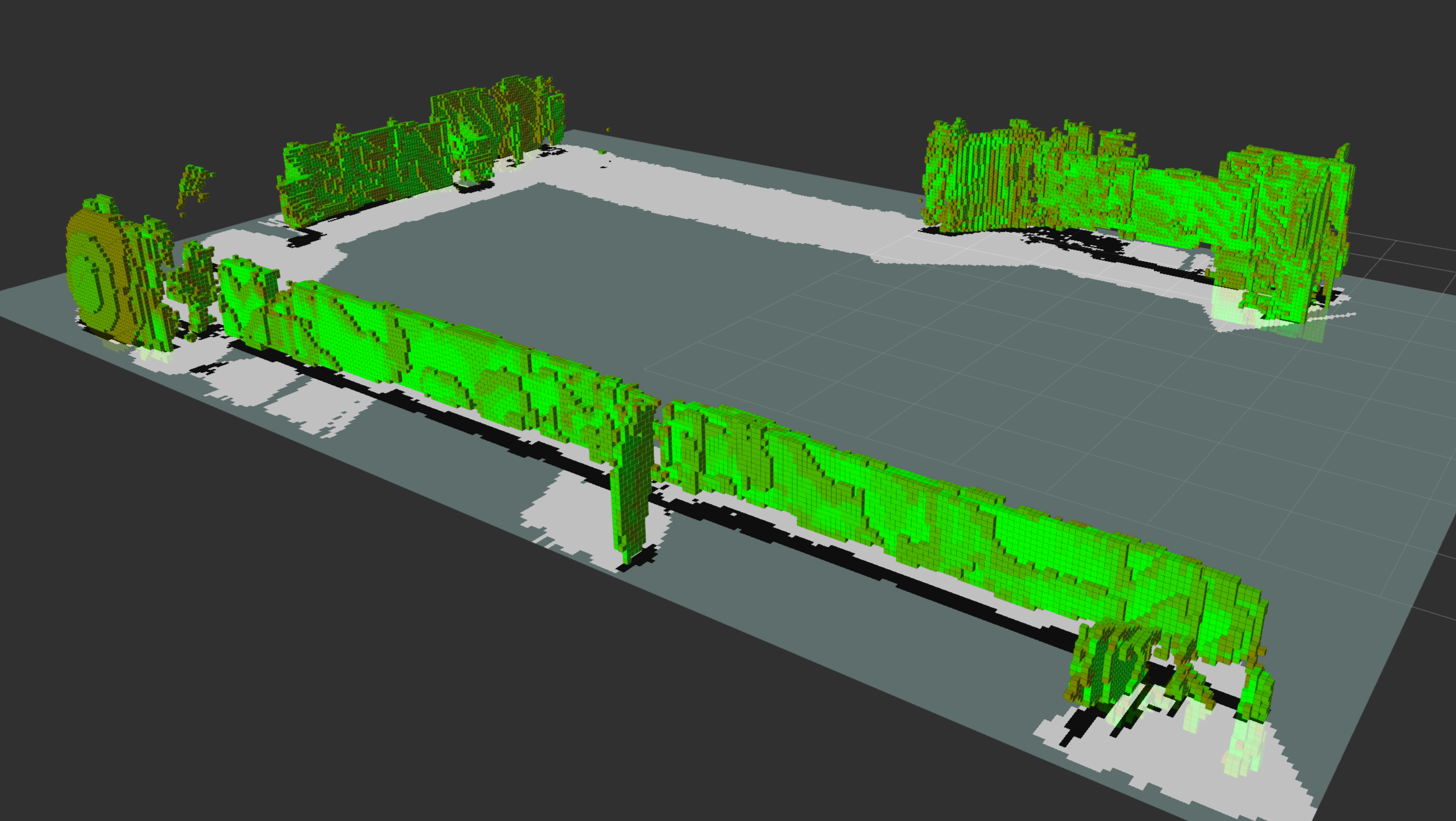}} &
{\includegraphics[width = 0.17\linewidth]{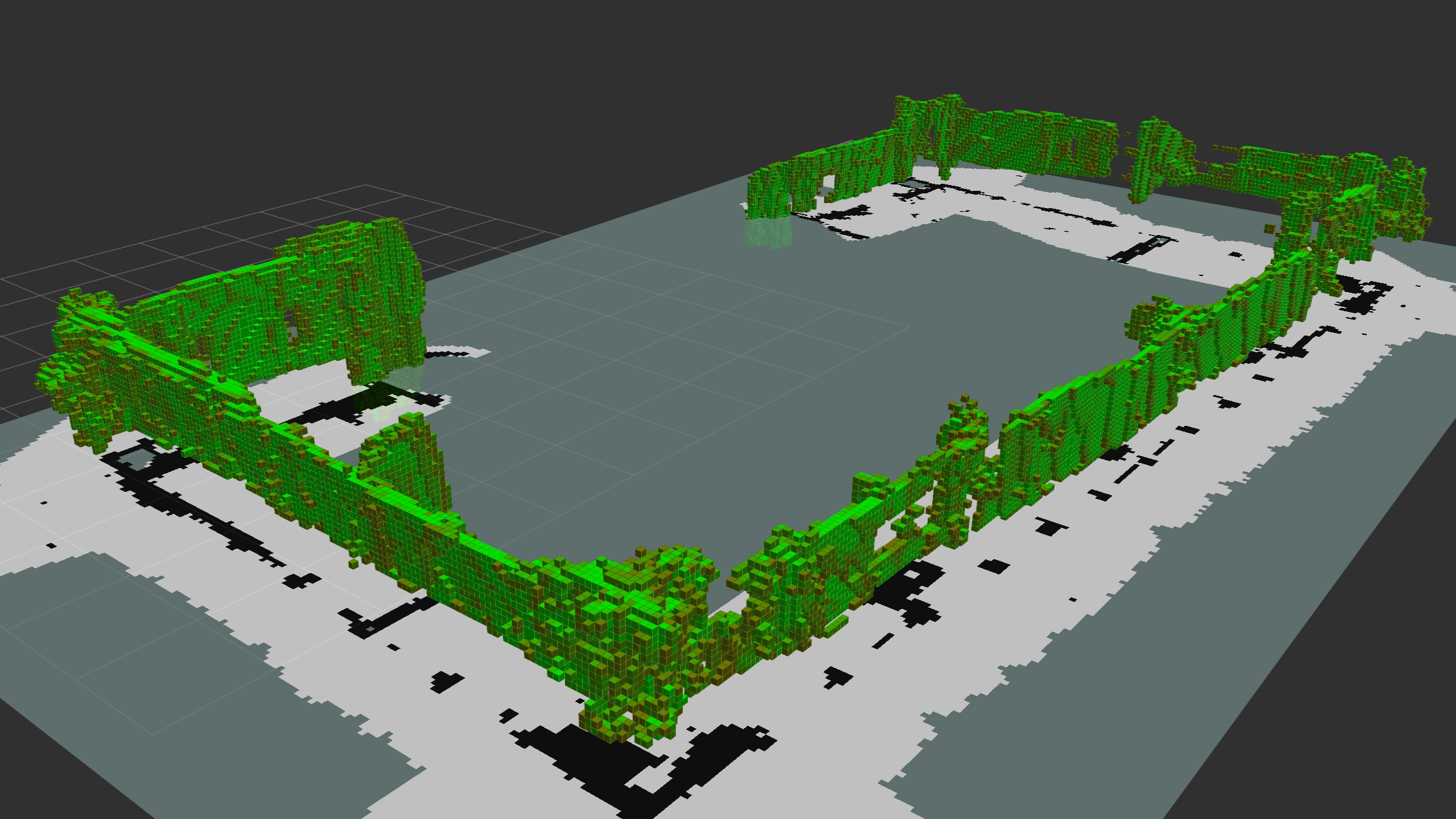}} \\
\subfloat[Vicon (Handheld)]
{\includegraphics[width = 0.17\linewidth]{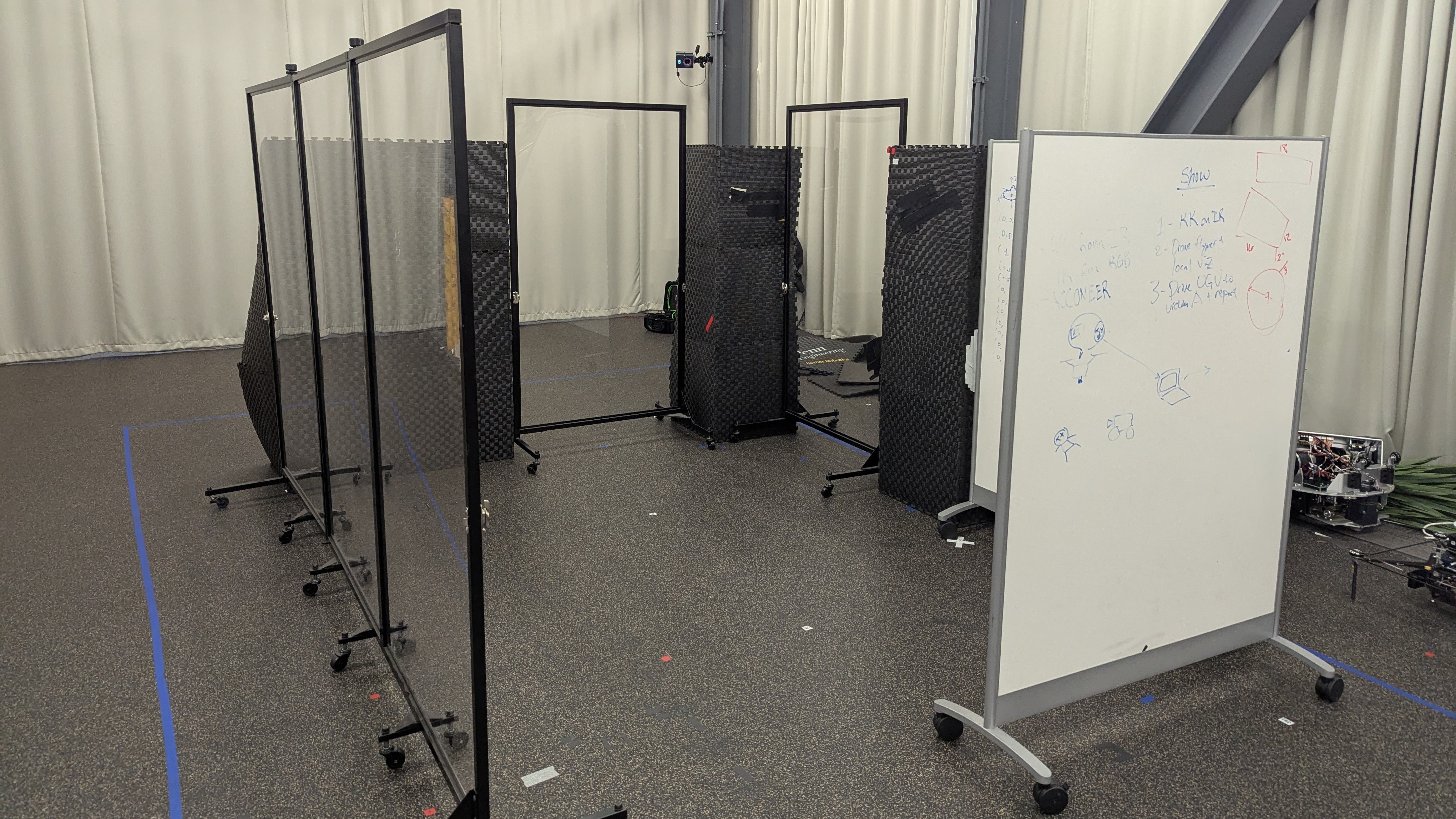}} &
\subfloat[Vicon (Autonomy)]
{\includegraphics[width = 0.17\linewidth]{images/newimages/scene-1-t_1.jpg}} &
\subfloat[Half-Glass Room]
{\includegraphics[width = 0.17\linewidth]{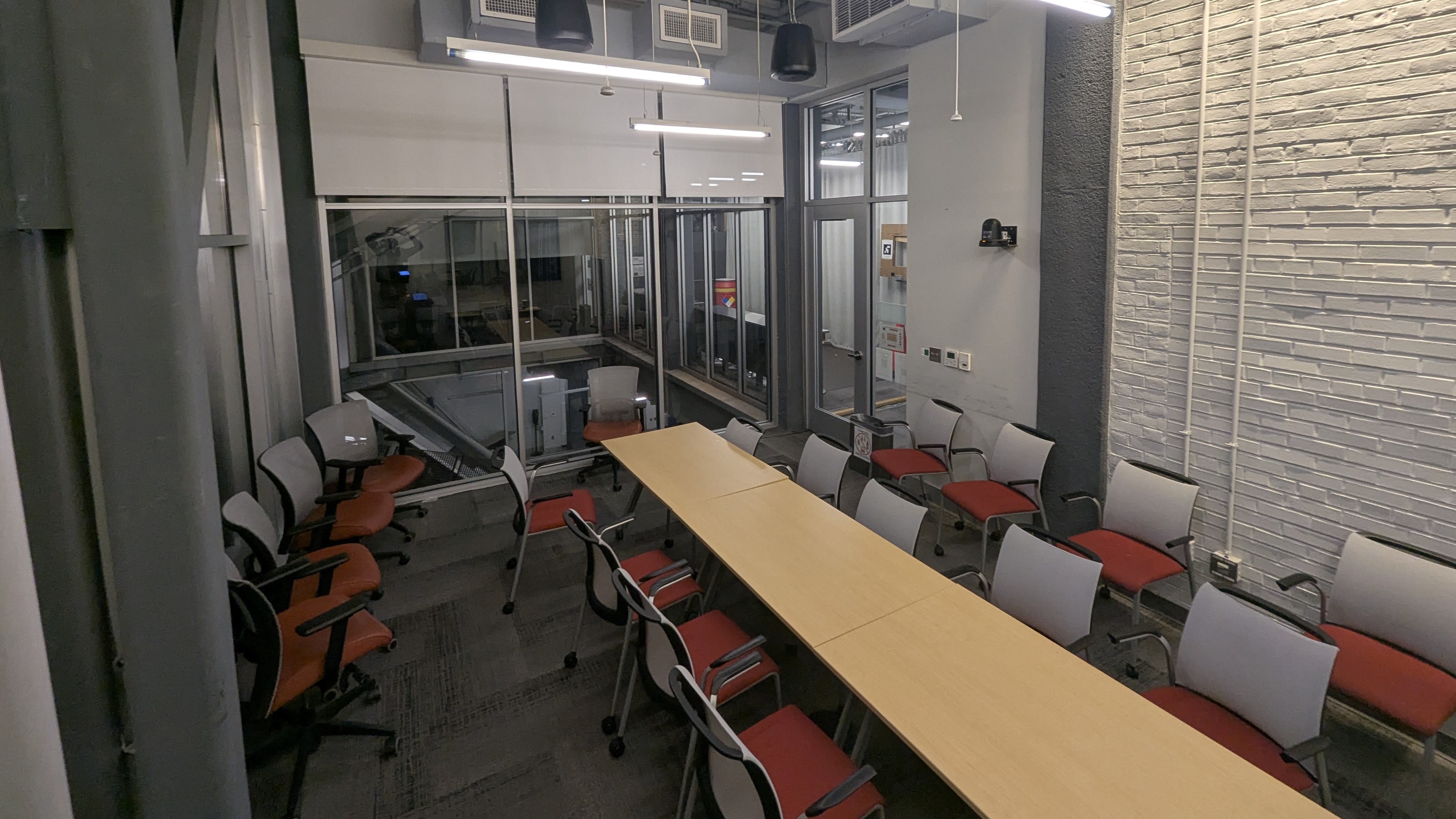}} &
\subfloat[Glass Hallway]
{\includegraphics[width = 0.17\linewidth]{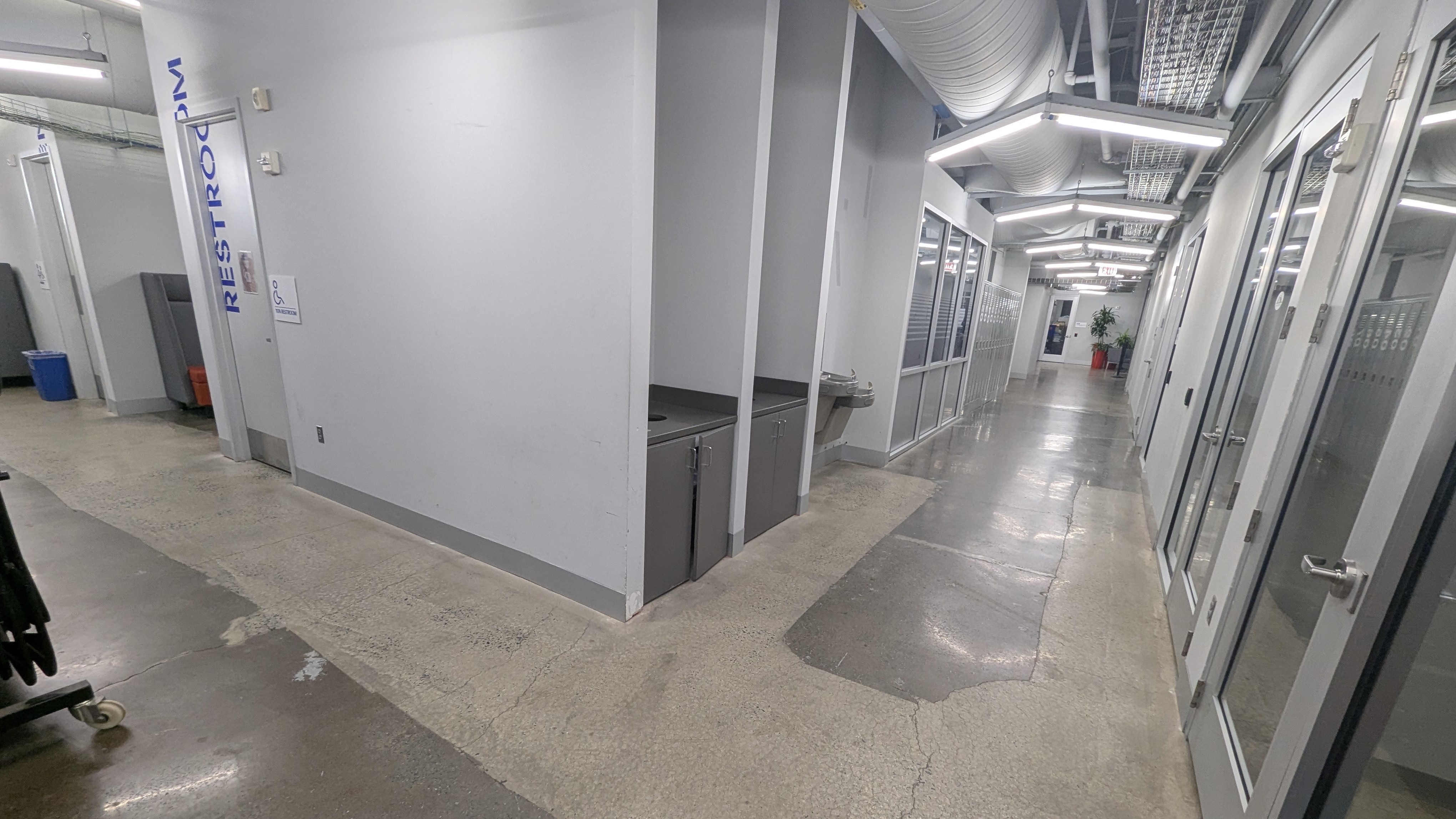}} &
\subfloat[Glass Perimeter]
{\includegraphics[width = 0.17\linewidth]{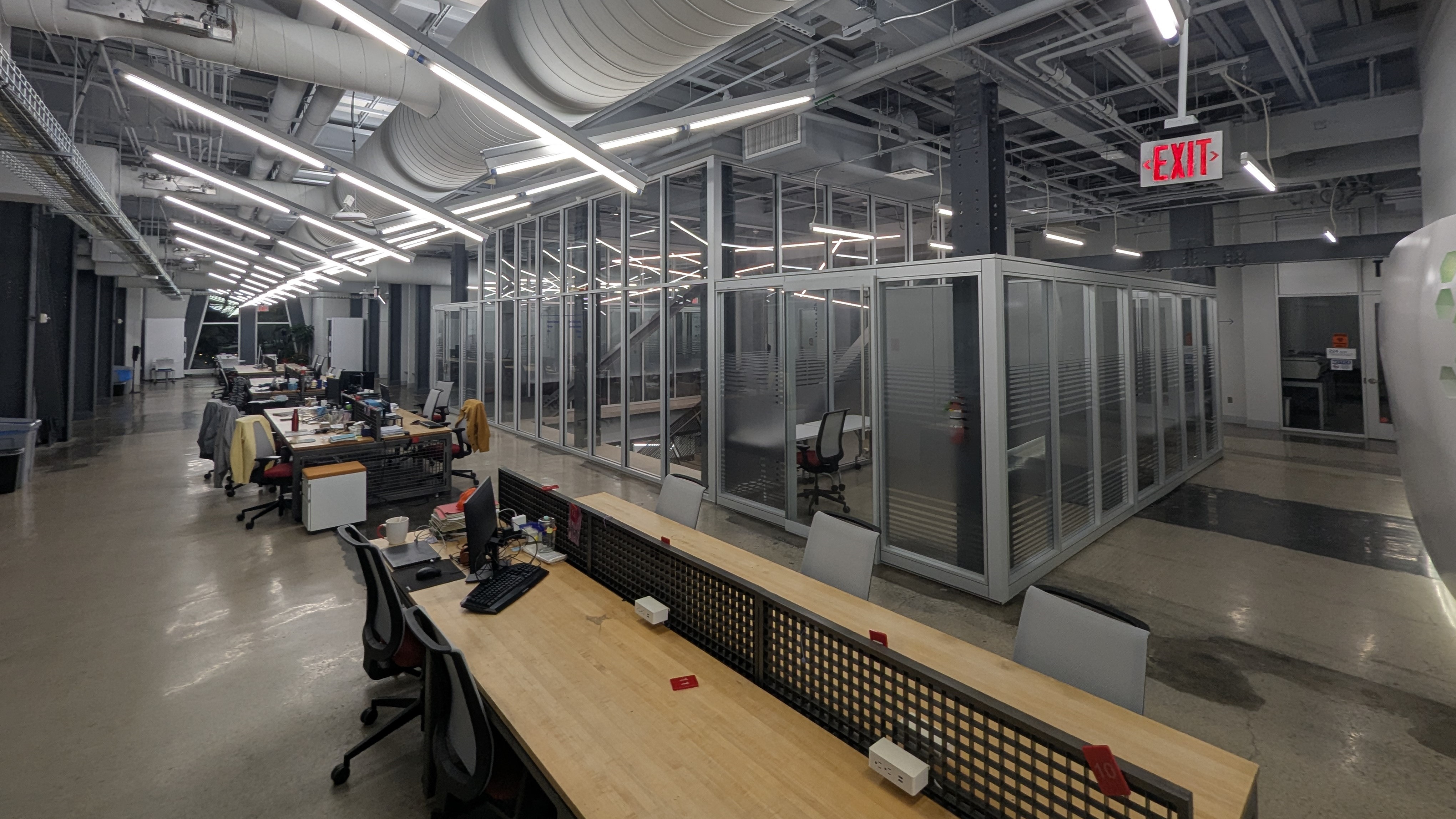}} \\
\end{tabular}
\caption{Qualitative results from real-world mapping experiments, showing a comparison between the (TOP) baseline depth map and (MIDDLE) our algorithm's reprojected map. The columns represent different environments (Vicon, Half-Glass Room, Glass Hallway, and Glass Perimeter), while the bottom row (BOTTOM) provides a visual reference of the scene. 
\vspace{-1em}
}
\label{fig:occupancy_maps2}
\end{figure*}

Following the controlled tests, we moved to more complex real-world environments to evaluate the system's robustness. During these experiments, all computation was done onboard the robot. The qualitative results in Figure 7 visually demonstrate the high precision and completeness of our mapping algorithm in these environments. The quantitative results are displayed in Table~\ref{tab:experimental_results}.

\noindent \textbf{Half-Glass Room: }
In this experiment, the robot operated within a room where half of the room is made of several glass windows instead of walls, while the other half is opaque. The algorithm correctly distinguished between the transparent and opaque surfaces, seamlessly integrating the mapped transparent planes with the existing depth data from the solid walls. This scene included stickers on some of the glass panes, preventing speckles from being detected. 

\noindent \textbf{Glass Hallway: }
We performed a mapping experiment in a hallway with large glass panes on one side, which presented challenges from complex reflections and varying ambient light. The algorithm successfully detected and mapped the glass planes along the hallway. 
In these experiments, multiple glass panes contained stickers that prevented the algorithm from detecting speckles, causing a decrease in mIOU.

\noindent \textbf{Glass Perimeter: }
We performed a glass mapping experiment in a building floor with a centroid area with large glass panes on all sides. The algorithm successfully detected and mapped the glass planes within the scene. 
This scene included multiple scenarios where an opaque object was directly behind a glass pane and within our defined sonar depth filtering threshold, which prevented the background object from being removed from the image. This is further discussed in our Discussion and Limitations Section~\ref{sec:Discussion and Limitations}. 

\subsection{Onboard Processing Performance}
Our full algorithmic pipeline executes directly on the quadrotor's embedded processor. The system operates at a real-time frequency of \textbf{2 Hz}, using approximately \textbf{20\% of a single CPU core} on the VOXL2's QRB5165 processor. This low computational footprint reserves power for other critical functions like odometry and flight control. We specifically conducted our experiments at this processing rate due to SWaP constraints to allow room for other onboard processes that the robot required for flight. To demonstrate scalability, the system can also run at a maximum rate of \textbf{10 Hz}, which aligns with our Time-of-Flight (ToF) sensor's maximum frame rate onboard our robot. This requires the full allocation of a single CPU core, providing a flexible trade-off between resource use and operational frequency. Compared to existing methods like GDNet and GlassSemNet, our approach offers significant advantages in performance and efficiency. Table~\ref{tab:performance_comparison} summarizes this comparison. Due to the variety of onboard processing our robot requires during flight (eg. odometry), it would be unable to process GDNet or GlassSemNet onboard the CPU.

\begin{table}[h!]
\centering
\caption{Comparison of Onboard Processing Performance}
\label{tab:performance_comparison}
\setlength{\tabcolsep}{7pt} 
\footnotesize 
\begin{tabular}{l|c|c|c}
\toprule
\textbf{Metric} & \textbf{Our Method} & \textbf{GDNet} & \textbf{GlassSemNet} \\
\midrule
Processing Rate (Hz) & 2-10 & 1.5 & 0.4  \\
Input Rate (Hz) & 2-10 & 19 & 19 \\
CPU Usage (\%) & 20-100 & 600 & 800 \\
\bottomrule
\end{tabular}
\end{table}

\section{Discussion and Limitations}
\label{sec:Discussion and Limitations}

Our algorithm was tested using three distinct presets. Preset 3, the most robust configuration, was employed for the Vicon handheld and autonomous flight experiments, utilizing a circularity threshold of 0.5, an empty threshold of 0.3, and an active sonar depth filter. A more strict Preset 2 was used in real-world experiments, with a circularity threshold of 0.56 and an empty threshold of 0.07. Preset 1, which did not use sonar depth filtering, was applied to the single pane experiments without occlusion. Table~\ref{tab:ablation_table} includes a comparison of all presets, including an extra version of Preset 3 without the sonar filter, validated on the Vicon Room (Autonomous) flight data.

\begin{table}[!ht]
    \centering
    \caption{Ablation table for Algorithm Preset Comparison}
    \label{tab:ablation_table}
    \begin{tabularx}{0.48\textwidth}{l l c c c}
        \toprule
        \textbf{Experiment Type} & \textbf{Preset} & \textbf{Precision} &
        \textbf{Recall} &
        \textbf{mIOU} \\
        \midrule
        \multirow{3}{*}{\shortstack[l]{\\ \\ \\ \\ Vicon Room\\(Autonomous)}} 
        & 1 & 83\% & 31\% & 19.4\% \\
        \cmidrule{2-5}
        & 2 & 96\% & 61.7\% & 56.8\%\\
        \cmidrule{2-5}
        & 3 (w/o sonar)  & 86.9\% & 45.9\% & 32.7\% \\
        \cmidrule{2-5}
        & \textbf{3} & 92.9\% & 72.3\% & 64.4\% \\
        \bottomrule
    \end{tabularx}
\end{table}


Regarding the qualitative results from the single glass pane experiments with background objects presented in Table~\ref{tab:performance_summary}, the minimal performance difference observed, particularly in the angled tests with and without sonar, is attributed to the speckle's location. In these specific scenarios, the speckle was not directly occluded by the background objects, allowing for successful detection even without the sonar filter, causing very similar results for both experiments.

\begin{figure}[!h]
\centering
\begin{tabular}{c c}
{\includegraphics[width = 0.31\linewidth]{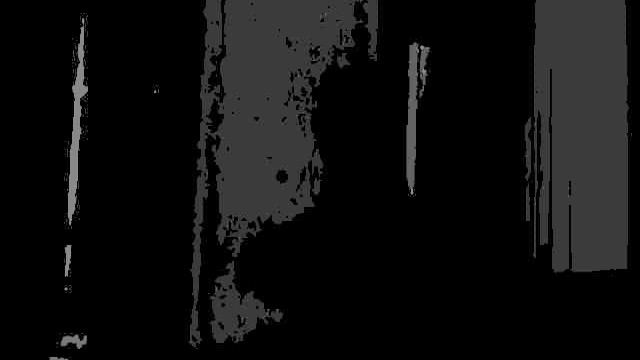}} &
{\includegraphics[width = 0.31\linewidth]{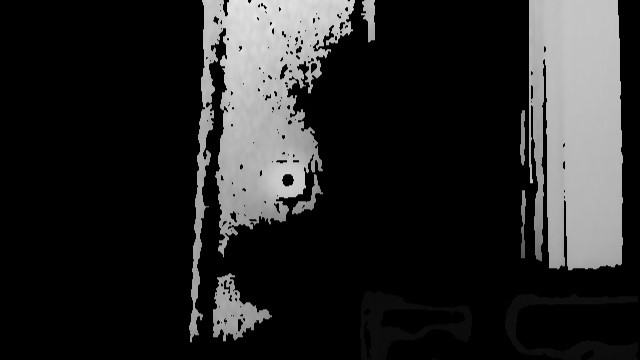}} \\
\subfloat[Raw Depth]
{\includegraphics[width = 0.31\linewidth]{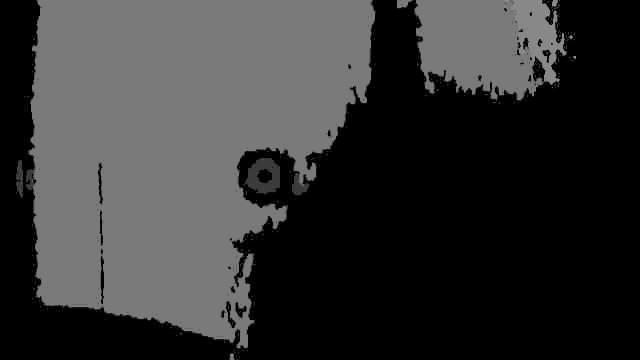}} &
\subfloat[Sonar-Filtered]
{\includegraphics[width = 0.31\linewidth]{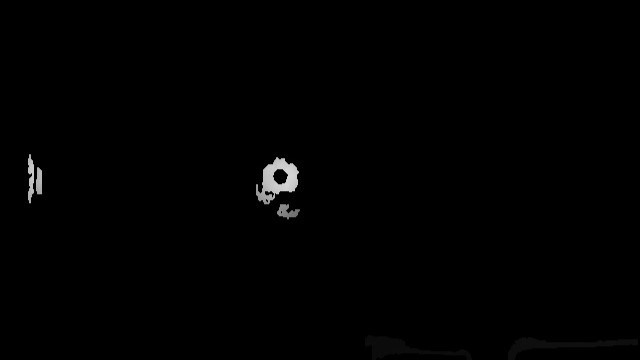}} \\
\end{tabular}
\caption{Comparison of sonar depth filter in (TOP) Glass Perimeter vs (BOTTOM) Vicon Room (Autonomous). 
\vspace{-2em}
}
\label{fig:occupancy_maps_ablation}
\end{figure}

As depicted in Fig.~\ref{fig:occupancy_maps_ablation}, a weakness in speckle detection is demonstrated by the Glass Perimeter experiment where a background object was so close to the glass that it prevented speckle detection. These objects were within the 0.1 m distance from the glass in which the speckle begins, making them impossible to filter by the sonar. Seven glass panes in the environment were occluded by such obstacles, leading to a significant decrease in performance metrics. In contrast, the sonar filter in the Vicon Room experiment successfully removed background objects that were further than this distance, allowing for better speckle detection.

\section{Conclusions}
\label{sec:conclusions}
In this work, we presented a novel method for transparent obstacle detection and mapping on a low-SWaP (Size, Weight, and Power) aerial robot. Our approach uniquely combines a custom 2D convolution for speckle detection in a Time-of-Flight depth image with a sonar-based depth filter to robustly identify glass surfaces, and reproject the speckle depth into segmented empty depth regions that contain it. This fusion technique effectively isolates speckles from background clutter and verifies their presence within a central region of interest, significantly enhancing the reliability of our detections and reprojections.
The algorithm was validated through a series of experiments, from controlled lab settings to complex real-world environments, demonstrating its ability to accurately and consistently map transparent planes. 
Our method successfully detected and reprojected glass surfaces during autonomous flight tests, with minimal performance degradation in comparison to handheld experiments.
The entire pipeline operates efficiently and in real-time on a resource-constrained embedded processor. 
This demonstrates our method's viability for onboard deployment and its computational superiority to more intensive, data-driven alternatives that require powerful GPUs. This makes it a practical and effective solution for enabling autonomous navigation in environments with transparent obstacles. 
To facilitate further research and development in this area, we will open-source our datasets and software. In future work, we plan to enhance our algorithm's robustness to handle more complex glass structures, and extend its capabilities to other transparent and reflective objects.


\bibliography{bibtex.bib}

\end{document}